\documentclass[times,11pt]{article} 
\usepackage[small,compact]{titlesec}
\usepackage{verbatim}
\usepackage{wrapfig}
\usepackage{mathrsfs}
\usepackage{amsmath,amssymb,latexsym}
\usepackage{amsfonts}
\usepackage{times,lastpage,bm,xspace}
\usepackage{mathrsfs,bbm}
\usepackage{dsfont}
\usepackage{amsmath,amssymb,algorithm}
\usepackage{stmaryrd}
\usepackage{booktabs}
\usepackage{algpseudocode}
\usepackage{multirow}
\usepackage{array}

\usepackage[caption=false,font=footnotesize]{subfig}

\usepackage{amsmath, amsfonts, amssymb, amsthm,color,bm}

\usepackage[pdftex]{graphicx}
\usepackage{url}

\def\ie{{\em i.e.,~}}
\def\eg{{\em e.g.,~}}
\def\cf{{\em cf.,~}}

\def\tol{{\mbox{\sc tol}}}
\def\deq{\triangleq}

\def\bI{\ones}

\def\cN{\mathcal{N}}

\def\cJ{\mathcal{J}}

\newcommand{\wt}[1]{{\widetilde{#1}}}
\def \reals{\mathbb{R}}

\def\bI{\mathbbm{1}}

\def\Kmax{K_{\mbox{max}}}

\begin{document}
	\begin{center}
	{\bf{\LARGE{Online Data Thinning via Multi-Subspace Tracking}}}

	\vspace*{.1in}
	\begin{tabular}{ccc}
	Xin Jiang$^{1}$ & Rebecca Willett $^{2}$ \\
	\end{tabular}

	\vspace*{.1in}

	\begin{tabular}{c}
	  $^1$ Department of Electrical and Computer Engineering, Duke University\\
	  $^2$ Department of Electrical and Computer Engineering, University of Wisconsin-Madison
	\end{tabular}

	\vspace*{.1in}


	\end{center}

\begin{abstract}
{In an era of ubiquitous large-scale streaming data,
  the availability of data
  far exceeds the capacity of expert human analysts.} In many settings,
  such data is either discarded or stored unprocessed in
  datacenters. This paper {proposes a method of} {\em online data thinning}, in
  which large-scale streaming datasets are winnowed to preserve
  unique, anomalous, or salient elements for timely expert analysis.
  At the heart of {this} proposed approach is an online anomaly detection
  {method} based on dynamic, low-rank Gaussian mixture
  models. Specifically, the high-dimensional covariances matrices
  associated with the Gaussian components are associated with low-rank
  models. {According} to this model, most observations lie near a
  union of subspaces. The low-rank modeling mitigates the curse of
  dimensionality associated with anomaly detection for
  high-dimensional data, and recent advances in subspace clustering
  and subspace tracking allow the proposed method to adapt to dynamic
  environments. {Furthermore, the proposed method allows subsampling, 
  is robust to missing data, and uses a mini-batch online optimization approach. 
  The resulting algorithms are scalable, efficient, and are capable of
  operating in real time.} Experiments on wide-area motion imagery and
  e-mail databases illustrate the efficacy of the proposed approach.
\end{abstract}

\section{Introduction}\label{sec:introduction}
Modern sensors are collecting
high-dimensional data at unprecedented volume and speed; human {analysts}
cannot keep pace.
For instance, many sources of intelligence data must be translated by human experts before they can be widely accessible to analysts and actionable; the translation step is a significant bottleneck \cite{sept11}. 
Typical NASA
missions collect terabytes of data every day
\cite{JPLBigData,WiredSKA,LSSTWeb,LSST}. {Incredibly,} the Large
Hadron Collider (LHC) at CERN ``generates so much data that scientists
must discard the overwhelming majority of it---hoping hard they've
not thrown away anything useful.'' \cite{LHC} There is a pressing need
to help analysts {prioritize} data {\em accurately and efficiently} from
a storage medium or a data stream. This task is complicated by the
fact that, typically, the data is neither thoroughly annotated nor
meaningfully catalogued. Failure to extract relevant data could lead
to incorrect conclusions in the analysis, while extraction of
irrelevant data could overwhelm and frustrate human analysts,
throttling the discovery process.

{This paper focuses on} {\em scalable online data processing
  algorithms that can winnow large datasets to produce smaller subsets
  of the most important or informative data for human analysts}. 
  {This process is described as ``data thinning.''} Often, the data thinning process
involves flagging observations which are inconsistent with previous
observations from a specified class or category of interest, or are
ranked highly according to a learned ranking function. Typically we
are interested in methods which can perform these assessments from
streaming data, as batch algorithms are inefficient on very large
datasets.

{One} generic approach to the problem of data thinning for large
quantities of (possibly streaming) high-dimensional data {requires
estimating and tracking} a probability distribution $f_t$ underlying the
stream of observations $x_t$, and {flagging} an observation as anomalous
whenever $\widehat{f}_t(x_t) < \tau$ for some small threshold
$\tau>0$, as demonstrated in past work
\cite{raginsky_OCP,onlineSocialAnomalies}. Ultimately, {the} goal is
to ensure that the flagged data is salient to human analysts on
the receiving end without being buried in an avalanche of irrelevant
data.  Within this general framework, {there are} three key
challenges:
\begin{itemize}
\item {\bf Dynamic environments:} The data may not be from a
  stationary distribution. For {example}, it may exhibit diurnal,
  location- or weather-dependent patterns. Effective data thinning
  methods must adapt to those dynamics and sources of bias. Global
  summary statistics and naive online learning algorithms will fail in
  this context.
\item {\bf High-dimensional data:} Individual data points $x_t$ may be
  high-dimensional, resulting in the classical ``curse of
  dimensionality'' \cite{bellman1961adaptive,hastie2009element}. While
  large quantities of data {may be} available, the
  combination of high-dimensional data and a non-stationary
  environment still results in an ill-posed estimation problem.
\item {\bf Real-time processing:} In applications {like those
  with NASA and CERN}, large quantities of streaming data preclude
  computationally intensive or batch processing.
\end{itemize}

\subsection{Data thinning for wide-area motion imagery}
While our approach is not restricted to imaging data, one important application of our data thinning approach is real-time video analysis. 
Recent advances in optical engineering have led to the advent of new
imaging sensors that collect data at an unprecedented rate and scale;
these data often cannot be transmitted efficiently or analyzed by
humans due to their sheer volume. For example, the ARGUS system
developed by BAE Systems is reported to collect video-rate gigapixel
imagery \cite{argus,argus2}, and even higher data rates are anticipated
soon \cite{mosaic3,mosaic4,mosaic2}. This type of data is often
referred to as wide-area motion imagery (WAMI). Currently WAMI streams
are used primarily in a forensic context -- after a significant event
occurs (\eg a security breach), the data immediately preceding the
event are analyzed {\em reactively} to piece together what led to that
event. However, there is a strong need for predictive analysis which
can be used to help {\em anticipate} or detect negative events in real
time.

Unfortunately, the latter form of analysis is often infeasible for two
reasons: {(1)} the data acquisition rate exceeds the capacity of
many sensor platforms' downlinks; and {(2)} size, weight, and power
constraints limit processing capabilities on airborne sensor
platforms.  Thus an {\em emerging and fundamental challenge is
  efficiently downloading salient information to ground-based analysts
  over a limited-bandwidth channel}. While data compression has a long
history, conventional compression methods may distort information
particularly relevant to analysts. In particular, standard motion
imagery compression techniques typically focus on optimizing peak
signal-to-noise ratio or psycho-visual metrics which apply globally to
an entire video and are often unrelated to any specific task.

\begin{figure}
  \centerline{\includegraphics[width=.6\textwidth]{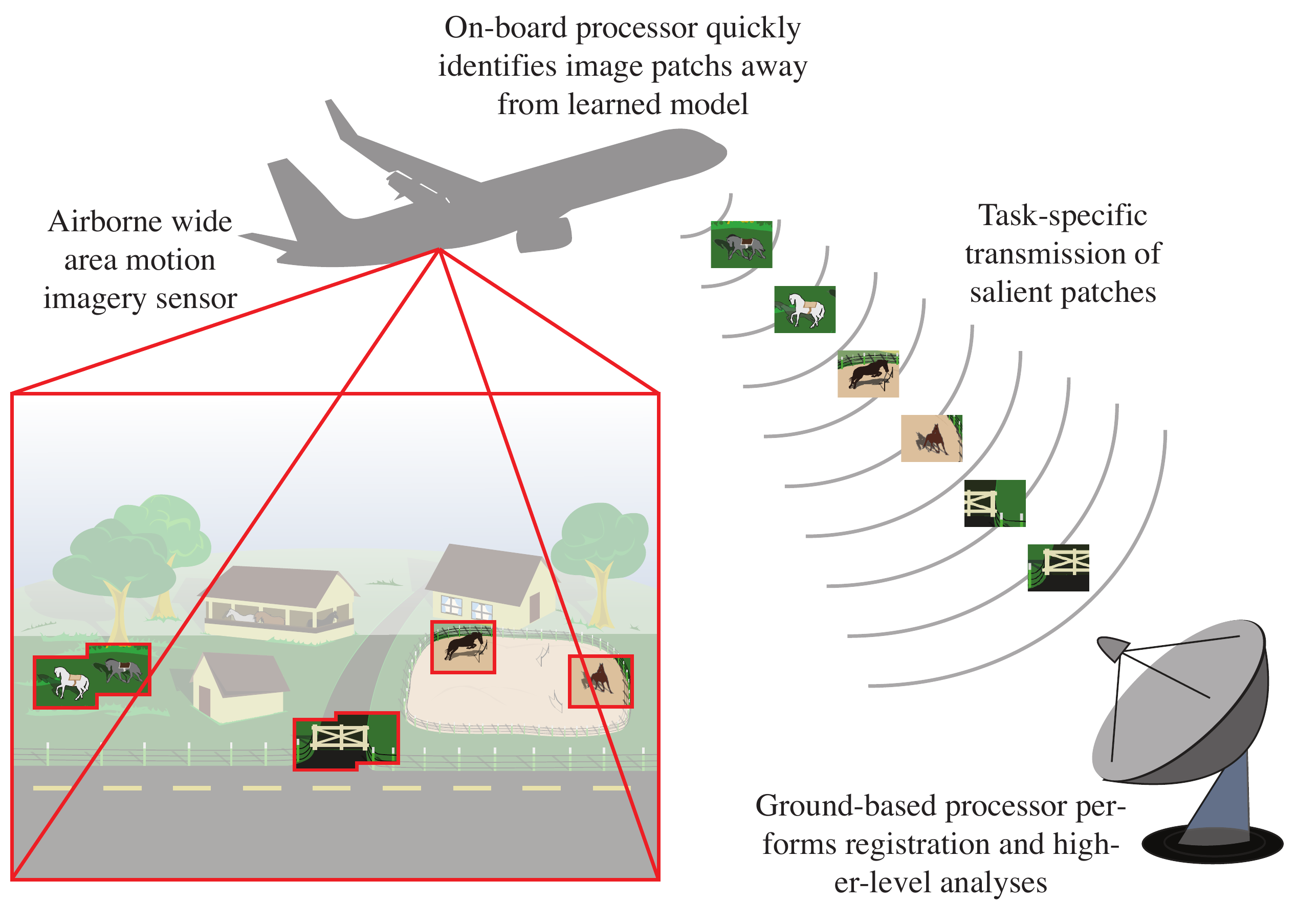}}
  \caption{{\small Conceptual illustration of proposed objectives. An airborne
    platform collects wide-area motion imagery (WAMI), identifies
    task-specific salient patches, and transmits only those
    patches. The ground-based receiver can then perform more
    sophisticated processing, including registration, geolocation, and
    activity analysis.}}
  \label{fig:concept}
\end{figure}

{Instead, a} better solution would be to identify unique objects or regions of
WAMI, and transmit only features of these objects.  This concept is
illustrated in Fig.~\ref{fig:concept}. Ideally, {this} method will
identify regions and features of a data stream most critical to a
given task, and prioritize these features when preparing data for
storage or transmission. This task is clearly related to ``visual
saliency detection'' (\cf
\cite{SaliencyItti,SaliencySR,SaliencyGBVS,rao2010using}); we describe
the connections between the proposed work and saliency detection in
Section~\ref{sec:RelatedWork}.

Note that in this setting a key challenge is that the sensor may be
placed on a vibrating platform that introduces significant jitter into
the data and precludes direct comparison of successive frames. While
real-time video stabilization has been considered in the video
processing literature (\cf
\cite{erturk2002real,hansen1994real,ratakonda1998real,chang2006robust,battiato2010robust}),
such methods are often robust for small motions associated with a
hand-held device and break down with large motions associated with
mechanical vibrations. More robust methods capable of processing
larger degrees of jitter can be computationally prohibitive on
energy-constrained platforms.

\subsection{Problem formulation and approach}
\label{sec:formulation}
Suppose we are given a sequence of data $x_1, x_2, \ldots,$ and for
$t = 1, 2, \ldots$, $x_t \in \mathbb{R}^{p}$, where $p$ denotes the
{\em ambient dimension}. Assume that $x_t$ {comes} from some unknown
distribution, \ie there exists some sequence of distributions $P_t$
such that
$$
x_t\sim P_t \quad t=1,2,\ldots
$$
where $P_t$ evolves over time, and its distribution density function
is denoted by $f_t$.  The goal is to find the $x_t$ that are unusual
or anomalous. In particular, we assign each observation $x_t$ {an} {\em
  anomalousness score} proportional to its negative log likelihood
under the estimated model---\ie $-\log f_t(x_t)$. Observations with a
low anomalousness score can then either be directed to a human
{analyst} or flagged for further processing and analysis.

The key challenge here is two-fold: (a) the dimension of the signal,
$p$, can be quite large, and (b) $f_t$ may evolve rapidly over
time. The combination of these factors means that our problem is
ill-posed, because we are unlikely to gather enough samples to
reliably learn the density $f_t$.

{This paper proposes} a method for estimating and tracking the
time-series of density functions $f_t$ over $\mathbb{R}^p$. In
stationary, low-dimensional settings, we might consider a Gaussian
mixture model that could be estimated, for instance, using an online
expectation-maximization (EM) algorithm
\cite{same2007online}. However, the non-stationary setting and high
dimensions make that approach unviable, as we demonstrate
experimentally later in the paper. {The proposed approach, by contrast,
considers} a
constrained class of Gaussian mixture models in which the Gaussian
covariance matrices (each in the positive-semidefinite cone
$\mathcal{S}^{p}_{+}$) are low-rank. This model is equivalent to
assuming most $x_t$ lie near a union of low-dimensional
subspaces. While this union of subspaces is unknown {\em a priori}, we
may leverage recent advances in subspace tracking (\cf
\cite{BalzanoNowakRecht2010,ChiJournal2012,ChiEldarCalderbank2012,roseta})
and subspace clustering (\cf \cite{aggarwal2000finding, bradley2000k,
  bohm2004computing, achtert2006mining, achtert2007robust,
  achtert2007exploring}) to yield an accurate sequence of density
estimates $\hat f_t$, and mitigate the curse of
dimensionality.

In addition, we consider certain computational {and} statistical
  tradeoffs associated with the data thinning problem. In particular,
{there are two ways to} reduce the computational complexity associated with computing
anomalousness scores. {First,} we can reduce the frequency with which we update our
model. {Second, we can} subsample the elements of each $x_t$ and leverage missing
data models for fast calculations and updates. We demonstrate that
these methods, which are not amenable to standard stochastic filtering
methods, can yield significant computational speedups with only small
decreases in thinning accuracy.

\subsection{Contributions and paper layout}
{This paper presents} a data thinning method for high-dimensional
streaming data in a dynamic environment. The algorithm adapts to
changing environments using tracking methods for union of
subspaces. {As shown by} both synthetic and real-data experiments,
{the} algorithm (a) efficiently tracks the subspaces in which most
observation lie and hence precisely detects observations that occur with low
probability, and (b) can be applied to a variety of real-world
applications and tasks.

{Section~\ref{sec:RelatedWork} describes related work.} Section~\ref{sec:model} explains {the}
probability density model based on unions of subspaces, and
Section~\ref{sec:tracking} presents the algorithm for tracking such
densities. Section~\ref{sec:computation} describes the computational
and statistical tradeoffs associated with the proposed approach.
Section~\ref{sec:syncexp} {reports} synthetic experiments
{which demonstrates} the ability of {the} algorithm to precisely track the
density and detect anomalous signals within a changing environment. 
Section~\ref{sec:WAMIexp}, {tests the} algorithm on the wide-area
motion imagery (WAMI) videos {to detect} salient objects, while
Section~\ref{sec:Enronexp} {tests the} algorithm on the Enron email
database to detect major events.

\section{Related work}
\label{sec:RelatedWork}
While data thinning is an emerging concept associated with modern
high-dimensional, high-velocity data streams, the formulation described
in Section~\ref{sec:formulation} is closely related to anomaly
detection, visual saliency detection, and subspace clustering and
tracking.

\subsection{Anomaly detection} 
The study of anomaly detection has a long and rich history, where the
earliest papers can date back to the 19th century
\cite{edgeworth1887xli}.  Despite the long history of the study of
anomaly detection, most existing detection methods {do not work well with high
dimensional data}, and often do not work online.  A 2009 survey on
anomaly detection \cite{chandola2009anomaly} categorizes the available
methods into classification-based methods, nearest neighbor-based
methods, cluster-based methods, information theoretic methods,
statistical anomaly detection methods, and spectral methods.

Among the six categories, classification based methods (\cf
\cite{de2000reject, barbara2001detecting, scholkopf2001estimating,
  roth2004outlier, roth2006kernel}) require a large training pool with
labeled data that is typically unavailable in the settings of
interest. Also, {the classification based methods depend highly} on the training data, 
{and} do not {effectively} adapt to changing environments. Nearest neighbor (\cf
\cite{zhang2006detecting, otey2006fast, ghoting2008fast,
  tao2006mining, wu2006outlier}) and cluster-based methods (\cf
\cite{ertoz2004finding, yu2002findout, budalakoti2006anomaly,
  pires2005using}) can both be extended to work online, but the
computational costs are usually high, scaling with the amount of
data. Furthermore, the performance of the nearest neighbor and
cluster-based methods highly depend on the distance measure, and the
optimal distance measure is highly problem-dependent.

Certain statistical methods (\cf \cite{solberg2005detection,
  aggarwal2008outlier, chen2005simultaneous, agarwal2007detecting})
assume that the data are drawn from some standard or predetermined
distribution, and determines outliers by computing the likelihood of
the signal coming from such distributions. These methods can often
work online, and do not rely on a big training set, but estimating the
distribution of high-dimensional data is a non-trivial task, and the
statistical assumptions {do} not always hold true, especially for
high-dimensional data where there could be spatial correlations.

Information theoretic techniques (\cf \cite{he2005optimization,
  ando2007clustering, keogh2004towards}) identify the anomalies by
trying to find a small subset such that removing the subset will
greatly reduce the complexity of the whole set. The approach requires
no supervision, and does not make assumptions about the underlying
statistical distribution of the data. However, they usually have
exponential time complexity and are batch methods. Additionally, it is
difficult to assign anomalousness scores to a single data point.

Spectral methods (\cf \cite{agovic20086, dutta2007distributed,
  ide2004eigenspace, shyu2003novel}) assume that data can be embedded
into a lower dimensional subspace, and detect anomalies over the
embedded space rather than the original space. Because {spectral methods}
essentially operate on a reduced-dimensional representation of the
data, {they} are well-suited to high-dimensional
data. Spectral methods can also be integrated with other methods, and
are thus highly versatile. However, spectral methods can incur high
computational costs; even online anomaly detection algorithms (\cf
\cite{breunig2000lof}, \cite{kriegel2008angle} and
\cite{ahmed2009online}) face this challenge.  Furthermore, the
subspace model underlying spectral methods is less flexible than the
union of subspace model underlying {this paper's proposed} method.

\subsection{Visual saliency detection}
In the special case of imagery or video data, data thinning is closely
related to visual saliency detection.  Like anomaly detection,
saliency detection has been widely studied over the last few decades.
A standard benchmark for comparison in image saliency detection is
proposed by Itti et al. in \cite{SaliencyItti}. {This} paper attempts to
explain human visual search strategies, using biologically motivated
algorithms. However, this algorithm is too slow to apply to real time
videos.  Hou and Zhang in \cite{SaliencySR} use spectral analysis to
detect salient objects for faster speed. However, {the analysis} breaks down when
multiple types of salient objects are present in the scene.
Graph-based methods (\cf \cite{SaliencyGBVS}) work well even when
there is no central object in the scene, which is often difficult for
other methods to handle, but suffers from high computational
complexity.  Rao et al. proposed a cluster-based algorithm in
\cite{rao2010using}, where the salient object is identified by first
clustering all the pixels according to their local features, and then
finding the group of pixels that contains the most salient
information. It works better than \cite{SaliencyItti}, but not as well
as the graph-based algorithms.  The information theoretic model based
algorithm proposed in \cite{SaliencyAIM} claims to work as well as
\cite{SaliencyItti}, but requires less tuning. This work is
improved in \cite{zhang2008sun} for faster speed and better
performance.

Methods for image saliency detection have been extended to video
saliency detection, but those methods assume a stable imaging platform
and video stream free of jitter. In the WAMI application described
above, however, sensors can be placed on vibrating platforms that
preclude most video saliency detection methods.
 
\subsection{Subspace clustering and tracking}
{The proposed method} is also closely related to the subspace clustering and
tracking algorithms. Subspace clustering is a relatively new, but
vibrant field of study. These methods cluster observations into
low-dimensional subspaces to mitigate the curse of dimensionality,
which often make nearest-neighbors-based methods inaccurate
\cite{beyer1999nearest}. Early works in the field can only identify
subspaces that are parallel to the axes, which is not useful when the
data is not sparse, but lives on an arbitrarily oriented
hyperplane. Newer methods \cite{aggarwal2000finding, bradley2000k,
  bohm2004computing, achtert2006mining, achtert2007robust,
  achtert2007exploring}, which are also called correlation clustering
methods, can identify multiple arbitrarily angled subspaces at the
same time, but all share the same problem of high computational
cost. Even \cite{achtert2007exploring}, which is shown to beat other
methods in speed, still has an overall complexity of $O(p^2T^2)$,
where $p$ is the dimension of the problem, and $T$ is the total number
of data points. More recent methods based on sparse modeling (\cf
\cite{elhamifar2009sparse,elhamifar2013sparse,wang2016noisy,
  vidal2010tutorial,groupsparse_ssp}) require solving convex
optimization problems that can be inefficient in high-dimensional
settings.  Thus, the high complexity of the algorithms make them less
than ideal candidates for an efficient online algorithm.

Subspace tracking is a classical problem that experienced recent
attention with the development of algorithms that are robust to
missing and outlier elements of the data points $x_t$.  For example,
the Grassmannian Rank-One Update Subspace Estimation (GROUSE)
\cite{BalzanoNowakRecht2010}, Parallel Estimation and Tracking by
REcursive Least Squares (PETRELS)
\cite{ChiJournal2012,ChiEldarCalderbank2012}, and Robust Online
Subspace Estimation and Tracking Algorithm (ROSETA) \cite{roseta}
effectively track a single subspace using incomplete data
vectors. These algorithms are capable of tracking and adapting to
changing environments.  The subspace model used in these methods,
however, is inherently strong, whereas a plethora of empirical studies
have demonstrated that high-dimensional data often lie near manifolds
with non-negligible curvature
\cite{allard2012multi,roweis2000nonlinear,belkin2003laplacian}.

In contrast, the non-parametric mixture of factor analyzers
\cite{ChenSilvaPaisley2012} uses a mixture of low-dimensional
approximations to fit to unknown and spatially-varying (but static)
curvatures. The Multiscale Online Union of SubSpaces Estimation
(MOUSSE) method developed by Xie et al. \cite{xie2013change} employs
union of subspaces tracking for change point detection in
high-dimensional streaming data. Thanks to the adoption of the
state-of-the-art subspace tracking techniques, the algorithm is both
accurate and efficient (with complexity linear in $p$).  However,
MOUSSE cannot be directly applied for our data thinning task for a few
reasons. First, MOUSSE is designed for change-point detection and does
not have a probabilistic model. Thus observations in a rare subspace
would still be treated as typical, which makes it difficult to
discover the rare observations. Second, MOUSSE can only process one
observation at a time, \ie it does not allow for mini-batch updates
that can be helpful in data thinning applications, where data could
arrive in blocks. Last but not least, although MOUSSE is able to deal
with missing data, \cite{xie2013change} does not explore the
computational-statistical tradeoffs that are important for time- or
power-sensitive applications. {This paper presents} a method
that is designed for the data thinning task, has a specific
statistical model, and allows for mini-batch updates which increases
the algorithm's efficiency. Also, we will explore the
computational-statistical tradeoffs in Section~\ref{sec:computation}.

\section{Data thinning via tracking union of subspaces}
\label{sec:tracking}

\subsection{Union of subspaces model}
\label{sec:model}
Recall from Section~\ref{sec:formulation} that each $x_t \in \reals^p$
is assumed {to be} drawn from a distribution with density $f_t$, and that
$f_t$ is modeled as a mixture of Gaussians where each Gaussian's
covariance matrix is the sum of a rank-$r$ matrix (for $r<p$) and a
scaled identity matrix. We refer to this as a {\em dynamic low-rank GMM}. In
particular, the $j^{\rm th}$ Gaussian mixture component is modeled as
$$\cN\left(\mu_{j,t},\Sigma_{j,t}\right)$$
where $\mu_{j,t} \in \reals^p$ is the mean and 
$$\Sigma_{j,t} = V_{j,t} \Lambda_{j,t} V_{j,t}^T + \sigma_j^2 I.$$
Here $V_{j,t} \in \reals^{p \times r}$ is assumed to have orthonormal
columns, and $\Lambda_{j,t} \in \reals^{r \times r}$ is a diagonal
matrix with positive diagonal entries.
If $\sigma_j = 0$, then $\Sigma_{j,t}$ would be rank-$r$ and any point
drawn from that Gaussian would lie within the subspace spanned by the
columns of $V_{j,t}$ -- shifted by $\mu_{j,t}$. By allowing $\sigma_j
> 0$ we model points drawn from this Gaussian lying near
that $r$-dimensional shifted subspace.
Overall, we model
\begin{equation}
f_t = \sum_{j=1}^{K_t} q_{j,t} \cN\left(
\mu_{j,t}, V_{j,t} \Lambda_{j,t} V_{j,t}^T + \sigma_j^2 I
\right)
\label{eq:GMMmodel}
\end{equation}
where $K_t$ is the number of mixture components in the model at time
$t$ and $q_{j,t}$ is the probability of $x_t$ coming from mixture component $j$.

To better understand this model, we can think of each observation
  $x_t$ as having the form $v_t + w_t$, where $v_t$ lies in a union of
subspaces (or more precisely, because of the Gaussian means, a union
of hyperplanes) defined by the $V_{j,t}$s and within  ellipsoids
embedded in those hyperplanes, where the ellipsoid axis lengths are
determined by the $\Lambda_{j,t}$s. 

\begin{figure}
  \centering
  \subfloat[Image of a pedestrian walking on a road with trees on the sides]{\includegraphics[width=.45\textwidth]{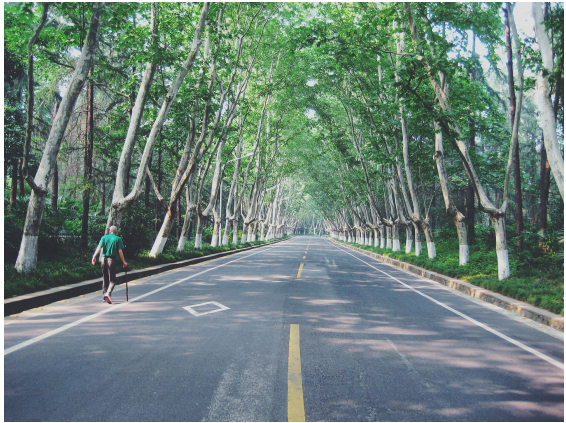}\label{fig:biker}} ~
  \subfloat[Illustration of the union of subspaces
  idea]{\includegraphics[width=.45\textwidth]{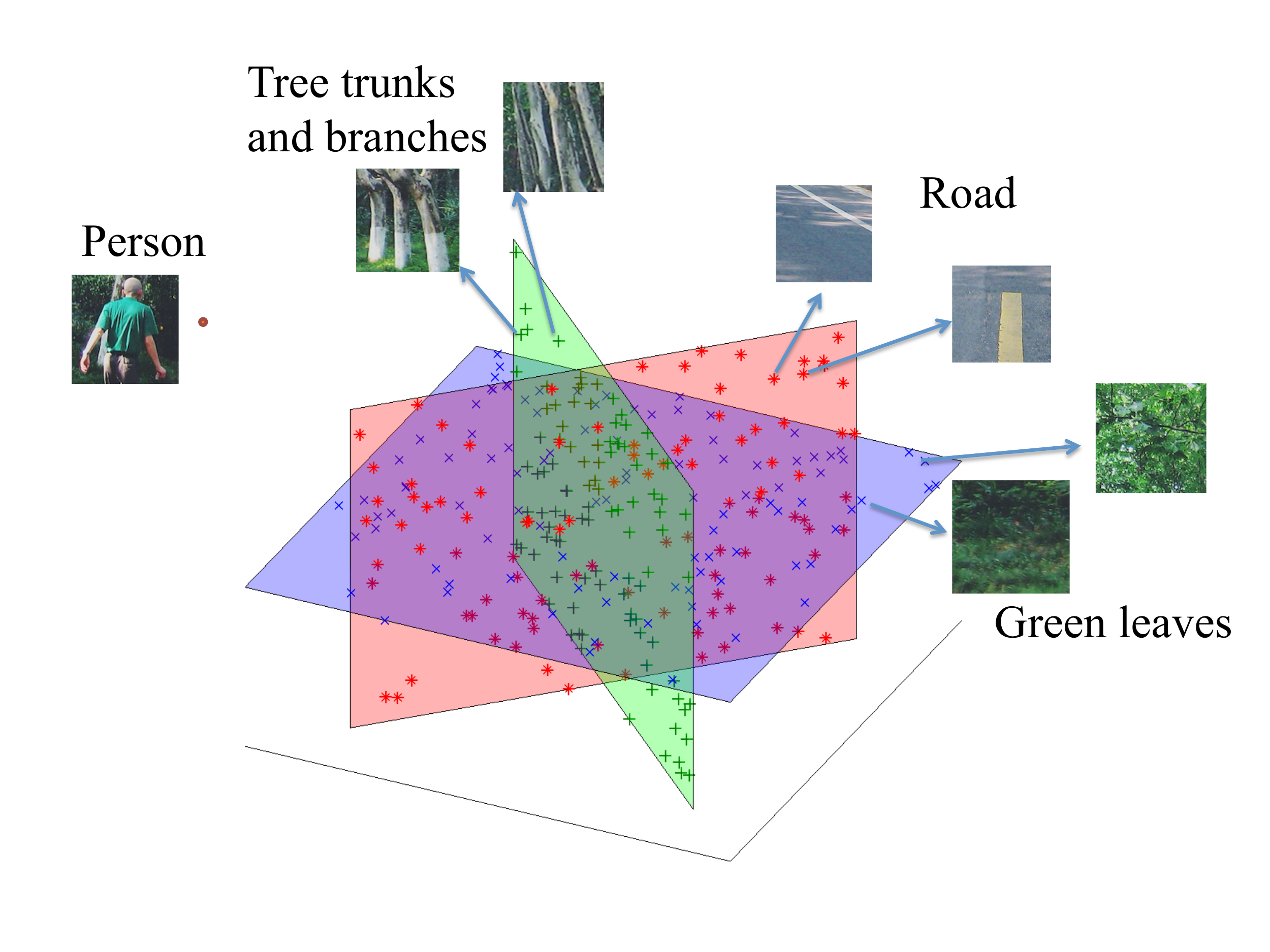}\label{fig:bikerill}}
  \caption{{Illustration of the union of subspaces
      idea. Fig.~\ref{fig:biker} shows a pedestrian walking on a road
      with trees on the sides \cite{sampleimage}. The road and the
      plants occupy most of the pixels, and they can be considered
      living in a union of subspaces. The person on the road
      would be considered as an outlier. }}
\label{fig:subspaceill}
\end{figure}

Fig.~\ref{fig:subspaceill} illustrates the union of subspaces
model. Fig.~\ref{fig:biker} shows a sample image where one person is
walking on a road with trees on both sides \cite{sampleimage}. In such
a situation, we would want to be able to learn from a sequence of such
images that the trees, grass and the road which occupy most of the
pixels are typical {of} the background, and label the person as salient
because it is uncommon in the scene. Fig.~\ref{fig:bikerill}
illustrates the union of subspaces model. When we divide the image
into patches, the vast majority of patches are plant, and road
patches, and only a few patches contain the person. Thus, the plant
and road patches live on a union of subspaces as illustrated and can
be thinned, leaving anomalous patches for further analysis.

\subsection{Algorithm highlights}
\begin{figure}
\centering
    \includegraphics[width=.45\textwidth]{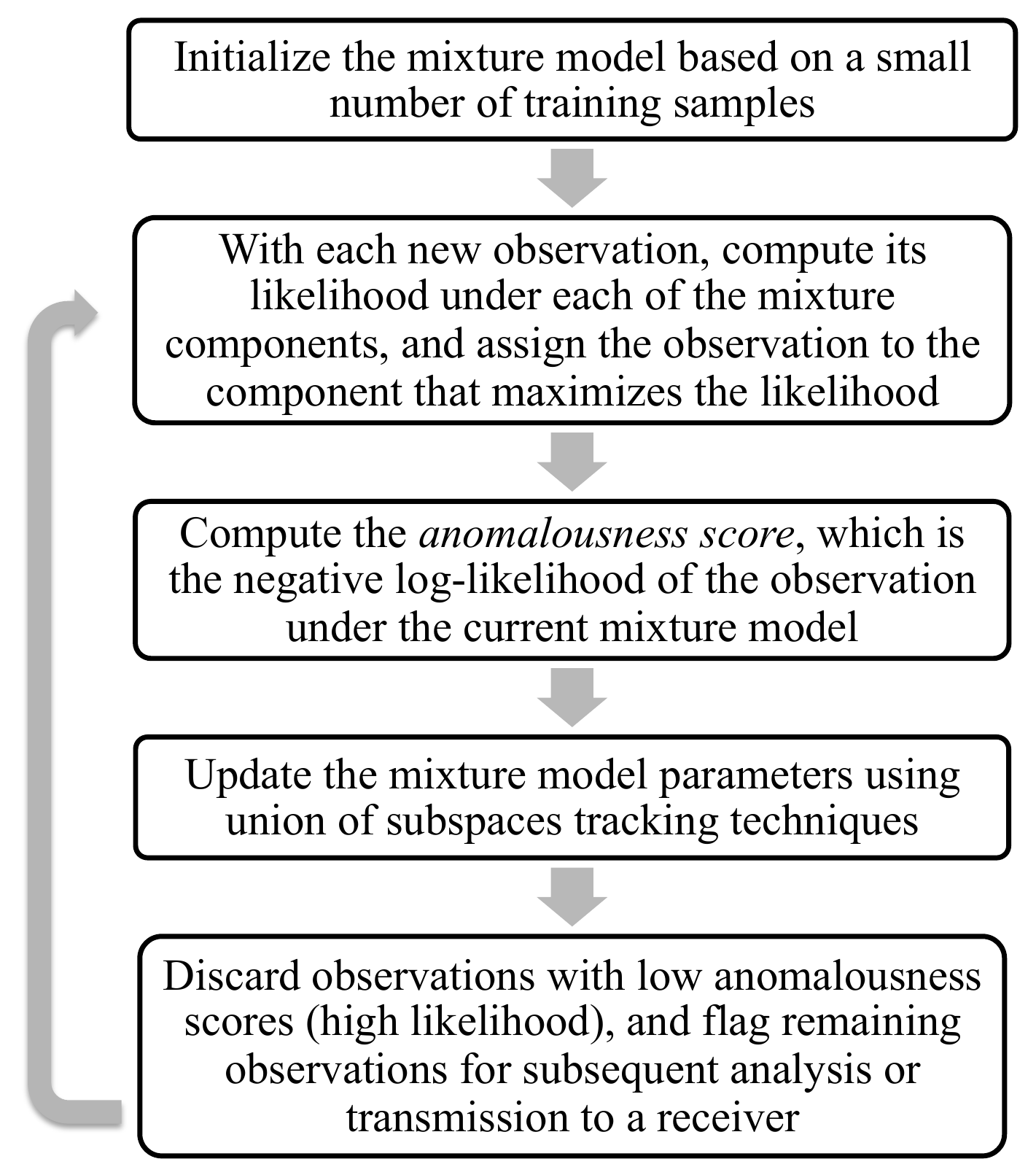}
  \caption{Flow chart of the main steps in the data thinning method.}
  \label{fig:flowchart}
\end{figure}

{This section explains how the proposed method estimates} the evolving Gaussian
mixture model using the techniques from the union of subspaces
tracking algorithms. 
These steps are summarized in in Fig.~\ref{fig:flowchart}.
As seen, {this} data thinning method shares some features with the online EM
algorithm for GMM estimation. However, there are a few key differences
{which are elaborated} below:
\begin{itemize}
\item We constrain covariances to lie in a {\em union of subspaces},
  which significantly reduces the problem size for estimating the
  covariance matrices. This constraint improves the accuracy of the
  algorithm, and also makes our method much stabler when the
  environment is changing rapidly relative to the data
  availability. This constraint also reduces computation time. (More
  details of computational complexity are discussed in
  Section~\ref{sec:complexity}.)
\item In some settings, such as when working with WAMI data, we
  receive groups of $x_t$'s simultaneously and can perform model
  updates more efficiently using {\em mini-batch techniques}. (The
  mini-batch approach is discussed in Section~\ref{sec:Multi}.) 
\item For large, high-velocity data streams, real-time processing
  {is} paramount. Even evaluating the likelihood of each new observation
  can be time consuming. We explore {\em subsampling-based
    approximations} which reduce computational burden yet still yield
  accurate results.  (Accuracy and computational complexity tradeoffs
  are discussed in Section~\ref{sec:computation}.)
\item For the online EM algorithm for GMM estimation, the number of
  mixture components is selected {\em a priori}, and does not change for the
  duration of the task. This would work when the environment does not
  change over time, but is inappropriate for applications that work in
  dynamic environments. {The proposed} method {\em adapts to changing numbers of
    mixture components}, which allows the mixture model to better
  track the environmental dynamics. {The method} adapts the number of mixture
  components using a multiscale representation of a hierarchy of
  subspaces, which allows us to reduce the model order using a simple
  complexity regularization criterion. {The method} also tracks hidden subspaces
  which are then used to increase the model order when data
  calls for it. (More details about the multi-scale model is discussed
  in Section~\ref{sec:algorithm}.)
\end{itemize}

\subsection{The Online Thinning algorithm}
\label{sec:algorithm}

This section describes the updates of the parameters associated with
{the proposed} dynamic low-rank GMM in \eqref{eq:GMMmodel}. The updates of the
mixture component weights ($q_{j,t}$) and means ($\mu_{j,t}$) are
computed using stochastic gradient descent. The updates of the
covariance matrices are more sophisticated and leverage subspace
tracking methods. In particular, we focus on methods which admit
observations $x_t$ with missing elements; this will allow us to
subsample $x_t$ for computational speedups. These updates are detailed
below.

The biggest challenge is updating $K_t$, the number of mixture
components.  In real-life applications, the number of mixture
components is in general (a) not known {\em a priori}, and (b) can
change with $t$.  Thus a mechanism for adaptively choosing the number
of subspaces is needed.  Reducing model order is slightly less
challenging because it is relatively simple to merge two nearby
mixture components. However, increasing model order is a much more
complex issue, especially in an online setting.

\begin{figure}
\centering
\includegraphics[width=.45\textwidth]{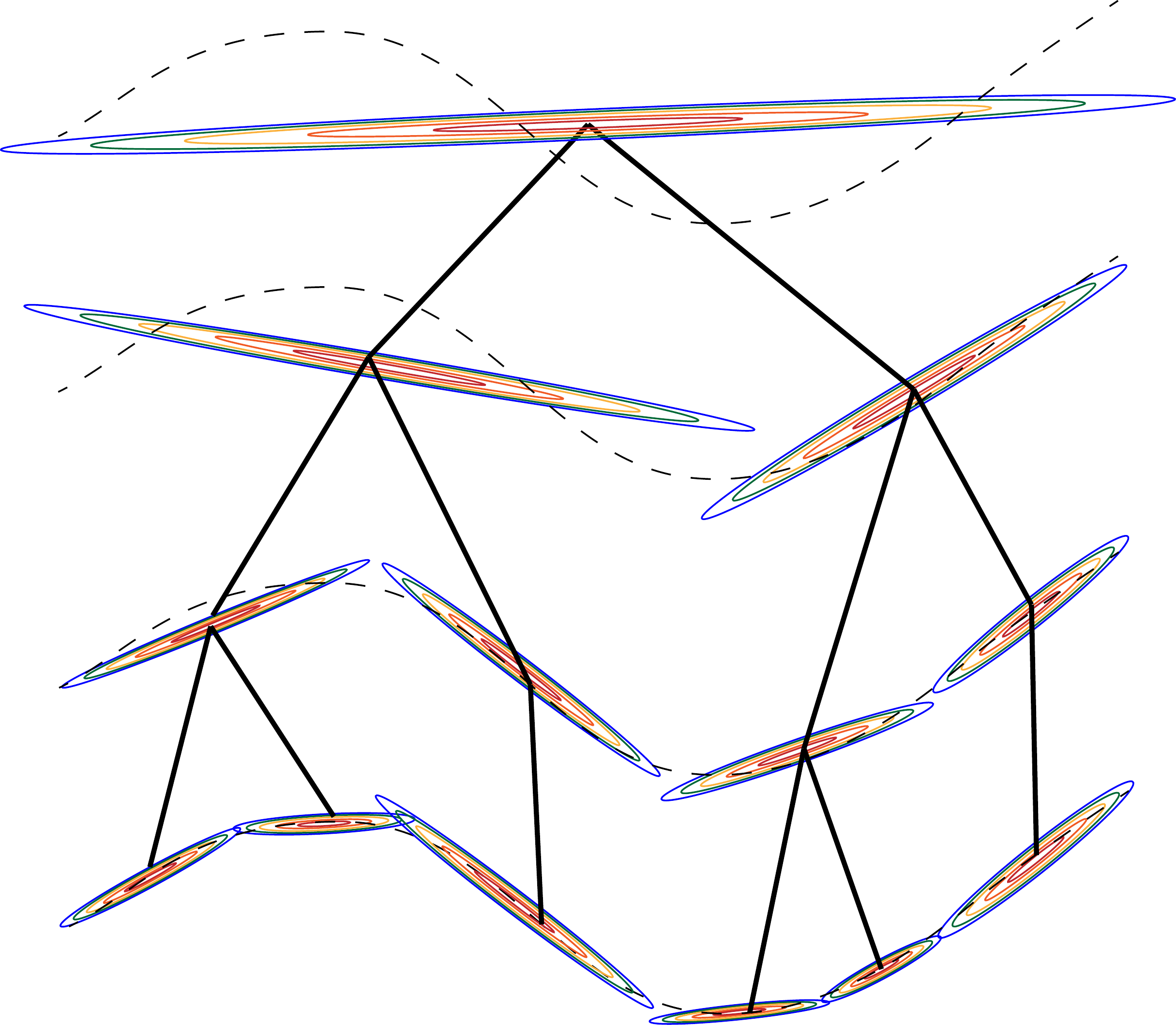}
\caption{Multiscale representation of low-rank Gaussian mixture
  model. Consider a density with its mass concentrated along the black
  dashed curve. Each successive level in the multiscale representation
  has more Gaussian mixture components (depicted via contour plots)
  with covariance matrices corresponding to more compact ellipsoids,
  and hence yields a more accurate approximation of the underlying
  density. Given a particular binary tree representation of a GMM, the
  approximation error can be allowed to increase or decrease by
  pruning or growing the binary tree connecting the different
  scales. The ellipsoids are all very compact along some axes because
  they correspond to covariance matrices that are the sum of a
  low-rank matrix and a scaled identity matrix.}
  \label{fig:tree}
\end{figure}

To address these challenges, we organize these mixture components
using a tree structure, as illustrated in Fig.~\ref{fig:tree}.  The
idea for a multiscale tree structure stems from the multiscale
harmonic analysis literature \cite{donoho97cart} and online updates of
such models are introduced in \cite{xie2013change}. In our setting, at
time $t$, the $j^{\mbox{th}}$ node is associated with a Gaussian
distribution parameterized by its mean vector $\mu_{j,t}$, low-rank
covariance matrix parameters $V_{j,t}, \Lambda_{j,t}$, and weight
$q_{j,t}$. Most of the probability mass associated with each Gaussian
is an ellipsoid centered at $\mu_{j,t}$, where $V_{j,t}$ and
$\Lambda_{j,t}$ characterize the principle axes and principal axis
lengths, respectively, of the ellipsoid. Finally, $q_{j,t}$ is
approximately the probability of an observation falling inside this
ellipsoid.

In the tree structure, we denote the set of leaf nodes as
$\cJ_t\deq\{j:j^{\mbox{th}} \mbox{ node is a leaf node at time } t\}$
and have $K_t \deq |\cJ_t|$.  The leaves of the tree correspond to the
Gaussian mixture components in the model shown in
Eq.~\eqref{eq:GMMmodel}. Each parent node corresponds to a single
Gaussian which approximates the weighted sum of the Gaussians
associated with its two children, where the weights correspond to the
children's $q$ parameters. Each of the tree leaves is also associated
with two {\em virtual} children nodes. The virtual children nodes
correspond to their own Gaussian distributions that can be used to
grow the tree. The decision of pruning and growing are made based on
(a) the accuracy of the Gaussian mixture model, \ie the cumulative
(with a forgetting factor) anomalousness score, and (b) the size of
the mixture model, \ie the total number of leaf nodes at time $t$.

\subsubsection{Computation of the Gaussian mixture likelihood (and
  anomalousness score)} 
{The proposed} algorithm uses the negative log-likelihood of {the} Gaussian mixture
model give the data point as its anomalousness score. 
The likelihood of $x_t$ under the Gaussian associated with
node $j$ is given by {(recall $\Sigma_{j,t} = V_{j,t} \Lambda_{j,t} V_{j,t}^T + \sigma_j^2 I$)}
\begin{equation}
 p_{j,t}(x_t)=\frac{1}{(2\pi)^{p/2}|\Sigma_{j,t}|^{1/2}}e^{-\frac{1}{2}(x_{t}-\mu_{j,t})^T{\Sigma}_{j,t}^{-1}(x_{t}-\mu_{j,t})}.
 \label{eq:likelihood}
\end{equation}

Using {the} model
in Eq.~\eqref{eq:GMMmodel}, the Gaussian mixture negative
log-likelihood function (and hence anomalousness score) for any $x_{t}\in\reals^{p}$ is:
\begin{equation}
\begin{aligned}
s_t(x_t) =& -\log f_t(x_t)\\
=& -\log \left(\sum_{j\in \cJ_t}q_{j,t} p_{j,t}(x_{t})\right).
\label{eq:score}
\end{aligned}
\end{equation}
\subsubsection{Selective update}
With the observation of each $x_t$, {the algorithm} first compute the likelihood of
$x_t$ under each of the Gaussian mixture components, and then assign
$x_t$ to the component that maximizes the likelihood. Specifically,
after the likelihood computations above, $x_t$ is assigned to the
mixture component
\begin{equation}
j^*_{t}\deq\arg\max_{j\in\cJ_t}\{p_{j,t}(x_{t})\}.
\label{eq:j_star}
\end{equation}
Note that {the weights $q_{j,t}$ are not used} here in order to avoid biasing towards
components with large weights.  {This} assignment is made in order to
reduce the computational complexity of the parameter update step: with
each $x_t$, instead of updating all the parameters of the entire tree,
{the algorithm} only updates the tree branch associated leaf node $j_t^*$. That is,
{the algorithm} updates the parameters of node $j_t^*$, all of its ancestors, and
one of node $j_t^*$'s virtual children (the one under which $x_t$ is
more likely). This approach significantly reduces the time complexity
of the updates, especially when the model is complex (\ie when the
number of leaf nodes is large).

\subsubsection{Mini-batch update} 
\label{sec:Multi}
In previous sections, we have always assumed that we have one
observation $x_t\in\reals^p$ arriving at each time $t$. However, in
many applications, multiple observations can arrive
simultaneously. For example, in WAMI settings, hundreds of image
patches in a single frame arrive at the same time. One way to deal
with this is simply treat each patch as arriving at a different time,
and update the model parameters separately with each
observation. However, when the number of patches is large (for HD
videos, {there can be} thousands of patches per frame), this sequential
processing can be extremely time-consuming.

To reduce the computation cost, we can instead update the mixture
model in mini-batches, \ie when multiple observations {are received} at
the same time, we first compute the anomalousness score of each
observation, and assign them to their own mixture component. The
collection of observations assigned to a given mixture component then
form a mini-batch. {The} mixture model and tree
structure {are then updated} only once for each mini-batch. When the size of mini-batches
is much larger than 1 (\eg hundreds of image patches assigned to a
tree of size $K_t=10$), this approach significantly reduces the number
of times {needed} to update the mixture component parameters and tree
structures, and thus saves computation time. Note that this mini-batch
processing does not affect the computation of the anomalousness score
and component assignment, where each observation {is processed}
sequentially as if they arrive separately.

Thus, now instead of assuming a single vector $x_t$ arrives at time
$t$, we assume that we receive a collection of observations stored in
matrix $X_t = [x_{t,1},\ldots,x_{t,N_t}]\in\reals^{p\times N_t}$ at
time $t$, where $x_{t,i} \in\reals^p$ for all $i=1,\ldots,N_t$. A
special case of this is $N_t = 1$, which is the sequential update
without mini-batches. After assigning each column in $X_t$ to the
$K_t$ leaf nodes in the hierarchical tree structure based on their
distance to the corresponding mixture components, we can rewrite $X_t$
into mini-batches, $X_t= [X_{j_{1},t},\cdots, X_{j_{K_t},t}]$, where
$\{j_1,\ldots,j_{K_t}\} \subseteq \cJ_t$. Here each
$X_{j_i,t}\in\reals^{p\times n_{j,t}}, i=1,\ldots,K_t$ is a block of
$n_{j,t}$ data points that are assigned to the $j_i^{\mbox{th}}$ node
in the tree (must be a leaf node). Note that
$\sum_{j\in\cJ_t} n_{j,t} = N_t$.

Our update equations are based on a ``forgetting factor''
$\alpha \in (0,1)$ that places more weight on more recent
observations; this quantity affects how quickly a changing
distribution is tracked and is considered a tuning parameter to be set
by the end user.  Then for each leaf node $j$ that needs updates (\ie
with assigned observations), the weights $q_{j,t}$ are then updated by
\begin{equation}
q_{j,t+1} = \alpha q_{j,t} + (1-\alpha)\frac{n_{j,t}}{N_{t}}.
\label{eq:q}
\end{equation}
Note that for the leaf nodes the weights need to add to 1, \ie
$\sum_{j\in\cJ_t} q_{j,t} = 1$ for all $t$. If we initialize $q_{j,1}$
s.t. $\sum_{i\in\cJ_1} q_{j,1}=1$ , and the weight of any parent node
is the sum of the weights of its two children, then this update
preserves $\sum_{i\in\cJ_t} q_{j,t}=1$ for all $t$.  The mixture
component means $\mu_{j,t}$ are updated by
\begin{equation}
\mu_{j,t+1} = \alpha \mu_{j,t} + \frac{(1-\alpha)}{n_{j,t}}X_{j,t}\bI_{n_{j,t}\times 1}.
\label{eq:mu}
\end{equation}
The diagonal matrix 
\[{\Lambda}_{j, t} \triangleq \mbox{diag}\{\lambda_{j, t}^{(1)}, \ldots, \lambda_{j, t}^{(r)}\} \in \mathbb{R}^{r\times r},\]
with $\lambda_{j, t}^{(1)}, \ldots, \lambda_{j, t}^{(r)}\ge 0$, contains eigenvalues of the
covariance matrix of the projected data onto each subspace. Let 
\begin{equation}
M_{j,t} = [\mu_{j,t},\ldots,\mu_{j,t}] \in\reals^{p\times n_{j,t}}
\label{eq:M}
\end{equation}
be a means matrix computed by concatenating $n_{j,t}$ copies of $\mu_{j,t}$ together. Let
\begin{equation}
B_{j,t} =  V_{j,t}^{\#} (X_{j,t} - M_{j,t}), 
\label{eq:B}
\end{equation}	 
be the residual signal, {where the superscript $^{\#}$ denotes the pseudo-inverse of a matrix (for orthonormal $V_{j,t}$, the pseudo-inverse is its transpose).} Denote its $m^{\mbox{th}}$ row as
$B_{j,t}^{(m)}$. Then we can update
\begin{equation}
\lambda_{j, t+1}^{(m)}=
\alpha\lambda_{j, t}^{(m)} + (1-\alpha) \|B_{j,t}^{(m)}\|_2^2, m = 1,
\ldots, r.
\label{eq:Lambda}
\end{equation}
The subspace matrices $V_{j,t}$ are updated using
Algorithm.~\ref{alg:mpetrels}. The updates of $V_{j,t}$ and
  $\Lambda_{j,t}$ are a mini-batch
  extension of the PETRELS \cite{ChiEldarCalderbank2012,ChiJournal2012} update equations, 
  with an added step of orthonormalization of $V_{j,t+1}$ since PETRELS does not guarantee the orthogonality of $V_{j,t+1}$.

For the ancestors of each leaf node that got updated, we
  combine all the mini-batches assigned to its children, and update
  the node with the same formulae as above using the combined
  mini-batches. For the virtual children of leaf nodes that got
  updated, we divide each mini-batch into two sub-mini-batches based
  on the likelihood of each observation under the Gaussian of the
  virtual node, and update each virtual node with its assigned
  sub-mini-batch.  
  
\begin{algorithm}
\caption{Mini-Batch Update of Covariance Parameters}\label{alg:mpetrels}
\begin{algorithmic}[1]
\State {\bf Initialize:} $V_{j,1}$ (with training data), $R_{j,1}=c\bI_{r\times r}, c\ll1$
\State {\bf input:} $X_{j,t}, V_{j,t}, R_{j,t}, M_{j,t}$
\State $B_{j,t} = V_{j,t}^{\#} (X_{j,t} - M_{j,t})$
\State $R_{j,t+1} = \alpha R_{j,t} + B_{j,t}B_{j,t}^T$
\State $\wt{V}_{j,t+1} = {V}_{j,t} + \left( (X_{j,t} - M_{j,t})B_{j,t}^T - V_{j,t}B_{j,t}B_{j,t}^T \right) R_{j,t+1}^{\#}$
\State {\bf Orthonormalization} 
\Statex \quad ${V}_{j,t+1} = \wt{V}_{j,t+1}\left(\wt{V}_{j,t+1}^T\wt{V}_{j,t+1}\right)^{-\frac{1}{2}}$
\State {\bf Output:} $V_{j,t+1}, R_{j,t+1}$
\end{algorithmic}
\end{algorithm}

\subsubsection{Tree structure update}
The growing (splitting nodes) and pruning (merging nodes) of the tree
structure allow the complexity of the GMM to adapt to the diversity of
the observed data.  The number of nodes in the tree controls the
tradeoff between the model accuracy and complexity. {The} proposed
method determines whether to grow or prune the tree by greedily
minimizing a cost function consisting of the weighted cumulative
anomalousness score (with weights corresponding to the forgetting
factor $\alpha$ described above) and the model complexity ($|\mathcal{J}_t|$).

Define $\epsilon_t$ as the cumulative anomalousness score where
  $\epsilon_0 = 0$, and
$$\epsilon_{t+1} = \alpha\epsilon_t + \frac{1}{N_t}\sum_{i=1}^{N_t}s_t(x_{t,i}).$$ 
For each node $j$ (including virtual children), a similar
cumulative score $e_{j,t}$ {is kept} based only on the mini-batches assigned to
that node. Let
$\mathcal{I}_{j,t}\deq\{i:x_{t,i} \mbox{ assigned to } j^{\mbox{th}}
\mbox{ node or its children}\}$
(for virtual nodes this set is the indices of its sub-mini-batch), {initialize} 
$e_{j,0}=0$, and $e_{j,t}$ is updated by
$$
e_{j,t+1} = \alpha e_{j,t} 
	+ \frac{1}{|\mathcal{I}_{j,t}|}
	\sum_{i\in\mathcal{I}_{j,t}}-\log\left(p_{j,t}(x_{t,i})\right).
$$

Let $\tol$ be a pre-set error tolerance. For each leaf node
$j_1\in\cJ_t$ that is assigned new observations, let $j_0$ be its
parent, $j_2$ be its sibling, and $j_{1,1}, j_{1,2}$ be its virtual
children. Let $\gamma$ be a positive constant. Split node
$j_1$ if
\begin{equation}
\epsilon_{t+1} \le \tol,
\label{eq:split1}
\end{equation}
and
\begin{equation}
  e_{j_1,t} + \gamma K_t > \frac{q_{j_{1,1},t}e_{j_{1,1},t} + q_{j_{1,2},t}e_{j_{1,2},t}}{q_{j_{1,1},t}+q_{j_{1,2},t}} + \gamma (K_t+1).
\label{eq:split}
\end{equation}
Note the left side of Ineq.~\eqref{eq:split} is the penalized
cumulative score of node $j_1$ (where the penalty is proportional to
the number of nodes in the tree), while the right side of
Eq.~\eqref{eq:split} is the average penalized cumulative score of node
$j_1$'s two virtual children. We split node $j_1$ if the average
penalized cumulative score is smaller at the virtual children level.

Similarly, merge nodes $j_1$ and $j_2$ if
\begin{equation}
\epsilon_{t+1} \ge \tol
\label{eq:merge1}
\end{equation}
and
\begin{equation}
e_{j_0,t} + \gamma (K_t-1) < \frac{q_{j_1,t}e_{j_1,t} + q_{j_2,t}e_{j_2,t}}{q_{j_1,t}+q_{j_2,t}} + \gamma K_t,
\label{eq:merge}
\end{equation}
Note the left side of Ineq.~\eqref{eq:merge} is the penalized (with
tree size) cumulative score of node $j_1$'s parent $j_0$, while the
right side of Eq.~\eqref{eq:split} is the average penalized cumulative
score of node $j_1$ and its sibling $j_2$. We merge $j_1$ and $j_2$ if
the average penalized cumulative score of $j_1$ and $j_1$ is larger
than the penalized score of their parent. The use of these penalized
scores to choose a tree which is both (a) a good fit to the observed
data and (b) a small as possible to avoid overfitting is common in
classification and regression trees
\cite{CART,willett:tmi03,willett:jsac04,WillettNowak2005,scott2006minimax,willett:density}.
The splitting and merging operations are detailed in
Algorithm~\ref{alg:split} and Algorithm~\ref{alg:merge}. The complete
Online Thinning algorithm is summarized in Algorithm~\ref{alg:OTM}.

\begin{algorithm}[h!]
   \caption{Grow tree}
   \begin{algorithmic}[1]
 \State {\bf Input:} Node $j$ with virtual children nodes $k$ and $\ell$
 \State Update $\cJ_{t+1} = \cJ_{t}\bigcup \{k,\ell\} \backslash \{j\} $
 \State Create new virtual children: $k_1, k_2$ for new leaf node $k$, and $j_{1,1}, j_{1,2}$ for new leaf node $\ell$
 \State Let $v_{i,t}^{(1)}$ be the first column of $V_{i,t}, i\in\{k,\ell\}$
    \State Initialize virtual
    nodes $k_1, k_2, j_{1,1}$ and $j_{1,2}$:
    \Statex \quad for $i\in \{k,\ell\}$
    \begin{align*}
      \mu_{i_1, t+1} &= \mu_{i, t} + \sqrt{\lambda_{i,
          t}^{(1)}} v^{(1)}_{i, t}/2 \\
      \mu_{i_2, t+1} &= \mu_{i, t} - \sqrt{\lambda_{i,
          t}^{(1)}} v^{(1)}_{i,  t}/2\\
      V_{i_1, t+1} &= V_{i, t} \\
      V_{i_2, t+1} &= V_{i, t} \\
      \lambda_{i_1, t+1}^{(1)} & = \lambda_{i, t}^{(1)}/2 \\
      \lambda_{i_2, t+1}^{(1)} &= \lambda_{i, t}^{(1)}/2 \\
      \lambda_{i_1, t+1}^{(m)} &=\lambda_{j, t}^{(m)}, \quad
      m = 2, \ldots, r \\
      \lambda_{i_2, t+1}^{(m)} &= \lambda_{j, t}^{(m)}, \quad
      m = 2, \ldots, r \\
      q_{i_1, t+1} &= q_{j,t} / 2\\
      q_{i_2, t+1} &= q_{j,t} / 2
    \end{align*}
   \end{algorithmic}
   \label{alg:split}
\end{algorithm}

\begin{algorithm}[h!]
  \caption{Prune tree}
  \begin{algorithmic}[1]
 \State {\bf Input}: Node $j$ with children nodes $j_1$ and $j_2$ to be
 merged
 \State Delete all four virtual children nodes of $j_1$ and $j_2$
\State    Update $\cJ_{t+1} = \cJ_t\bigcup\{j\}\backslash \{j_1, j_2\}$
\State Define $j_1$, $j_2$ as the virtual children nodes of the
    new leaf node $j$
  \end{algorithmic}
  \label{alg:merge}
\end{algorithm}

\begin{algorithm}[h!]
  \caption{Online Thinning with Mini-Batch Updates}
  \begin{algorithmic}[1]
    \State {\bf input:}  error tolerance $\tol>0$, threshold $\tau>0$, forgetting factor $\alpha\in(0,1)$
    \State {\bf initialize:} tree structure, set initial error
    $\epsilon_1 = 0$ 
    \For{$t = 1, 2, \ldots$} 
    \State receive new data
    $X_{t}\in\reals^{p\times N_t}$
    \For{$i = 1, 2, \ldots, N_t$} 
    \State let $x_{t,i}$ be the $i^{\mbox{th}}$ column of $X_{t}$
    \State for all $j\in\cJ_t$, compute likelihood of $x_{t,i}$ under
    node $j$:
$$ p_{j,t}(x_{t,i})=\frac{1}{(2\pi)^{p/2}|\Sigma_{j,t}|^{1/2}}e^{-\frac{1}{2}(x_{t,i}-\mu_{j,t})^T{\Sigma}_{j,t}^{-1}(x_{t,i}-\mu_{j,t})}$$
    \State compute anomalousness score $s_t(x_{t,i})$:
$$s_t(x_{t,i}) = -\log \left(\sum_{j\in \cJ_t}q_{j,t}
  p_{j,t}(x_{t,i})\right)$$
\State assign $x_{t,i}$ to leaf node $j^*_{t}\deq\arg\max_{j\in\cJ_t}\{p_{j,t}(x_{t,i})\}.$
\State compute the likelihood of $x_{t,i}$ under $j^*$'s two virtual
children nodes, and also assign $x_{t,i}$ to the virtual child with
higher likelihood
    \EndFor
\State update $\epsilon_{t+1} = \alpha\epsilon_t + \frac{1}{N_t}\sum_{i=1}^{N_t}s_t(x_{t,i})$
    \For{ all nodes $j$ in the tree}
\State set $\mathcal{I}_{j,t}\deq\{i:x_{t,i} \mbox{ assigned to } j^{\mbox{th}}
\mbox{ node or its children}\}$
    \If {$\mathcal{I}_{j,t}$ is not empty}
    \State {denote all data assigned to node $j$ or its children as $
      X_{j,t} = [x_1,\ldots,x_{n_{j,t}}]$}
\State update $e_{j,t+1} = \alpha e_{j,t} 
	+ \frac{1}{|\mathcal{I}_{j,t}|}
	\sum_{i\in\mathcal{I}_{j,t}}-\log\left(p_{j,t}(x_{t,i})\right)$
    \State update $q_{j,t+1} = \alpha q_{j,t} + (1-\alpha)\frac{n_{j,t}}{N_{t}}$
    \State update $\mu_{j,t+1} = \alpha \mu_{j,t} +
    \frac{(1-\alpha)}{n_{j,t}}X_{j,t}\bI_{n_{j,t}\times 1}$
\State set $M_{j,t} = [\mu_{j,t},\ldots,\mu_{j,t}] \in\reals^{p\times n_{j,t}}$
\State set $B_{j,t} =  V_{j,t}^\#   (X_{j,t} - M_{j,t})$
\For{$m = 1,\ldots, r$}
    \State update 
$\lambda_{j, t+1}^{(m)}=
\alpha\lambda_{j, t}^{(m)} + (1-\alpha) \|B_{j,t}^{(m)}\|_2^2$
\EndFor
    \State update $V_{j,t}$ by calling Algorithm~\ref{alg:mpetrels}
\If{$\epsilon_{t+1} \le \tol$ and $e_{j_1,t} + \gamma K_t > \frac{q_{j_{1,1},t}e_{j_{1,1},t} + q_{j_{1,2},t}e_{j_{1,2},t}}{q_{j_{1,1},t}+q_{j_{1,2},t}} + \gamma (K_t+1)$}
    \State call Algorithm~\ref{alg:split} 
\ElsIf{$\epsilon_{t+1} \ge \tol$ and $e_{j_0,t} + \gamma (K_t-1) < \frac{q_{j_1,t}e_{j_1,t} + q_{j_2,t}e_{j_2,t}}{q_{j_1,t}+q_{j_2,t}} + \gamma K_t$}
     \State call Algorithm~\ref{alg:merge}
\EndIf
      \Else { update $q_{j,t+1} = \alpha q_{j,t}$}
      \EndIf
    \EndFor
    \State $\mathcal{X}_t = \{x_{t,i}:s_t(x_{t,i}) > \tau\}$
    \EndFor
    \State {\bf output:} 
sequence of thinned data $\mathcal{X}_1,\ldots,\mathcal{X}_T$
  \end{algorithmic}
  \label{alg:OTM}
\end{algorithm}

\subsection{Subsampling observations}
When $p$ is large and computation time is critical, we can subsample
the elements of each $X_t$ and leverage missing data models for fast
calculations and updates. Algorithm~\ref{alg:mpetrels} is a modified
version of PETRELS \cite{ChiJournal2012,ChiEldarCalderbank2012} with
mini-batches. Note that PETRELS was specifically designed to work with
missing entries, where \cite{ChiJournal2012,ChiEldarCalderbank2012}
thoroughly investigated the effect of missing data in subspace
tracking algorithms.

Specifically, to modify our Online Thinning algorithm for $X_t$ with
subsampled entries, we define $\Omega_t\subseteq \{1,\ldots,p\}$ be
the subset of entries {used} at time $t$. Assume all $\Omega_t$ have
the same size and define $|\Omega|\deq|\Omega_t|, \forall t$.  Define
an operator $P_{\Omega_t}(\cdot)$ that selects the rows indexed by
$\Omega_t$. {Then, for the likelihood and score computation, denote 
$\Sigma_{j,t,\Omega_t} \deq P_{\Omega_t}(V_{j,t}) \Lambda_{j,t} P_{\Omega_t}(V_{j,t})^T + \sigma_j^2 I_{|\Omega_t|}$, and compute  
\begin{equation}
\begin{aligned}
p_{j,t}(x_t)&=\frac{1}{(2\pi)^{p/2}|\Sigma_{j,t}|^{1/2}}\exp\left\{-\frac{1}{2}\left[P_{\Omega_t}(x_{t,i})\right.\right. \\
&\left.\left.-P_{\Omega_t}(\mu_{j,t})\right]^T 
{\Sigma}_{j,t,\Omega_t}^{-1}\left[P_{\Omega_t}(x_{t,i})-P_{\Omega_t}(\mu_{j,t})\right]\right\}
\end{aligned}
\notag
\end{equation}
as the likelihood.
For the mini-batch update step, replace $X_{j,t}, \mu_{j,t},$ and $V_{j,t}$ with
$P_{\Omega_t}(X_{j,t}), P_{\Omega_t}(\mu_{j,t}),$
and $P_{\Omega_t}(V_{j,t})$, respectively in Eq.~\eqref{eq:mu},\eqref{eq:M}, and~\eqref{eq:B}, and
use Algorithm~\ref{alg:mpetrels2} instead of Algorithm~\ref{alg:mpetrels}.}


\begin{algorithm}
\caption{Mini-Batch Update of Covariance Parameters with Subsampling}\label{alg:mpetrels2}
\begin{algorithmic}[1]
\State {\bf Initialize:} $V_{j,1}$ (with training data), $R_{j,1}=c\bI_{r\times r}, c\ll1$
\State {\bf input:} $\Omega_t, P_{\Omega_t}(X_{j,t}), V_{j,t}, R_{j,t}, P_{\Omega_t}(M_{j,t})$
\State $B_{j,t} = P_{\Omega_t}(V_{j,t})^{\#} \left[P_{\Omega_t}(X_{j,t}) - P_{\Omega_t}(M_{j,t})\right]$
\State $R_{j,t+1} = \alpha R_{j,t} + B_{j,t}B_{j,t}^T$
\State $\wt{V}_{j,t+1} = {V}_{j,t}$
\State $P_{\Omega_t}(\wt{V}_{j,t+1}) = P_{\Omega_t}(\wt{V}_{j,t+1}) + \left[ (P_{\Omega_t}(X_{j,t})\right.$
\Statex $\left. - P_{\Omega_t}(M_{j,t}))B_{j,t}^T - P_{\Omega_t}(V_{j,t})B_{j,t}B_{j,t}^T \right] R_{j,t+1}^{\#}$
\State {\bf Orthonormalization} 
\Statex \quad ${V}_{j,t+1} = \wt{V}_{j,t+1}\left(\wt{V}_{j,t+1}^T\wt{V}_{j,t+1}\right)^{-\frac{1}{2}}$
\State {\bf Output:} $V_{j,t+1}, R_{j,t+1}$
\end{algorithmic}
\end{algorithm}

\subsection{Computational complexity}
\label{sec:complexity}
As discussed in Section~\ref{sec:tracking}, the union of subspaces
assumption significantly reduces the problem size for estimating the
covariance matrices. This not only improves the algorithm accuracy and
stability for high dimensional problems, but also reduces computation
time. 

{
Take the computation of $(x-\mu_{j,t})^T\Sigma_{j,t}^{-1}(x-\mu_{j,t})$ as an example. 
$\Sigma_{j,t}$ is the covariance matrix of the
$j^{\mbox{th}}$ mixture component at time $t$. For a full-rank GMM
model, computing the $\Sigma_{j,t}^{-1}$ takes $O(p^3)$
operations, and computing $(x-\mu_{j,t})^T\Sigma_{j,t}^{-1}(x-\mu_{j,t})$ given $\Sigma_{j,t}^{-1}$ takes $O(p^2)$ operations. Thus the total complexity is $O(p^3)$.
However, with the low-rank assumption we have
$\Sigma_{j,t}= V_{j,t}\Lambda_{j,t}V_{j,t}^T + \sigma^2 I$, and with
the Woodbury matrix identity \cite{woodbury1950inverting}, we can compute
\begin{equation}
\Sigma_{j,t}^{-1} = \sigma^{-2} I + \sigma^{-4} V_{j,t} (\Lambda_{j,t}^{-1} + V_{j,t}^T V_{j,t})^{-1} V_{j,t}^T,
\notag
\end{equation} 
and
\begin{equation}
\begin{aligned}
&(x-\mu_{j,t})^T\Sigma_{j,t}^{-1}(x-\mu_{j,t}) = \\ 
&\sigma^{-2} (x-\mu_{j,t})^T(x-\mu_{j,t}) \\
&+ \sigma^{-4} (x-\mu_{j,t})^TV_{j,t} (\Lambda_{j,t}^{-1} + V_{j,t}^T V_{j,t})^{-1} V_{j,t}^T(x-\mu_{j,t}).
\end{aligned}
\notag
\end{equation}
Note that computing $(\Lambda_{j,t}^{-1} + V_{j,t}^T V_{j,t})^{-1}$ is easy because (a) $V_{j,t}^T V_{j,t} = I_{r}$ since the columns of $V_{j,t}$ are orthonormal, and (b) $\Lambda_{j,t}$ is diagonal.
Computing the whole equation takes $O(pr+ r^2)=O(pr)$ operations. Thus, by using the low-dimensional structure, we reduced the computation complexity from $O(p^3)$ to $O(pr)$.
}

{
Another example is the computation of the determinant of $\Sigma_{j,t}$. For a full-rank GMM
model, computing $|\Sigma_{j,t}|$ takes $O(p^3)$ operations. For our low-rank model with $\Sigma_{j,t}= V_{j,t}\Lambda_{j,t}V_{j,t}^T + \sigma^2 I$, we can use the matrix determinant lemma \cite{harville1998matrix} and compute
\begin{equation}
\begin{aligned}
|\Sigma_{j,t}| &= \sigma^2\left|\Lambda_{j,t}\right| \left|\Lambda_{j,t}^{-1}+\sigma^{-2}V_{j,t}^TV_{j,t}\right|.
\end{aligned}
\notag
\end{equation}
The number of operation needed is O(r) since $V_{j,t}^T V_{j,t} = I_{r}$ and $\Lambda_{j,t}$ is diagonal.
}

{
\subsubsection{Computational complexity without subsampling}
{
Below we summarize the {computational} complexity of each major steps of the Online Thinning algorithm.{Let} $T$ {be} the total number of time steps. {For} simplicity, we assume the
mini-batch sizes $N_t$ are the same for all $t$. Let $N \deq T N_t$ be
the total number of observations {received}, and $\Kmax$ be the
maximum number of leaf nodes in the tree. 
}
\begin{itemize}
\item {Likelihood and score computation} has a complexity of  
$$O(N\Kmax pr),$$
where each likelihood computation takes $O(pr)$ operations, and this is computed $\Kmax+2$ times per observation ($\Kmax$ leaf nodes plus two virtual children of the assigned leaf node) for all $N$ observations. The computation of the anomalousness score is computationally inexpensive since it is a weighted sum of pre-computed likelihoods.
\item Mini-batch updates {have complexities of} at most  
$$O\left(N \Kmax pr + \frac{N}{N_t} \Kmax pr^2\right),$$
where the first term $N \Kmax pr$ comes from the calculation of $B_{j,t}$ and $\wt{V}_{j,t+1}$, and the second term comes from the orthonormalization of $\wt{V}_{j,t+1}$. The term $\frac{N}{N_t}$ is the number of batches {received}.
\item Tree structure updates {have complexities of} at most $$O(N \Kmax pr+\frac{N}{N_t}\Kmax pr),$$
where the first term is an upper bound of complexity for updating the cumulative likelihood $e_{j,t}$ of the parents of leaf nodes (for leaf nodes and their virtual children, the likelihood is computed when at the score computing stage). In the second term, the term $pr$ comes from the number of operations needed to copy the subspace when splitting nodes, and the maximum number of splitting at each time $t$ is bounded by the maximum number of leaf nodes $\Kmax$ (merging nodes is computational inexpensive).  
\end{itemize}
Adding three steps together, the Online Thinning algorithm has a complexity {of} at most
$$O\left(N \Kmax pr + \frac{N}{N_t} \Kmax pr^2\right).$$
}

{
\subsubsection{Computational complexity with subsampling}
Let $|\Omega|$ be the number of entries {observed} after subsampling, then the complexity of each major step {is} as follows.
\begin{itemize}
\item {Likelihood and score computation} has a complexity of
$$O(N\Kmax |\Omega|r),$$
which scales with $|\Omega|$.
\item Mini-batch updates {have complexities of} at most
$$O\left(N \Kmax (|\Omega|r+r^2) + \frac{N}{N_t} \Kmax pr^2\right),$$
where the first term $N \Kmax (|\Omega|r+r^2)$ comes from the calculation of $B_{j,t}$ and $\wt{V}_{j,t+1}$, and is affected by subsampling. Note the extra $N\Kmax r^2$ comes from the added complexity from computing $P_{\Omega_t}(V_{j,t})^{\#}$. When no subsampling is done, $V_{j,t}^{\#}=V_{j,t}^{T}$. However, this is {generally} not true when we perform subsampling, and this extra step adds complexity. However, in general $|\Omega|$ scales with $p$ and is much larger than $r$.
The second term comes from the orthonormalization of $\wt{V}_{j,t+1}$, which is not affected by subsampling.
\item Tree structure updates {have complexities of} at most  
$$O(N \Kmax |\Omega|r+\frac{N}{N_t}\Kmax pr),$$
where the first term comes from the likelihood computation, and scales with $|\Omega|$. The second comes from the tree splitting, which is not affected by subsampling.
\end{itemize}
Adding three steps together, the Online Thinning algorithm has a complexity {of} at most
$$O\left(N \Kmax (|\Omega|r + r^2) + \frac{N}{N_t} \Kmax pr^2\right).$$
}

\subsubsection{Remarks}
{
Subsampling changes the first term of {the} complexity. Higher subsampling rates (smaller $|\Omega|$) 
{reduce} the complexity of computing the likelihood, and affect some steps in the mini-batch update. However, {subsampling} does not affect the orthonormalization of $V_{j,t}$, or the splitting and merging of tree structures. Additionally, subsampling makes the computation of $P_{\Omega_t}(V_{j,t})^{\#}$ difficult, since in general $P_{\Omega_t}(V_{j,t})^{\#}\neq P_{\Omega_t}(V_{j,t})^{T}$. Still, the effect of this added complexity in computing $P_{\Omega_t}(V_{j,t})^{\#}$ is generally small since $|\Omega|$ is usually much larger than $r$.
}

{
Changing the size of mini-batches
$N_t$ changes $\frac{N}{N_t}$, and thus the second term of {the} complexity. Specifically, in the algorithm,
changing $N_t$ changes (a) the number of times {needed} to update the tree
structure, (b) the number of times {needed to} update $R_{j,t}$ and its pseudo-inverse (see
Algorithm~\ref{alg:mpetrels}), and (c) the
number of time {needed} to perform the orthonormalization of $V_{j,t}$
(see Algorithm~\ref{alg:mpetrels}).
When $N_t$ has the same order of (or larger than) $\Kmax r$, and without subsampling, the algorithm's complexity is linear to the total number of observations $N$, the observation dimension $p$, the tree size $\Kmax$, and the subspace dimension $r$.
}

\section{Computational and statistical tradeoffs}
\label{sec:computation}

Different systems have different delay allowances and precision
requirements, and it is natural to ask how much performance we
sacrifice by trying to reduce the computation time. Understanding such
tradeoffs is crucial for applications where real-time processing is
required, yet the computational power is limited, as with many mobile 
surveillance systems. {This section explores} the tradeoff between
processing time and detection accuracy for the Online Thinning
algorithm. 

There are two primary ways to reduce the computational complexity of
the data thinning: {(1)} by randomly subsampling the entries of $x_t$,
\ie we only use partially observed data to update the dynamic low-rank GMM
model parameters and
estimate the anomalousness score; and {(2)} by varying the size
  of the mini-batches.  Note that these are made possible because, as
discussed in Section~\ref{sec:tracking}, data thinning (a) is robust
to unobserved entries, and (b) can process data in mini-batches,
respectively.

To explore this further, two experiments {are conducted}---one in which we
vary the mini-batch size, and one in which we vary the subsampling
rate. For these experiments, {the data is generated} as follows: The ambient
dimension is $p=100$. We first generate points in $\reals^p$ in a
union of three (shifted) subspaces of dimension ten; in which 95\% of
the points lie in the union of the first two subspaces. The other 5\%
of the points lie in a third subspace that is orthogonal to the other
two. All three subspaces have shifts close to 0. We then add white
Gaussian noise with variance $\sigma^2 = 0.1$ to these points to
generate our observations.  The two subspaces where the 95\% of
observations come from are dynamic, where {the subspaces rotate at a}
speed $\delta>0$. For $j=1,2$, we have
$$
V_{j,t+1} =  V_{j,t} + \delta\frac{B}{\|B\|_F}V_{j,t},
$$
where $B$ is a $p\times p$ skew-symmetric matrix.  Denote the set of
$x_t$'s coming from each of the three subspaces as
$\mathcal{X}_j, j=1,2,3$, respectively.  The goal is to identify the
5\% of the observations that come from $\mathcal{X}_3$.

The experiment streams in four thousand observations in total.  An
initial model is estimated using the first one thousand samples, and
the models are then updated in an online fashion for the remaining
three thousand samples. The anomalousness score is calculated as the
negative log-likelihood of each data point according to the estimated
model. We then select observations $x_t$ for which $s_t(x_t) > \tau$,
and compute the detection rate and false alarm rate
$$P_D(\tau) = \frac{|\{t:x_t \in\mathcal{X}_3, s_t(x_t) > \tau\}|}{|\{t:x_t \in \mathcal{X}_3\}|},$$
$$P_F(\tau) = \frac{|\{t:x_t \in\mathcal{X}_1\bigcup\mathcal{X}_2, s_t(x_t) > \tau\}|}{|\{t:x_t \in \mathcal{X}_1\bigcup\mathcal{X}_2\}|}.$$
The threshold $\tau$ is tuned to minimize the detection error
$1-P_D(\tau)+P_F(\tau)$. Each experiment is averaged over ten random
realizations.

 {The first experiment varies} the percentage of entries {observed} in
each $x_t$. Subsampling reduces the dimension of $x_t$,
which saves time in many of the operations in the algorithm. With more
observed entries, {the} estimates of the likelihoods
under each mixture component {are more accurate}, and hence the thinning
performance {is better}. However, the computation of likelihoods and updates of
the dynamic low-rank GMM parameters will also be slower.

\begin{figure}[!t]
\centering
{\includegraphics[width=0.5\textwidth]{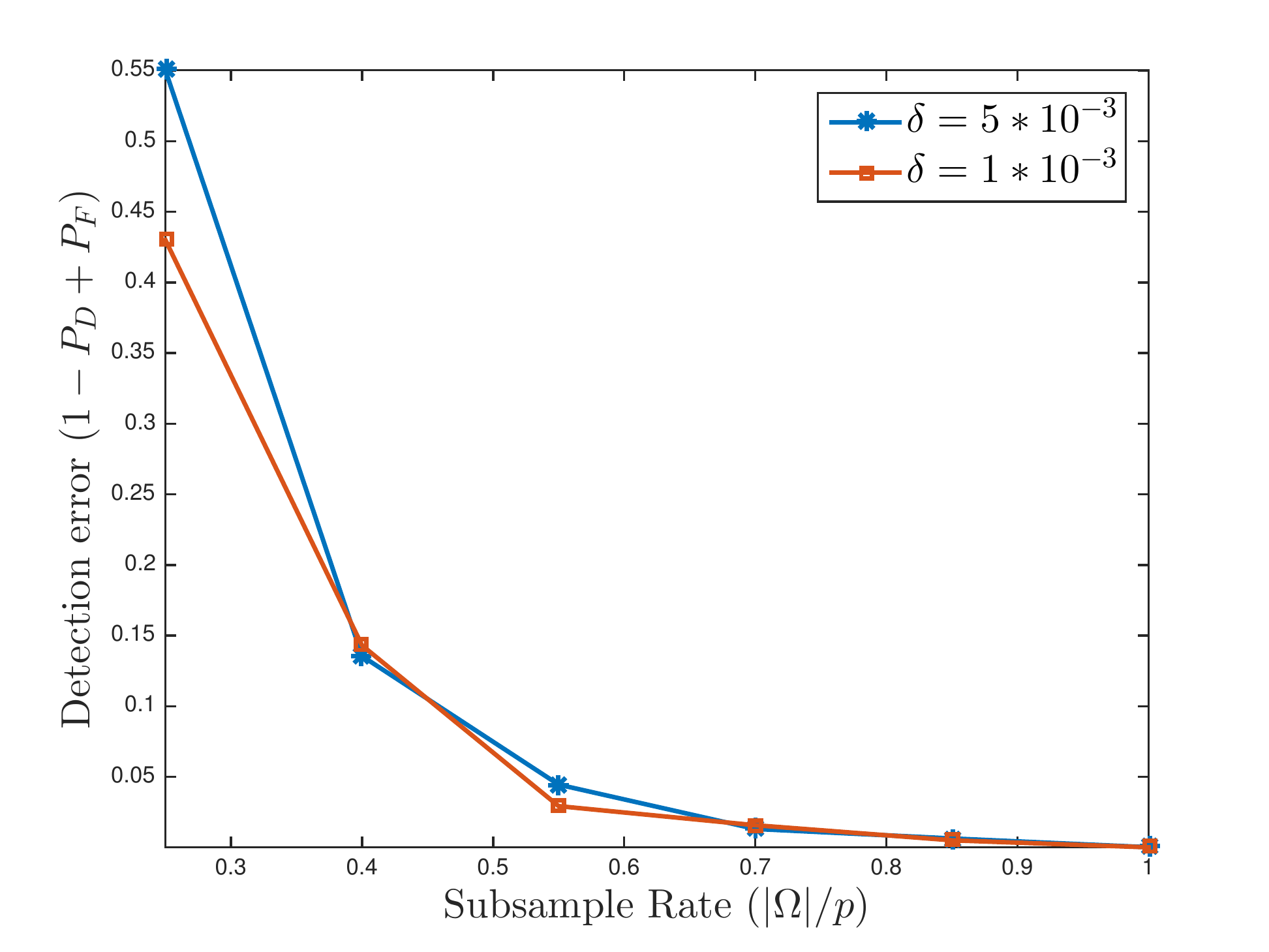}}
\caption{Detection error as a function of subsampling rate. The
    two curves correspond to different subspace rotation speed
    ($\delta$). A subsampling rate at 55\% still keeps the detection
    error less than 5\%. }
\label{fig:timeaccuracy1}
\end{figure}

Fig.~\ref{fig:timeaccuracy1} shows the detection error of our approach
as a function of subsampling rate ($|\Omega|/p$). The two curves
correspond to different subspace rotation speed ($\delta$). We vary
the subsampling rate from 25\% to 100\%.  The detection error is kept
at less than 5\% even at a subsampling rate of 55\%.

\begin{figure}[!t]
\centering
{\includegraphics[width=0.5\textwidth]{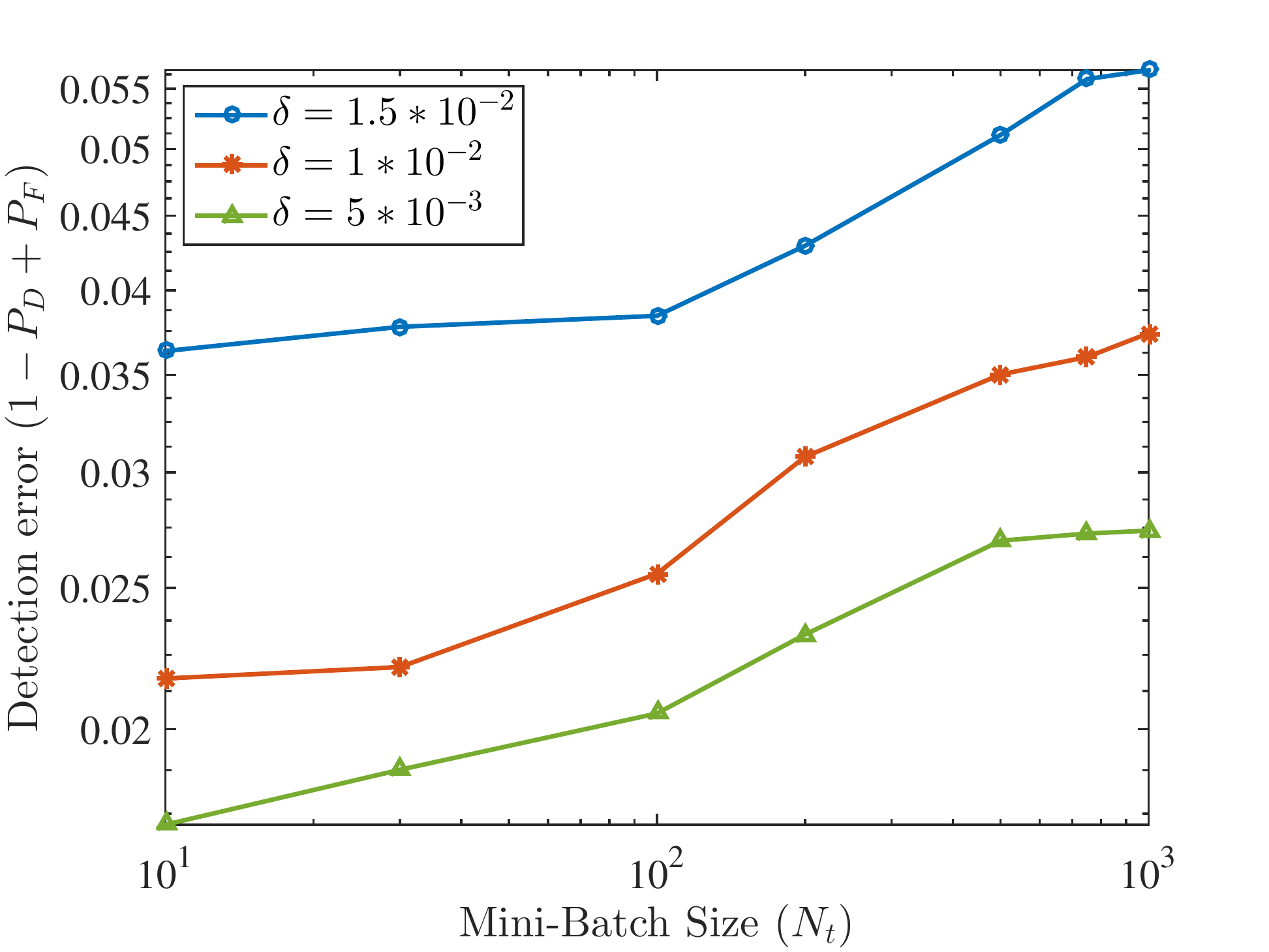}}
\caption{Detection error as a function of mini-batch size $N_t$. The three curves correspond to different
  subspace changing speeds ($\delta$). The detection error increases as
  $N_t$ increases. For all three $\delta$ values, the change in
  detection error relative to $N_t$ is less than 2\%.}
\label{fig:timeaccuracy2}
\end{figure}

{The second experiment varies} $N_t$, the size of the mini-batches.
{The batch size $N_t$ varies} from 10
to 1000.  Fig.~\ref{fig:timeaccuracy2} displays the detection error as
a function of $N_t$. The three curves correspond to different subspace
rotation speed ($\delta$). The detection error increases slightly as
$N_t$ increase, since reducing $N_t$ in general improves the ability
of the algorithm to follow the changing subspaces. For all three
values of $\delta$, the change in detection error relative to $N_t$ is
less than 2\%.

\section{Synthetic data experiments}
\label{sec:syncexp}
{This section compares the} Online Thinning approach based on
tracking a dynamic low-rank GMM with (a) a classical full-rank static GMM and
(b) an online GMM estimation algorithm. Neither
of these comparators has the low-rank structure {exploited by the Online Thinning algorithm}. The
synthetic data is generated according to the same model as in Section~\ref{sec:computation}.

\begin{figure*}[!t]
\centering
\subfloat[Static subspaces]{\includegraphics[width=0.33\textwidth]{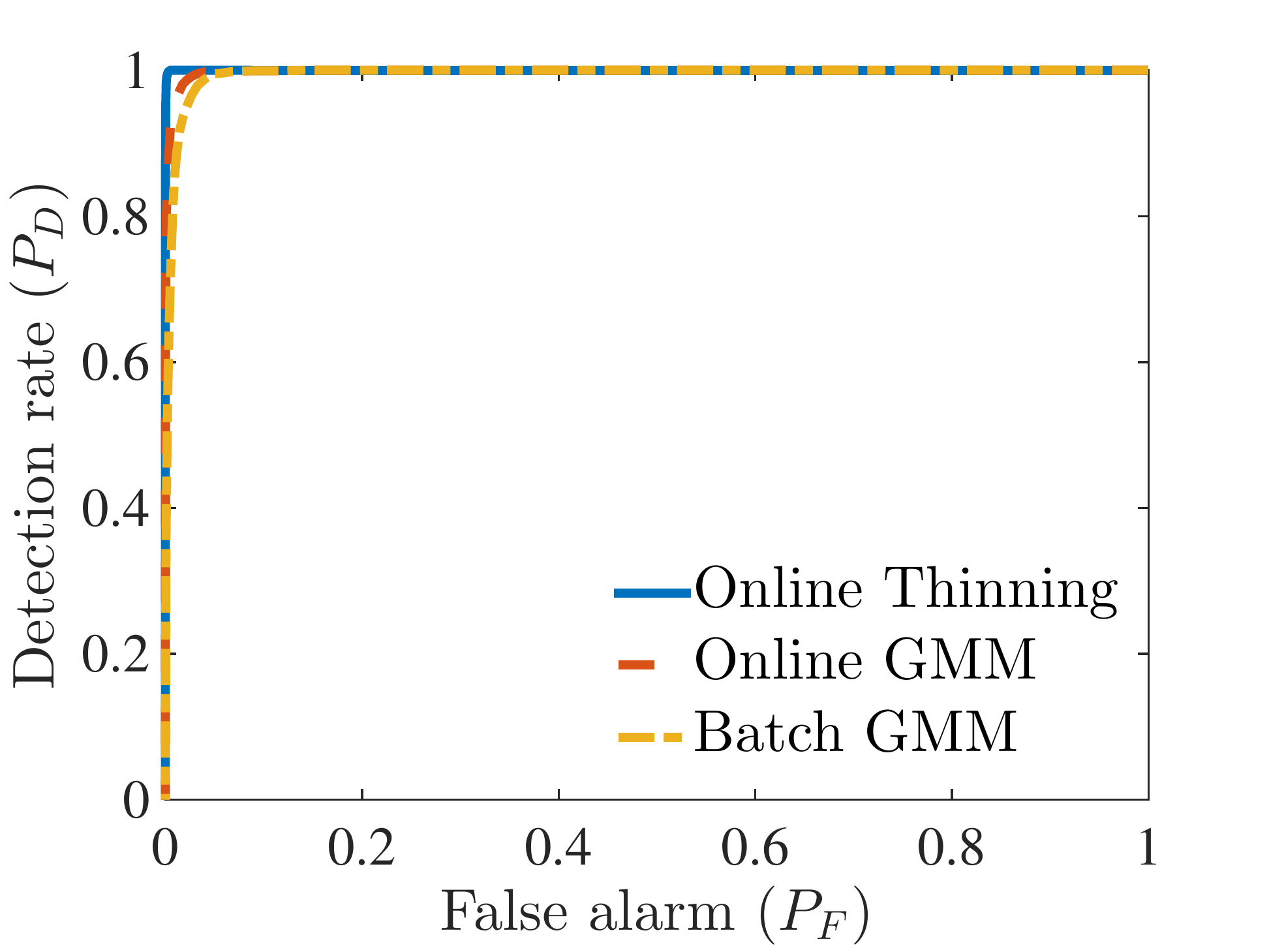}\label{fig:sync1}}~
\subfloat[Subspaces changing at rate $5\times10^{-3}$]{\includegraphics[width=0.33\textwidth]{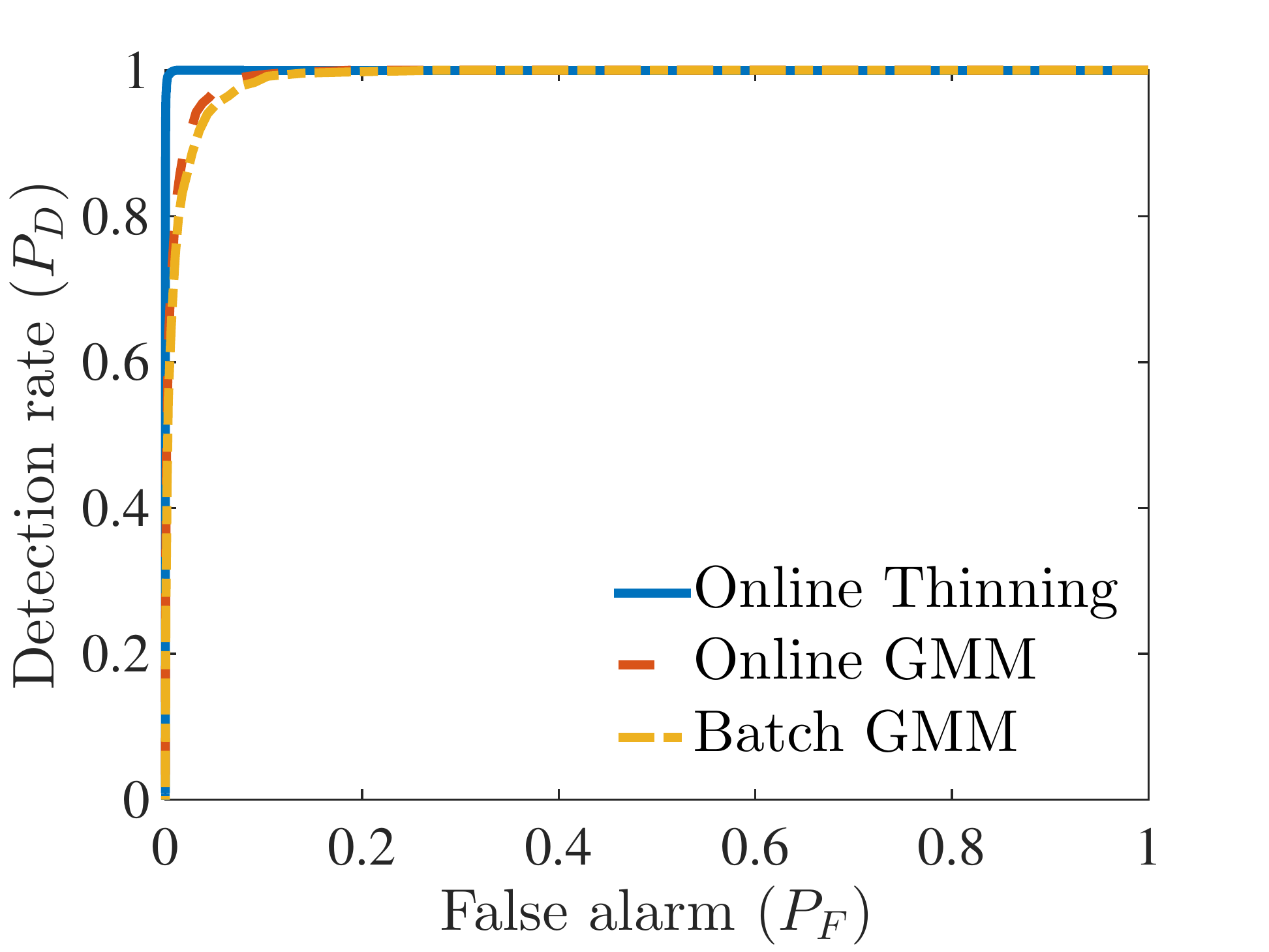}\label{fig:sync2}}~
\subfloat[Subspaces changing at rate $2\times10^{-2}$]{\includegraphics[width=0.33\textwidth]{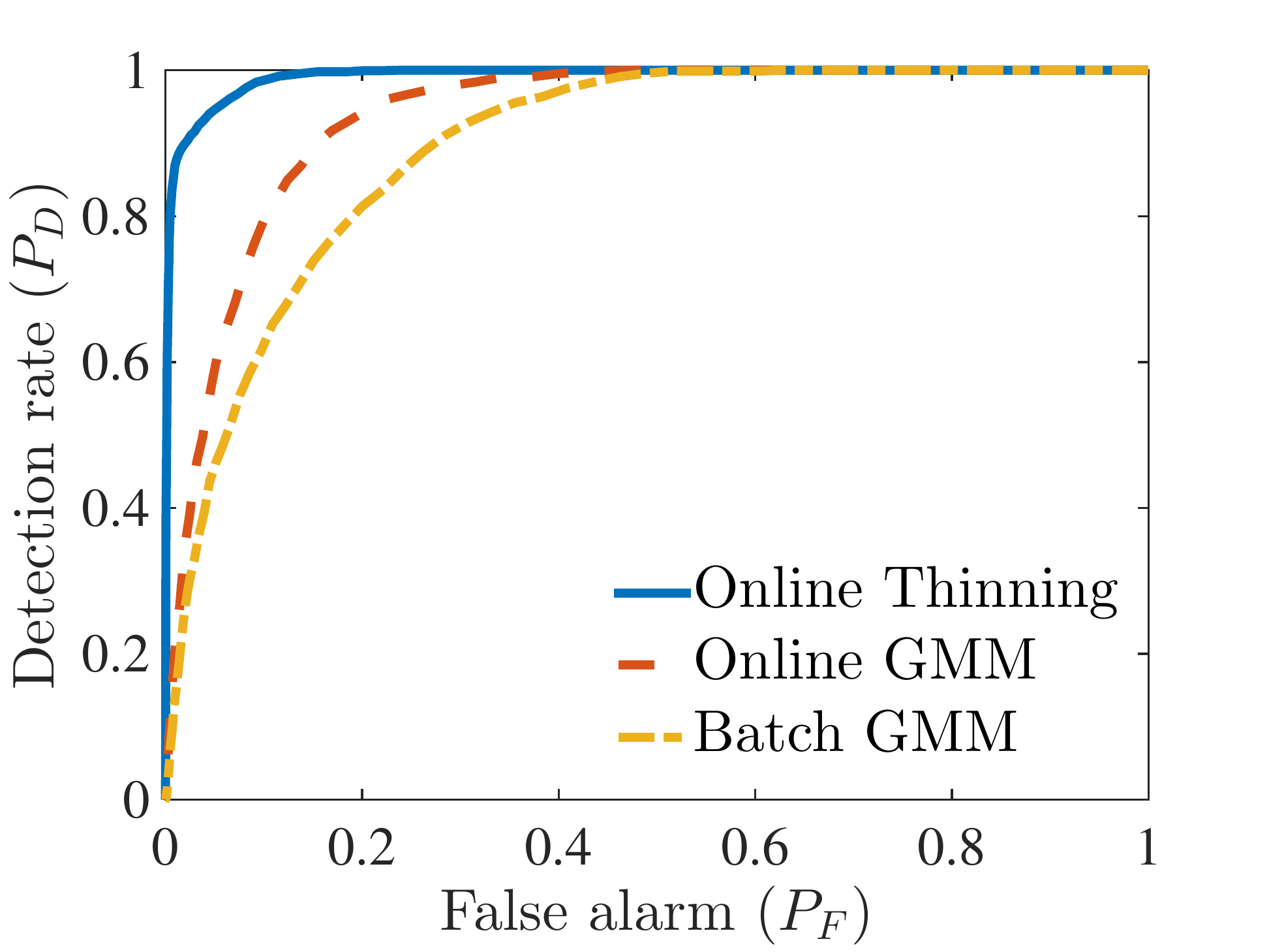}\label{fig:sync3}}
\caption{Comparison between Online Thinning using a dynamic low-rank
  GMM, a classical online GMM, and a classical static batch GMM assuming
  true subspace rank (ten) is {known}. \ref{fig:sync1}
  shows the comparison between Online Thinning and GMM when the subspaces
  are static. \ref{fig:sync2} and \ref{fig:sync3} show the comparison
  between Online Thinning and GMM when the subspaces change at a rate of
  $5\times10^{-3}$ and $2\times10^{-2}$, respectively. Online Thinning
  outperforms both the online and batch GMM algorithms in all cases,
  especially when the subspaces change over time.  }
\label{fig:sync}
\end{figure*}

The experiment streams in four thousand observations in total.  For
Online Thinning and the classical online GMM, an initial model is
estimated using the first one thousand samples, and the models are
then updated in an online fashion for the remaining three thousand
samples. The anomalousness score is calculated as the negative
log-likelihood of each data point according to the estimated
model. For the classical static GMM algorithm, we estimate a GMM model
on the entire four thousand data points (after all samples come in) at
once, and assign an anomalousness score to each sample proportional to
the negative log likelihood of the data point coming from the
estimated GMM model. Fig.~\ref{fig:sync} compares the detection
accuracy (in ROC curves) of Online Thinning and the two comparator
algorithms in two settings, where in \ref{fig:sync1}, the true
subspaces used to generate the data are kept static throughout the
experiment, and in \ref{fig:sync2} and \ref{fig:sync3}, the true
subspaces rotate at a small rate ($5\times10^{-3}$ and
$2\times10^{-2}$, respectively) at each time step. Each plotted
experiment is averaged over thirty random realizations. As seen in the
plots, Online Thinning using the dynamic low-rank GMM outperforms the
online and static algorithms based on a classical full-rank GMM model
in all cases, especially when the subspaces change over time.

The reasons behind the performance gap when the subspaces
change over time can be explained by the underlying models of the three 
algorithms. Both the batch GMM and online GMM algorithms rely on full-rank
GMM models, which make the problem ill-posed, and, {therefore,} estimating the
covariance matrices becomes difficult. Furthermore,  the batch GMM algorithm
relies on a static model, which introduces bias when the environment is dynamic. 
On the other hand, Online Thinning is based on a
dynamic low-rank GMM model, and thus faces a much less ill-posed problem
by having a union of subspace assumption (which significantly reduces
the number of unknowns in the covariance matrices). At the same time,
Online Thinning focuses on the most recent samples by weighing down
the past samples, and can thus follow the changes in the subspaces. 

\begin{figure}[!t]
\centering
{\includegraphics[width=0.5\textwidth]{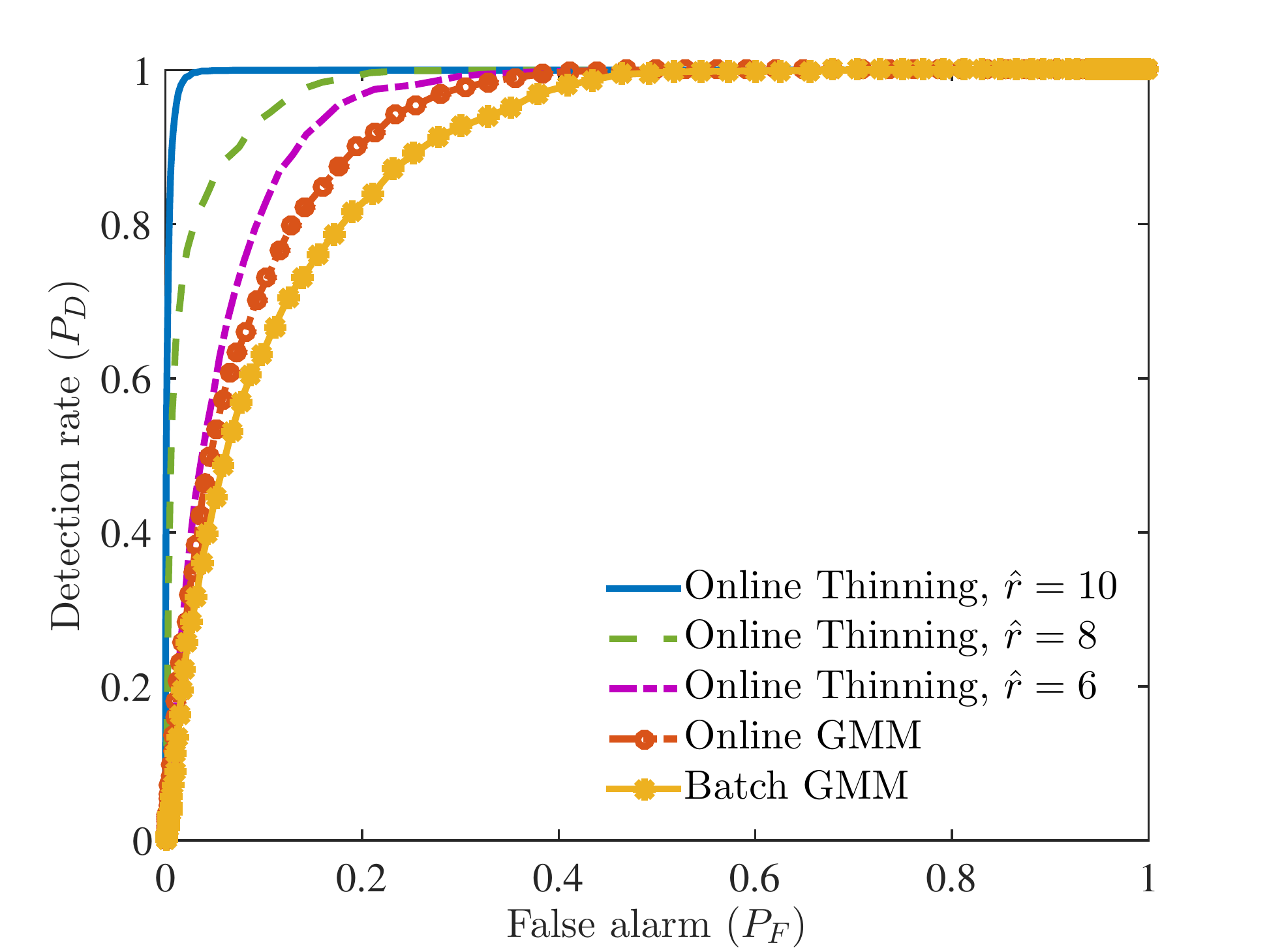}}
\caption{Comparison between Online Thinning, online GMM and batch GMM assuming the subspace rank {is under-estimated} at six and eight for the Online Thinning algorithm. We show the comparison between Online Thinning, online GMM and regular GMM algorithms when the subspaces change at a rate of $1\times10^{-2}$. Even when {the rank is incorrectly estimated at six (correct rank is ten)}, Online Thinning outperforms both classical batch GMM and online GMM algorithms.}
\label{fig:sync_r7}
\end{figure}

In Fig.~\ref{fig:sync}, the true subspace
rank {is assumed known}. However, the real rank of the subspaces is not always known {\em a
priori}. To further assess the performance of Online Thinning in such
situations, we repeat the above experiment but compute rank-six and rank-eight
approximations of the rank-ten subspaces; the results are displayed in
Fig.~\ref{fig:sync_r7}. Note that the classical (full-rank) GMM
algorithms are not affected by the rank assumption. As seen in the
plots, the performance of Online Thinning
slightly degrades when the rank of the subspace is given incorrectly
to the algorithm. However, Online Thinning still outperforms the classical batch GMM and online GMM
algorithms when the subspaces rotates at a rate of $\delta=1\times 10^{-2}$, even when the rank of the subspaces {is significantly under-estimated}.

\section{Wide-area motion imagery experiments}
\label{sec:WAMIexp}

{This experiment compares} Online Thinning with the SUN (Saliency
Using Natural statistics) algorithm proposed by Zhang {\em et al}
in~\cite{zhang2008sun}.  The SUN algorithm is representative of the
state-of-the-art saliency detection algorithms~\cite{borji2013state},
provides a general framework for many models, performs as well as or
better than previous models, and is computationally
efficient~\cite{zhang2008sun}.

We perform this comparison on a real surveillance video capturing an
empty field near a highway. In the video, a car is parked on the lot,
and two people can be seen walking in and out of the scene on the
field. We use this video because it is clear that the car and the
people are most salient in the scene. The original video can be found
at \url{https://youtu.be/mX1TtGdGFMU}.  For the Online Thinning
algorithm, we use SIFT (scale-invariant feature transform) features
\cite{lowe1999object} of frame $t$ as our observation $X_t$ at time
$t$.  Specifically, we use the package from \cite{vedaldi08vlfeat} to
compute the dense SIFT features (\ie SIFT features computed over a
pre-set grid of points on each frame) as features. Each frame of the video is of size
$960\times540$, and the {grid is placed} so that one SIFT feature is
computed for each $25\times 25$ patch.  {Each frame have} roughly
eight hundred SIFT feature vectors.  The dimension of each
SIFT feature vector is 128. 
  
\begin{figure}[!t]
\centering
\subfloat[Original, frame 50]{\includegraphics[width=0.4\textwidth]{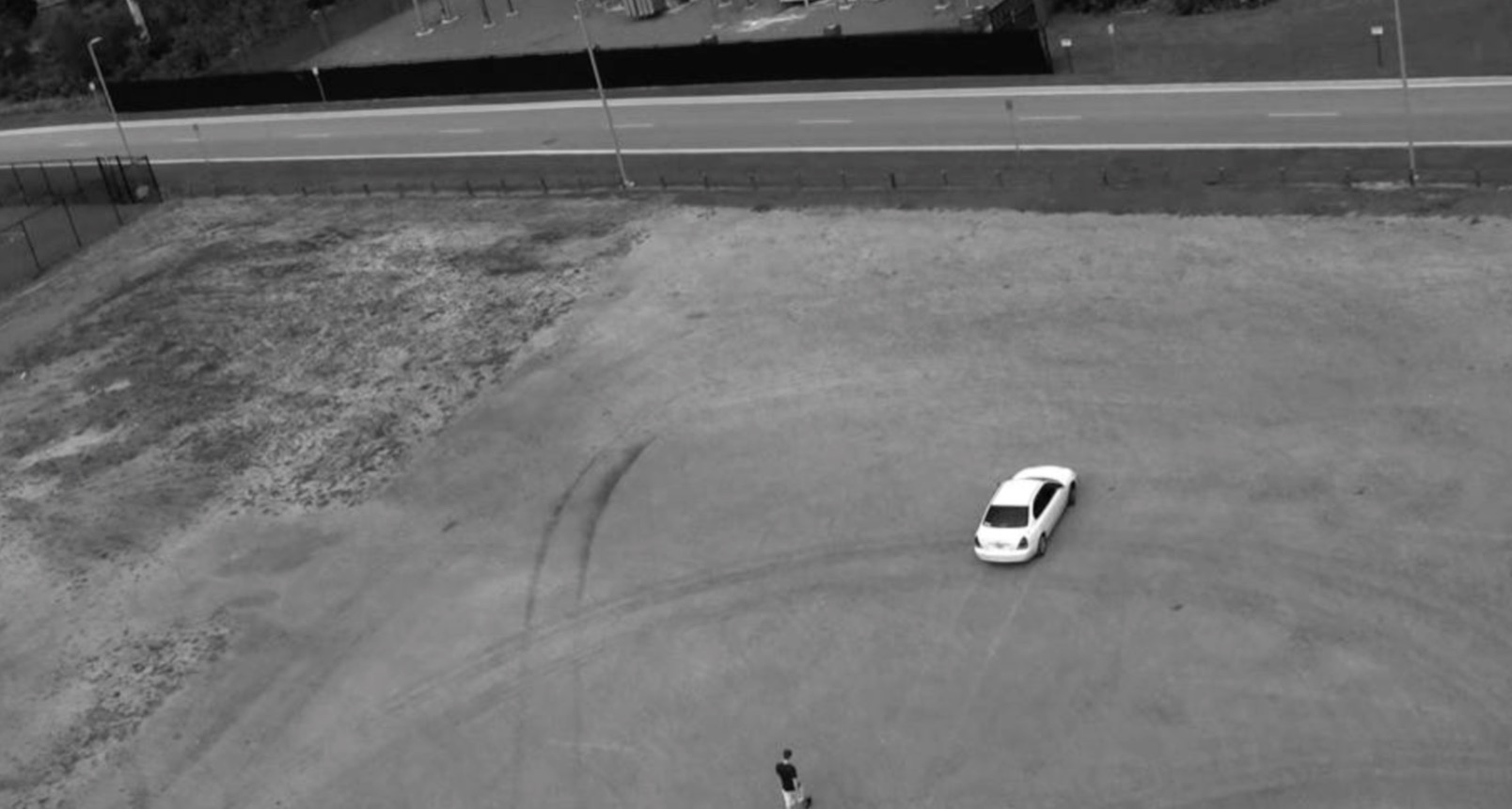}\label{fig:pklot11}}~
\subfloat[Original, frame 100]{\includegraphics[width=0.4\textwidth]{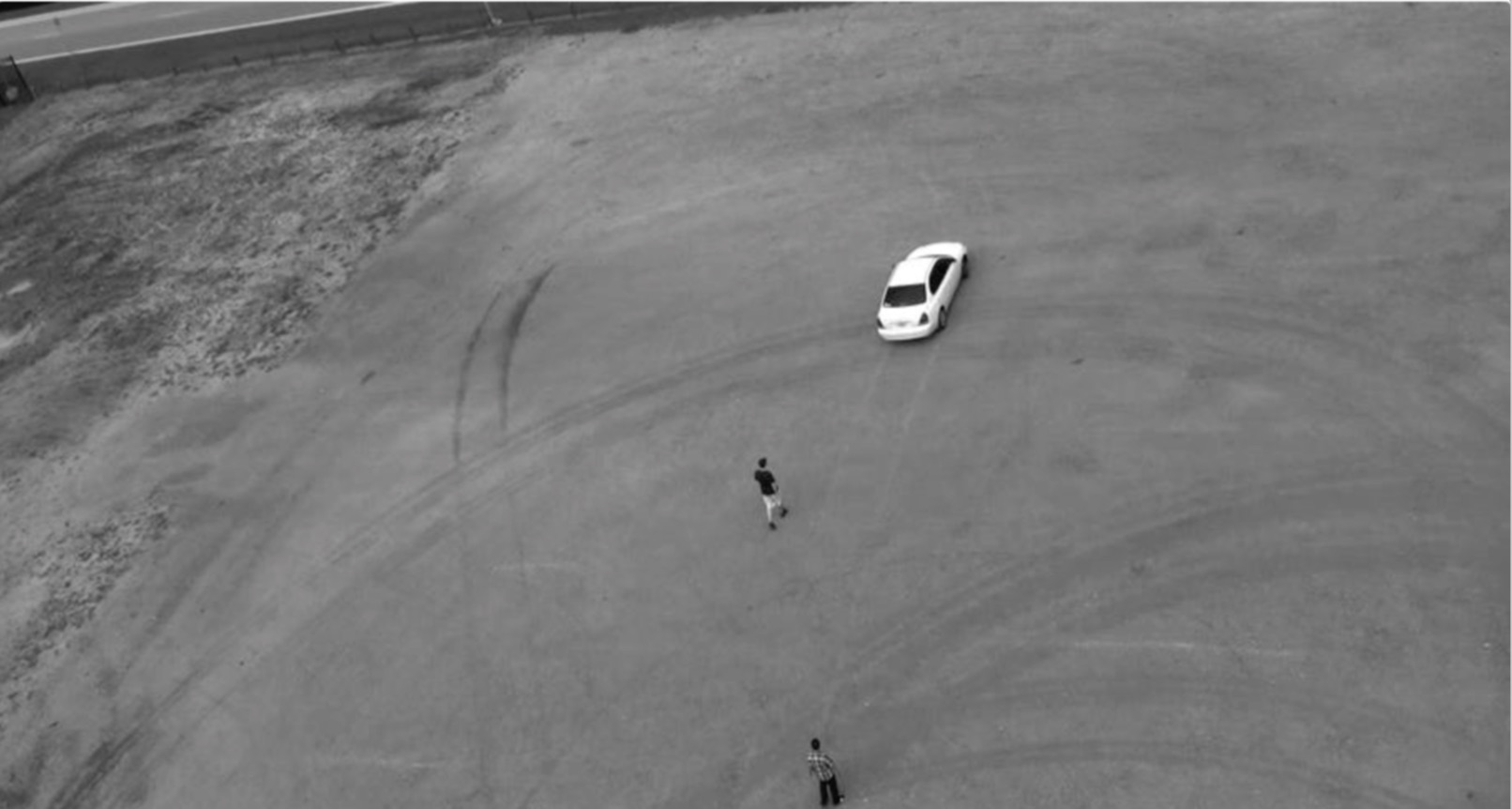}\label{fig:pklot12}}\\
\subfloat[5\% most salient patches, frame 50]{\includegraphics[width=0.4\textwidth]{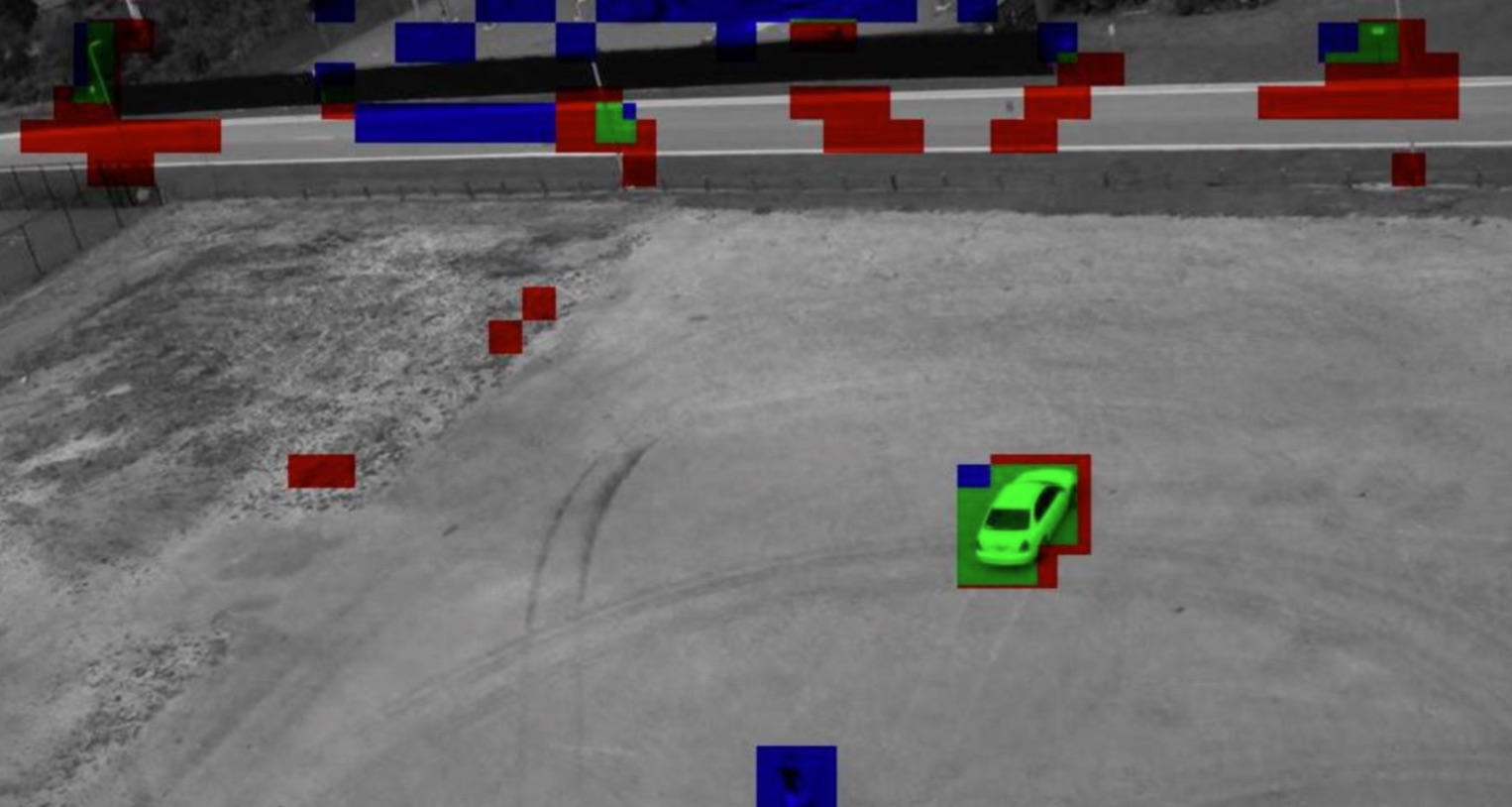}\label{fig:pklot21}}~
\subfloat[5\% most salient patches, frame 100]{\includegraphics[width=0.4\textwidth]{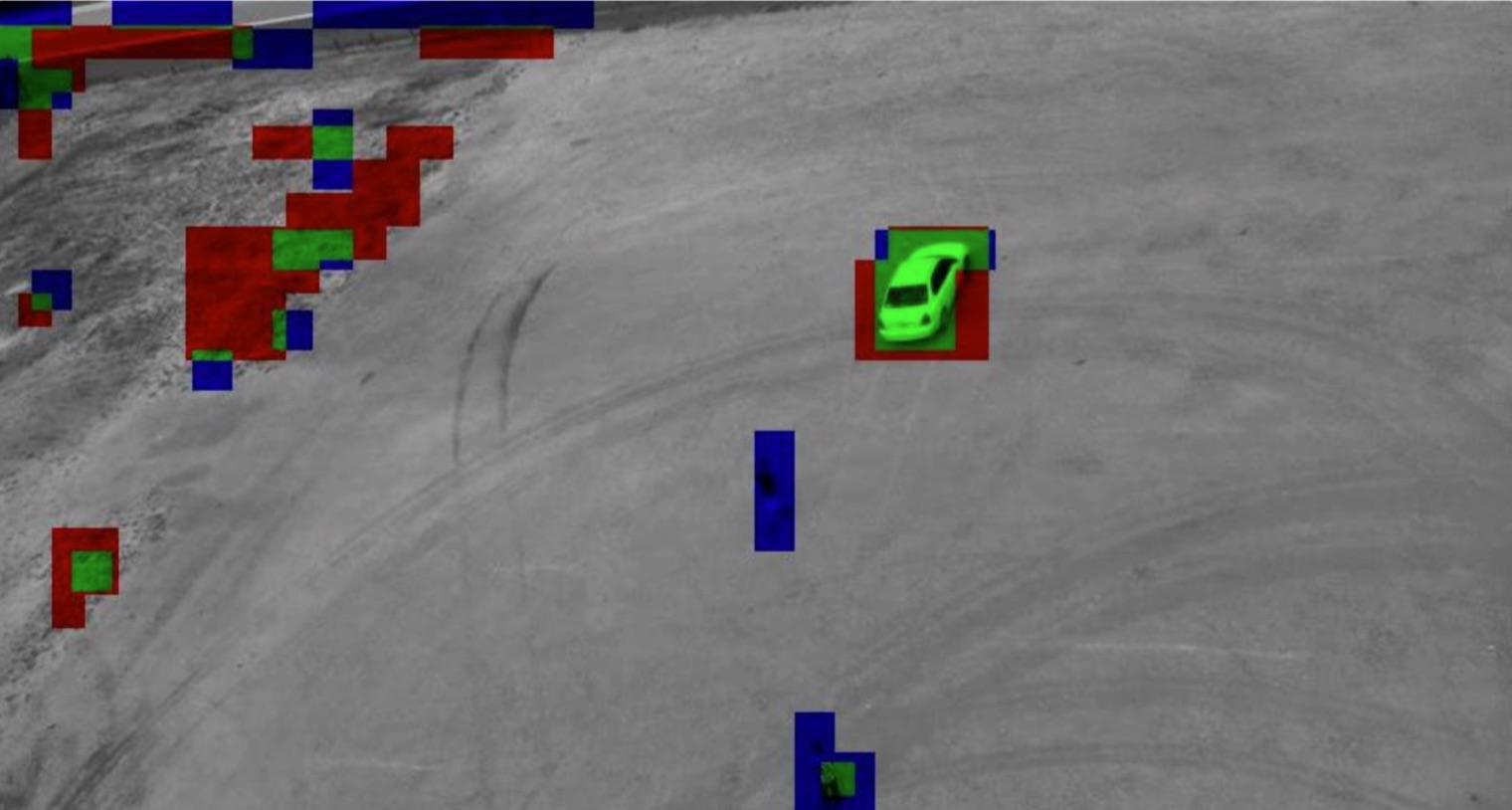}\label{fig:pklot22}}
\caption{Data thinning result using Online Thinning and SUN algorithms on the surveillance video at frames 50 and 100. The first row shows the original video, and the second row shows the data thinning results. In the results, green patches are flagged by both methods, blue patches are only flagged by Online Thinning, and red patches are only flagged by SUN. Online Thinning outperforms the SUN algorithm by consistently flagging the people, which are sometimes missed by the SUN algorithm.}
\label{fig:pklot}
\end{figure}

Fig.~\ref{fig:pklot} shows the result of Online Thinning and the SUN
algorithms on this surveillance video at frames 50 and
100. Figures~\ref{fig:pklot11} and~\ref{fig:pklot12} show the original
frames, while in~\ref{fig:pklot21} and~\ref{fig:pklot22}, we flag the
top 5\% patches with the highest anomalousness or saliency scores by
the Online Thinning and SUN algorithms. In the results, green patches
are flagged by both methods, blue patches are only flagged by Online
Thinning, and red patches are only flagged by SUN. Note that in both
frames, the people in the scene are mostly labeled by blue, \ie they
are only flagged by Online Thinning. {The} Online Thinning
outperforms the SUN algorithm by more consistently flagging small rare
patches such as the people; this is in part due to the adaptivity of
{Online Thinning} to dynamic environments. The result video can be found at
\url{https://www.youtube.com/watch?v=DyLJThawgi0}.

Motion imagery taken from a moving camera (\eg video taken from an
unmanned arial vehicle) is often jittery due to mechanical vibrations
in the camera platform. Such jittering often poses difficulty to the
data thinning task. The magnitude of the vibrations precludes
standard video stabilization techniques used, for instance, for
handheld video cameras.  {This experiment demonstrates} that the
proposed method can robustly flag salient objects from a jittery
video; the flagged patches can then be processed off-line (as
discussed in the introduction), and software video-stabilization
methods can be applied to these frames {alone} to co-register them.

\begin{figure}[!t]
\centering
\subfloat[Jittering video frame 60]{\includegraphics[width=0.33\textwidth]{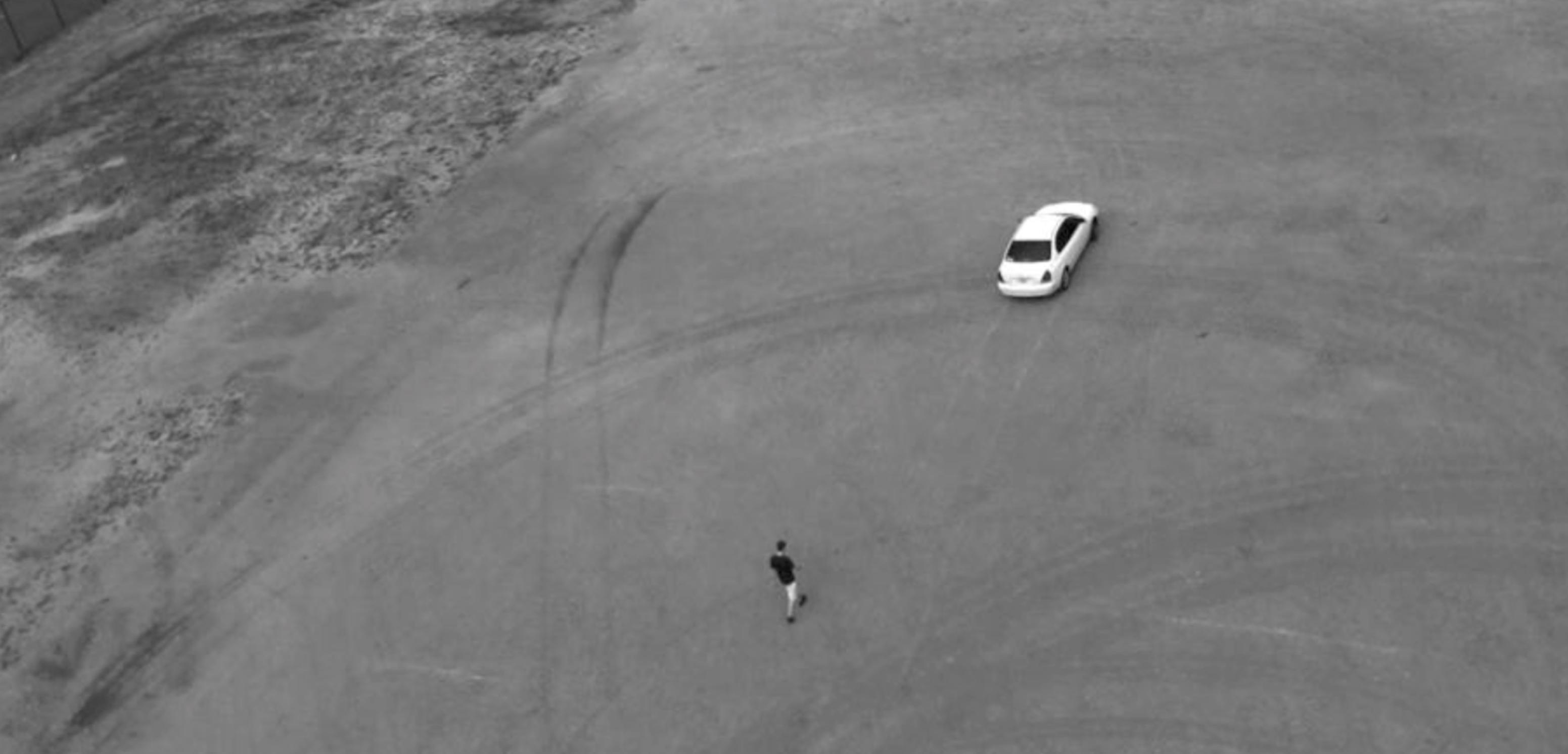}\label{fig:stb11}}~
\subfloat[Stabilized detection  frame 60]{\includegraphics[width=0.33\textwidth]{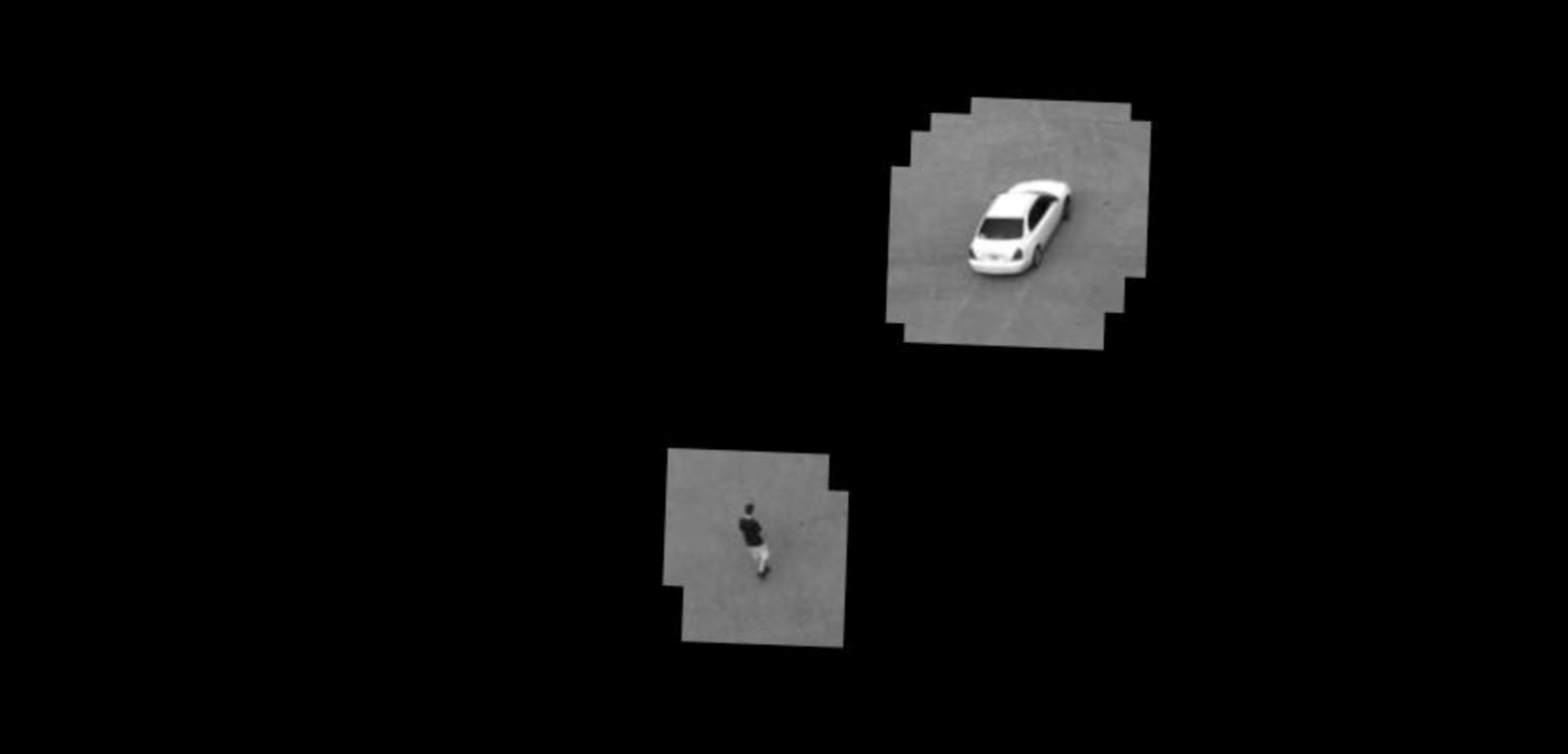}\label{fig:stb12}} \\
\subfloat[Jittering video frame 61]{\includegraphics[width=0.33\textwidth]{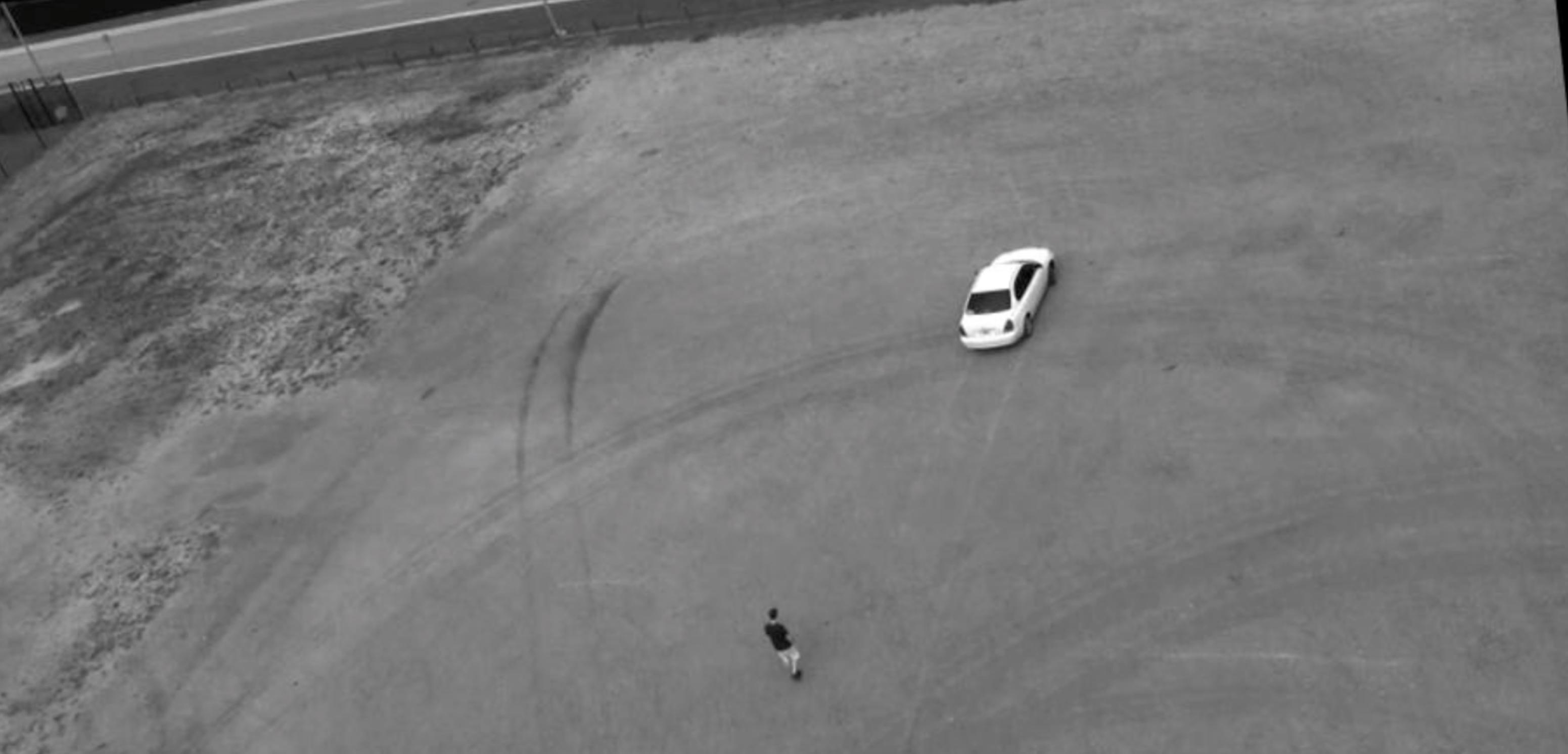}\label{fig:stb21}}~
\subfloat[Stabilized detection  frame 61]{\includegraphics[width=0.33\textwidth]{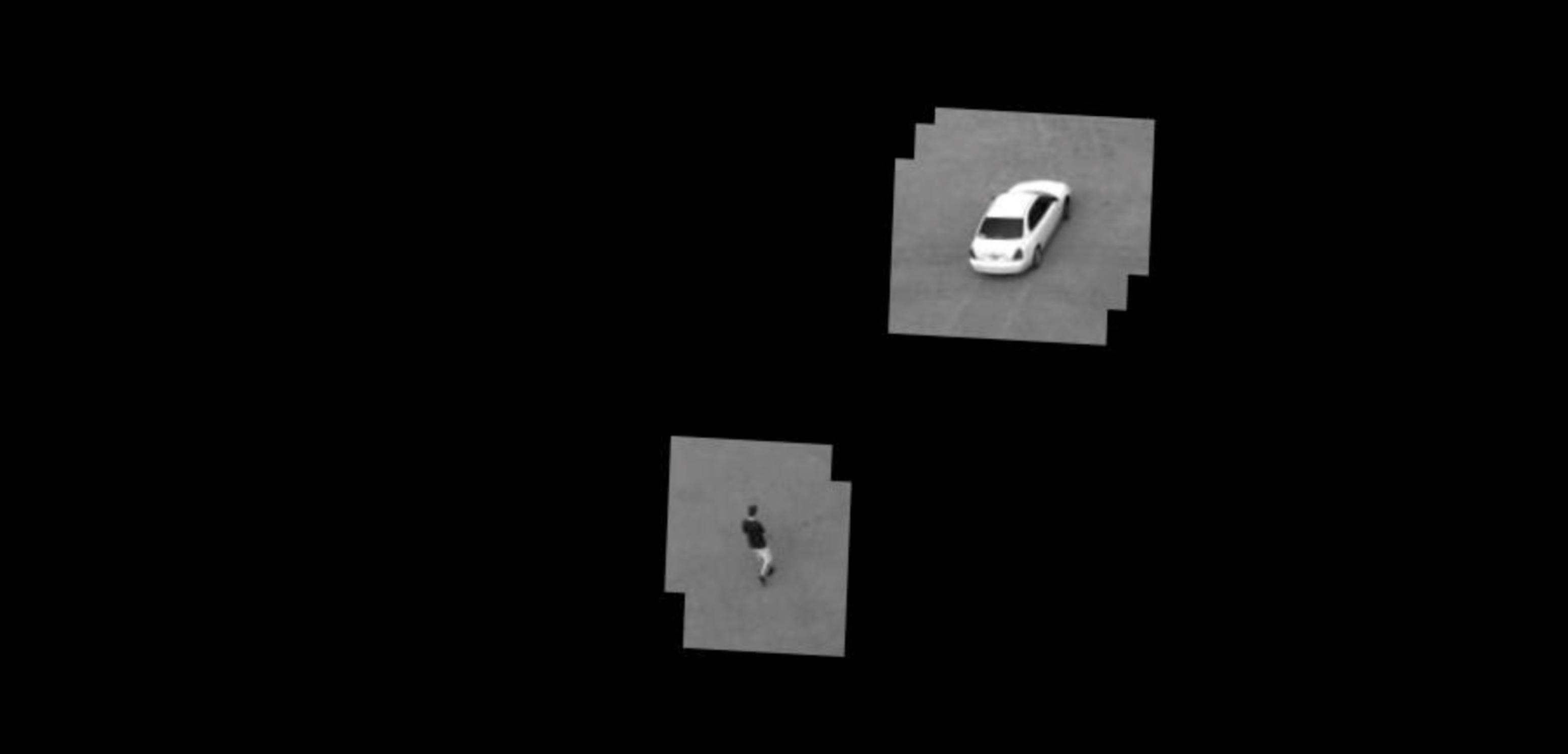}\label{fig:stb22}} \\
\subfloat[Jittering video frame 120]{\includegraphics[width=0.33\textwidth]{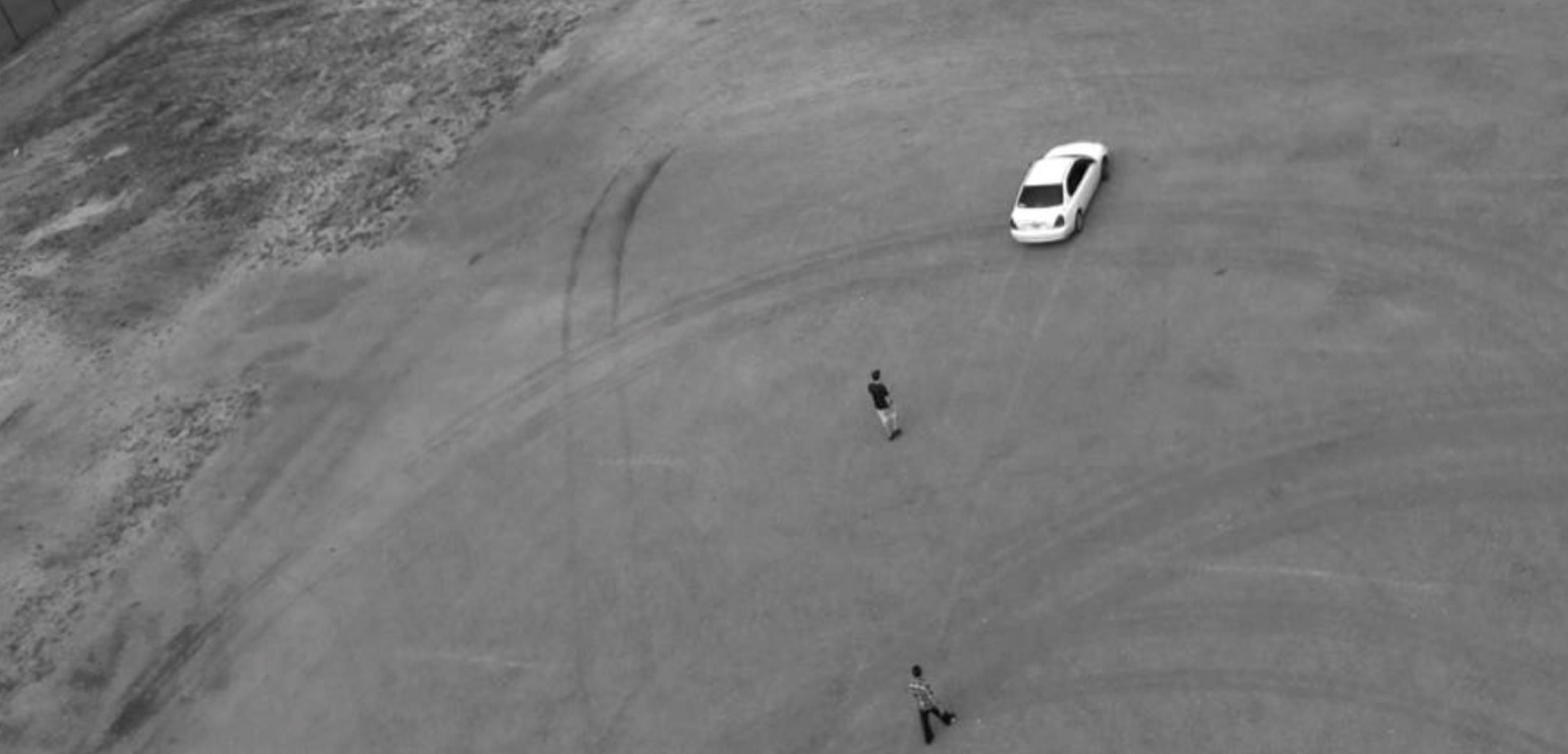}\label{fig:stb31}}~
\subfloat[Stabilized detection  frame 120]{\includegraphics[width=0.33\textwidth]{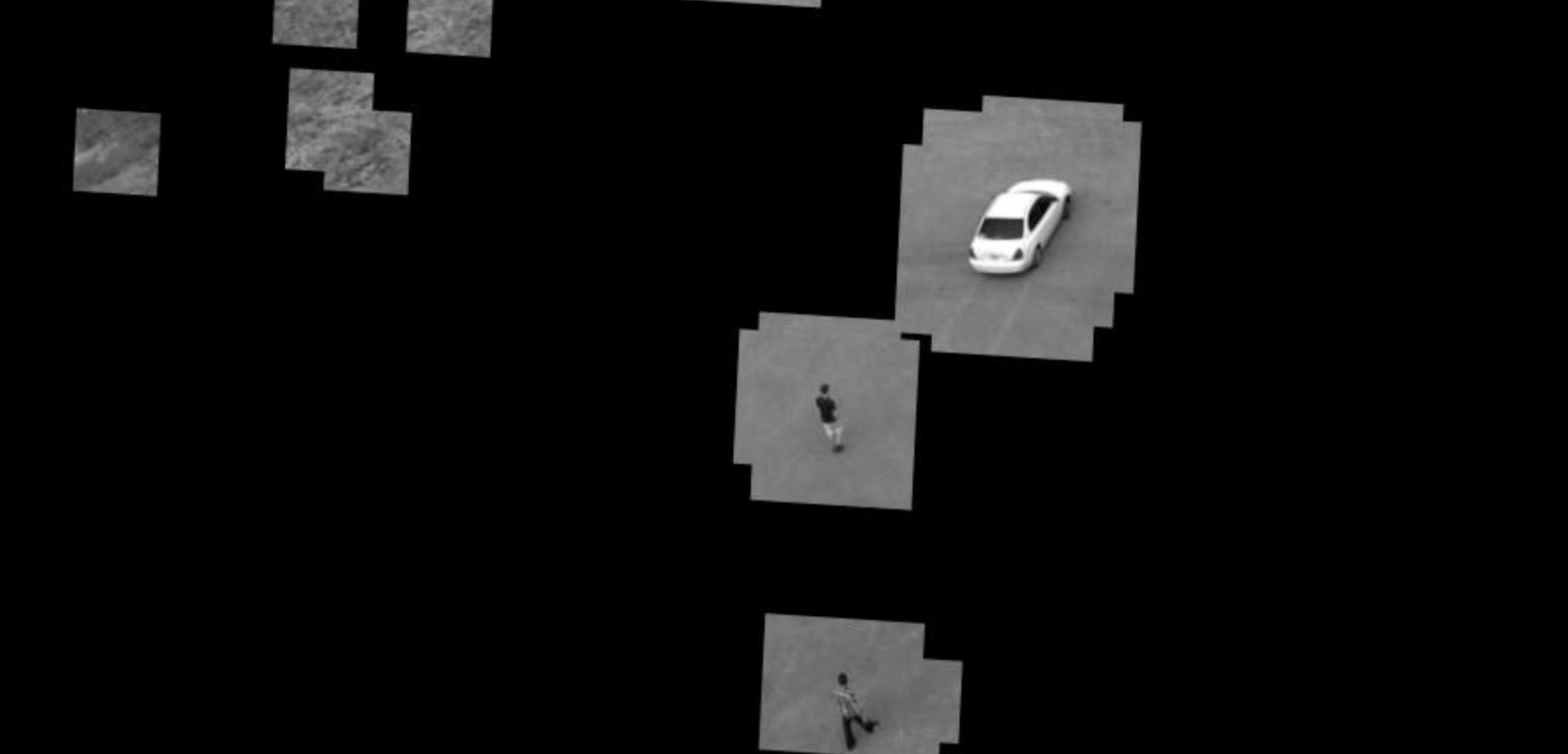}\label{fig:stb32}} \\
\subfloat[Jittering video frame 121]{\includegraphics[width=0.33\textwidth]{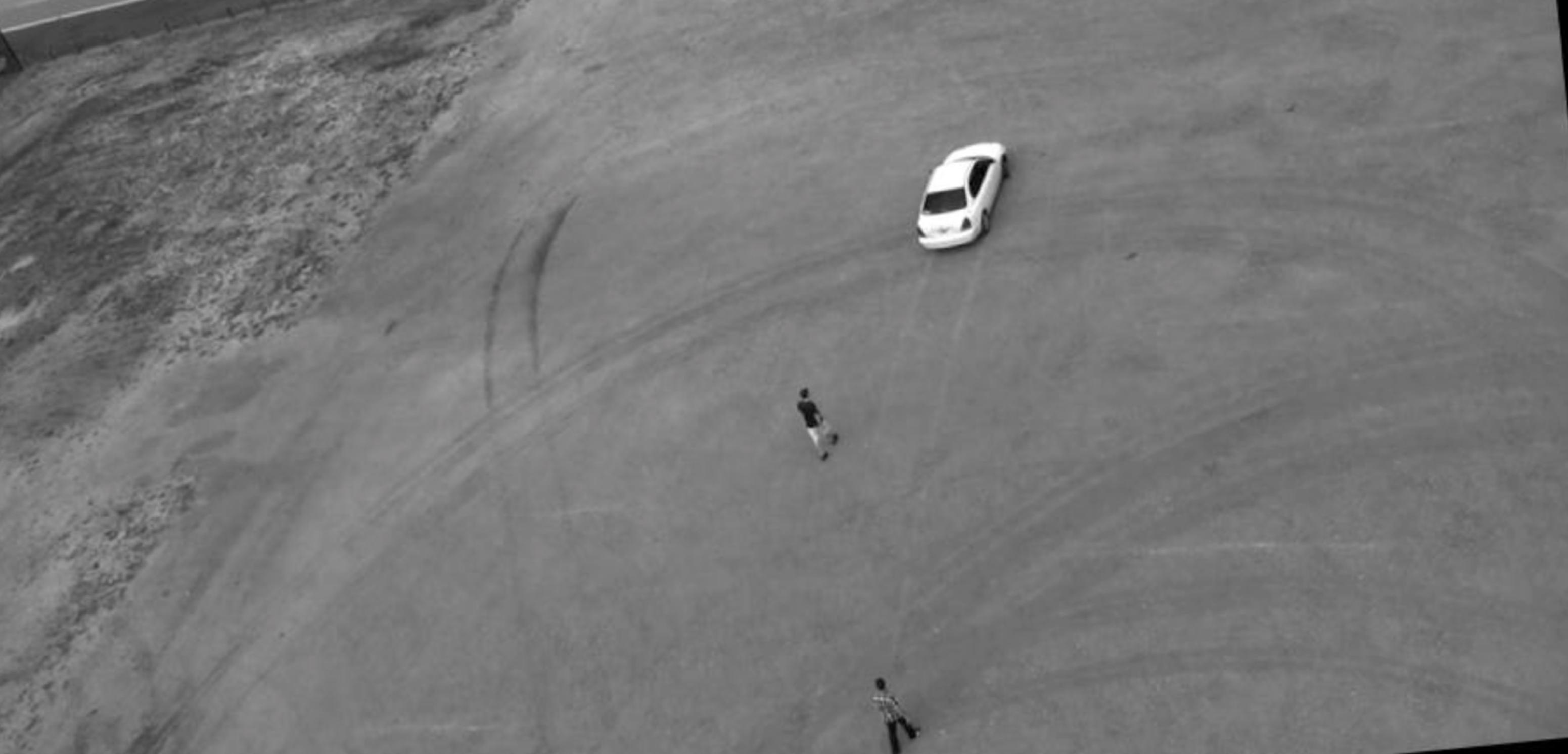}\label{fig:stb41}}~
\subfloat[Stabilized detection  frame 121]{\includegraphics[width=0.33\textwidth]{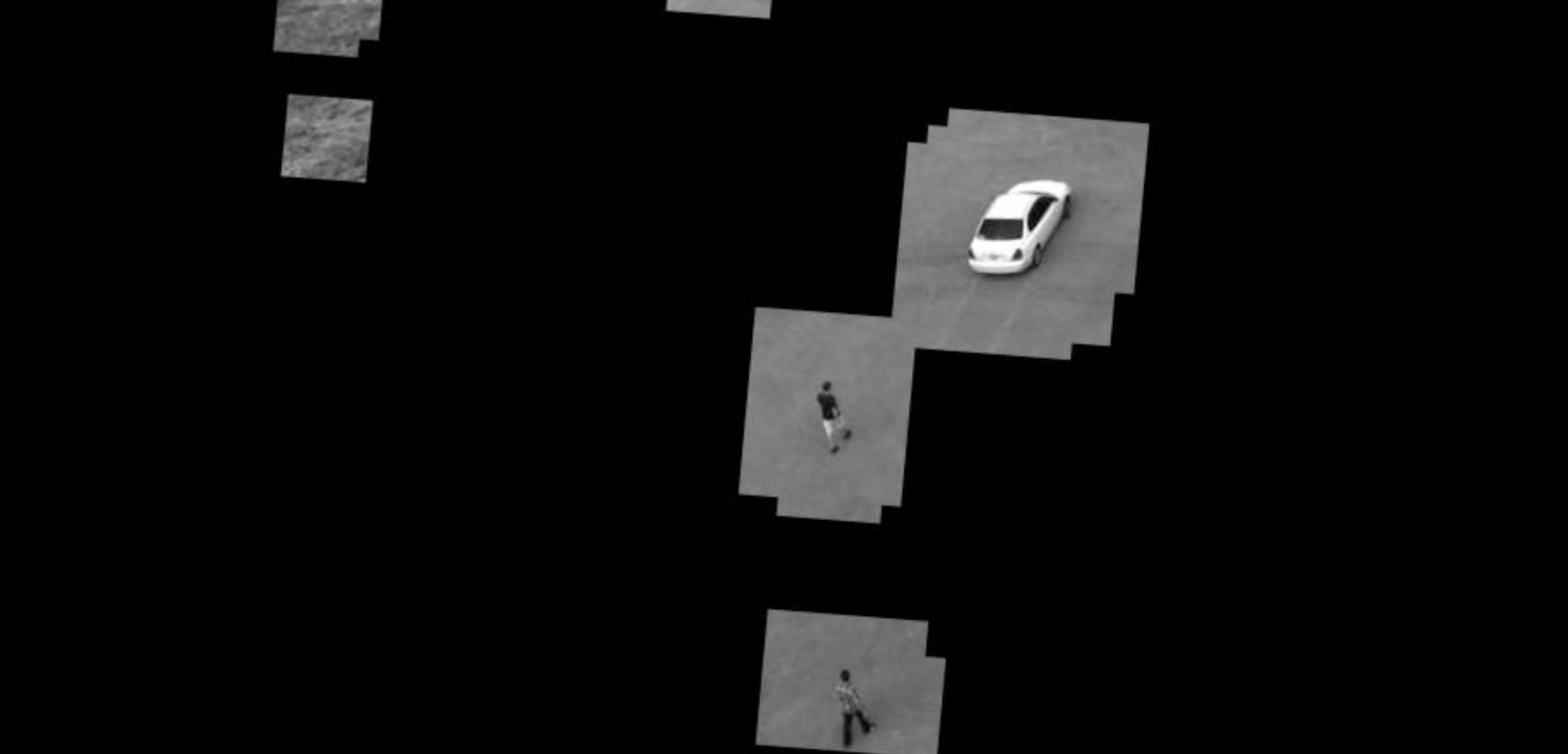}\label{fig:stb42}}
\caption{The original, jittered video frames (left columns) and
  corresponding thinning results after stabilization (right
  columns). The thinning result consistently shows the car and the
  people without significant shifting or shaking, even though the
  stabilization was performed using {\em only} the flagged patches.  }
\label{fig:stable2}
\end{figure}

Specifically, to demonstrate the effect of jittering, we artificially
add random rotations and small shifting to each of the frames before
processing.  The jittered video can be found at
\url{https://youtu.be/oKzIOryxR0s}.  Then, we flag and extract patches
with high anomalousness scores using {the} proposed Online Thinning
algorithm. Finally, we use a feature-matching-based approach {\em on
  only the flagged patches} to generate a stabilized, thinned
video~\cite{lee2009video,matsushita2005full}.  Fig.~\ref{fig:stable2}
shows the original jittered video frames
(left column) and corresponding stabilized detection results (right
column). Note that despite the rotation and shifting of the original
frames, the stabilized result is consistently showing the car
and the people without significant shifting or shaking. The
  result video can be found at \url{https://youtu.be/DyLJThawgi0}.

Under different situations, the meaning of ``anomalousness'' and ``saliency'' can also be
different. For example, a moving car on a busy street during daytime
may be seen as normal, while the same car should be considered
anomalous or salient if it appears in some vacant lot when no other cars are
around. Conventional non-adaptive saliency detection algorithms often
lack the flexibility of changing the definition of saliency over time,
while Online Thinning, as an online algorithm, has the ability to
learn the environment over time and adapt to new needs. 

{A third experiment compares} Online Thinning with the classical
batch and online GMM algorithm with a real-life parking lot
surveillance video data. The video is a time-lapse of a parking lot
where cars arrive and gradually fill up the entire lot.  For Online
Thinning and the classical online GMM, an initial model is estimated
using the first frame, and the models are then updated frame by
frame. The anomalousness score is calculated as the negative
log-likelihood of each data point according to the estimated
model. For the classical static GMM algorithm, we estimate a GMM model
on the first twenty frames, and assign an anomalousness score to each
data point proportional to the negative log likelihood of the data
point coming from the estimated GMM model.  {The batch
GMM is trained only with the patches from} the first twenty frames to simulate a setting
in which a probability model is learned in one set of environmental
conditions and does not adapt to a changing environment.
  
For all three algorithms, dense SIFT features from each frame
$t$ {are used} as the observation $X_t$.  Each frame of the video is of size
$960\times540$, and the {grid is placed} so that one SIFT feature is
computed for each $25\times 25$ patch.  {Each frame have} roughly
eight hundred SIFT feature vectors.  The dimension of each
SIFT feature vector is 128.  
   
\begin{figure}[!t]
\centering
\subfloat[Online Thinning, frame 21]{\includegraphics[width=0.33\textwidth]{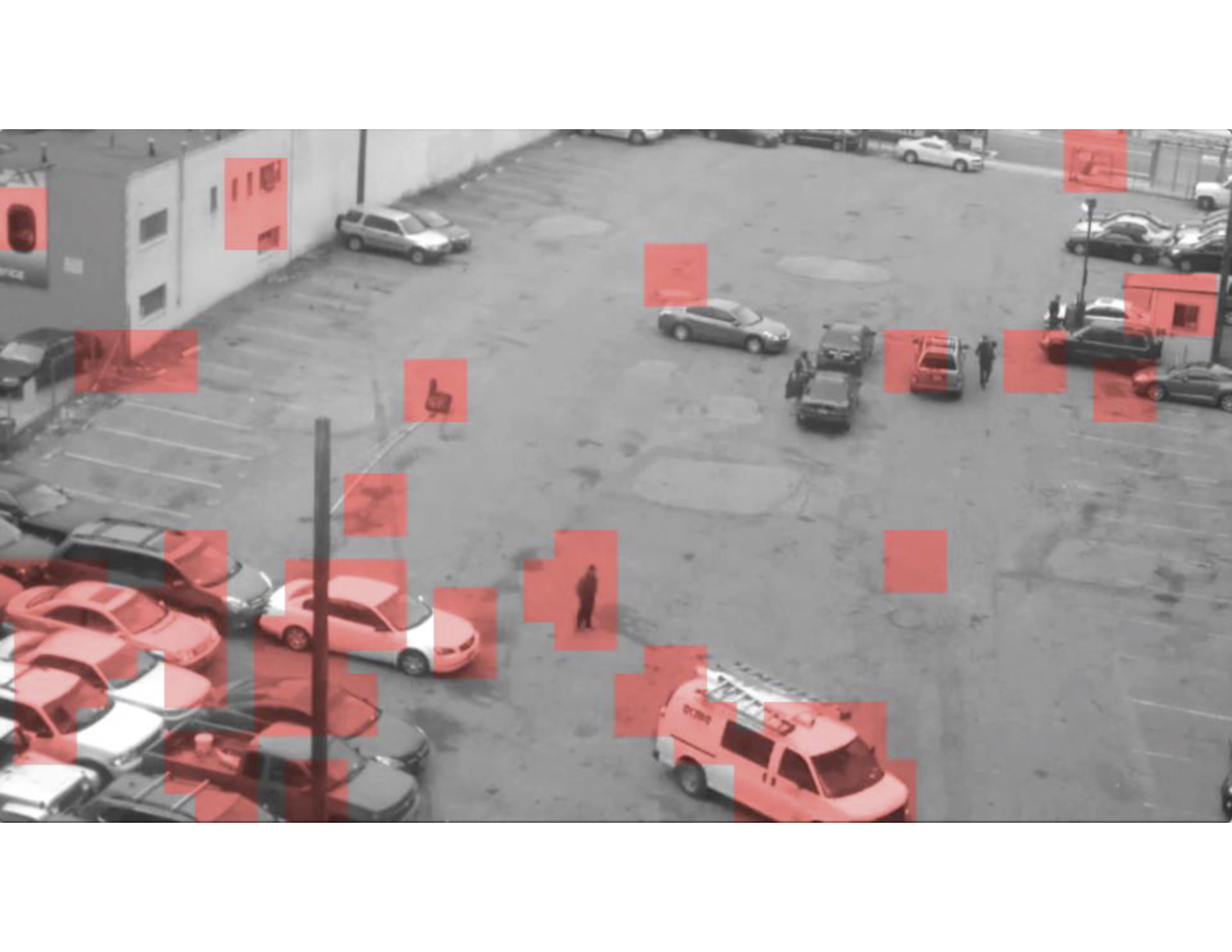}\label{fig:pklot211}}~
\subfloat[Online Thinning, frame 232]{\includegraphics[width=0.33\textwidth]{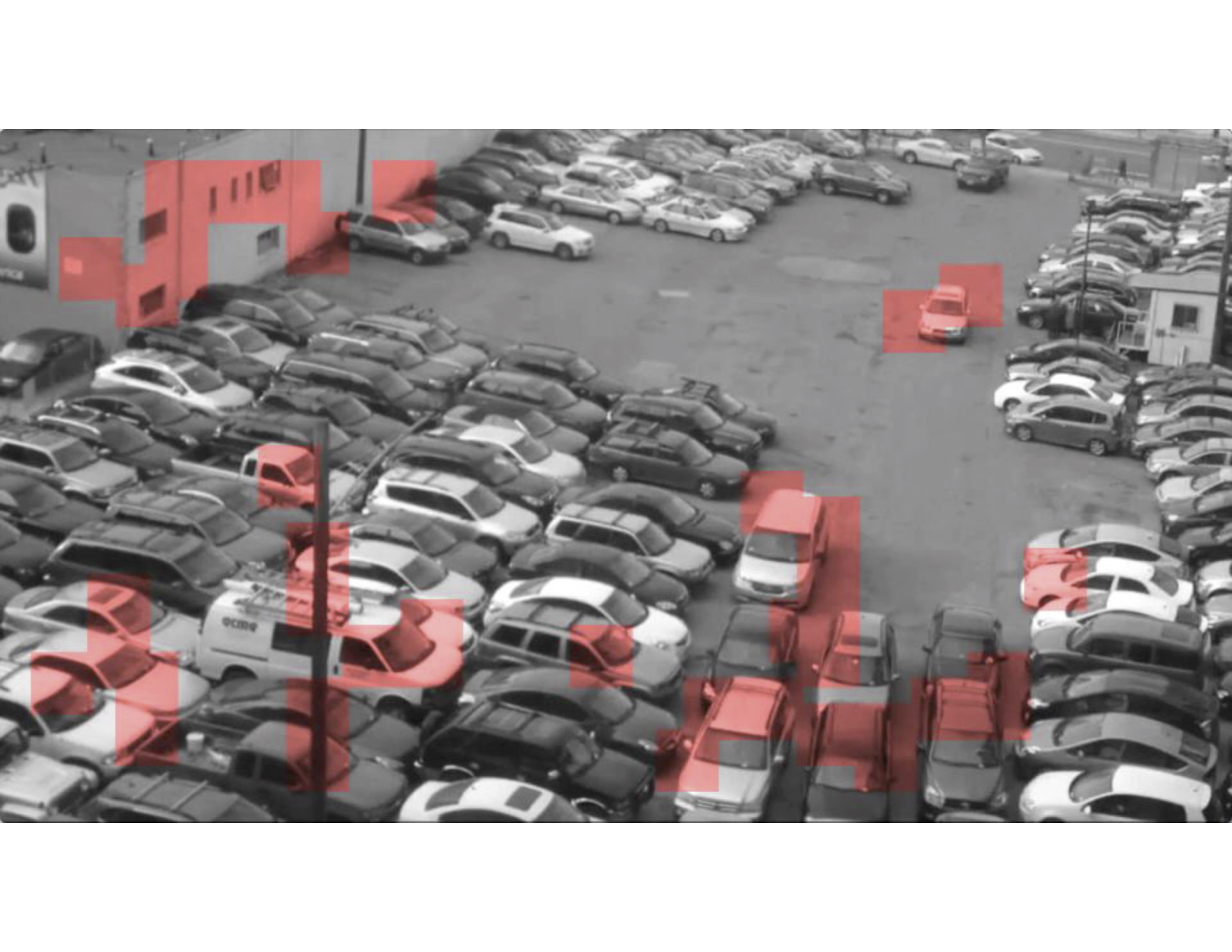}\label{fig:pklot221}}\\
\subfloat[Batch GMM, frame 21]{\includegraphics[width=0.33\textwidth]{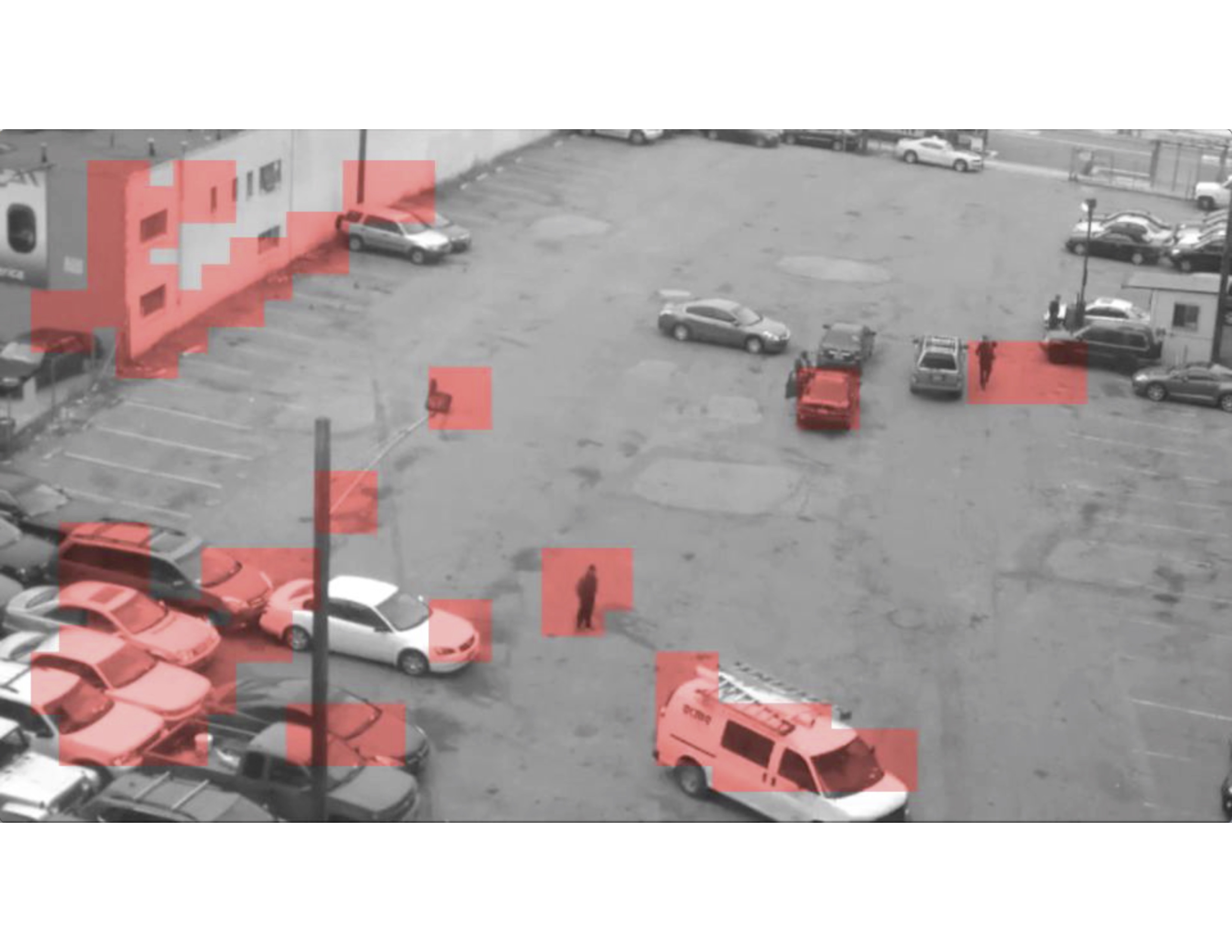}\label{fig:pklot212}} ~
\subfloat[Batch GMM, frame 232]{\includegraphics[width=0.33\textwidth]{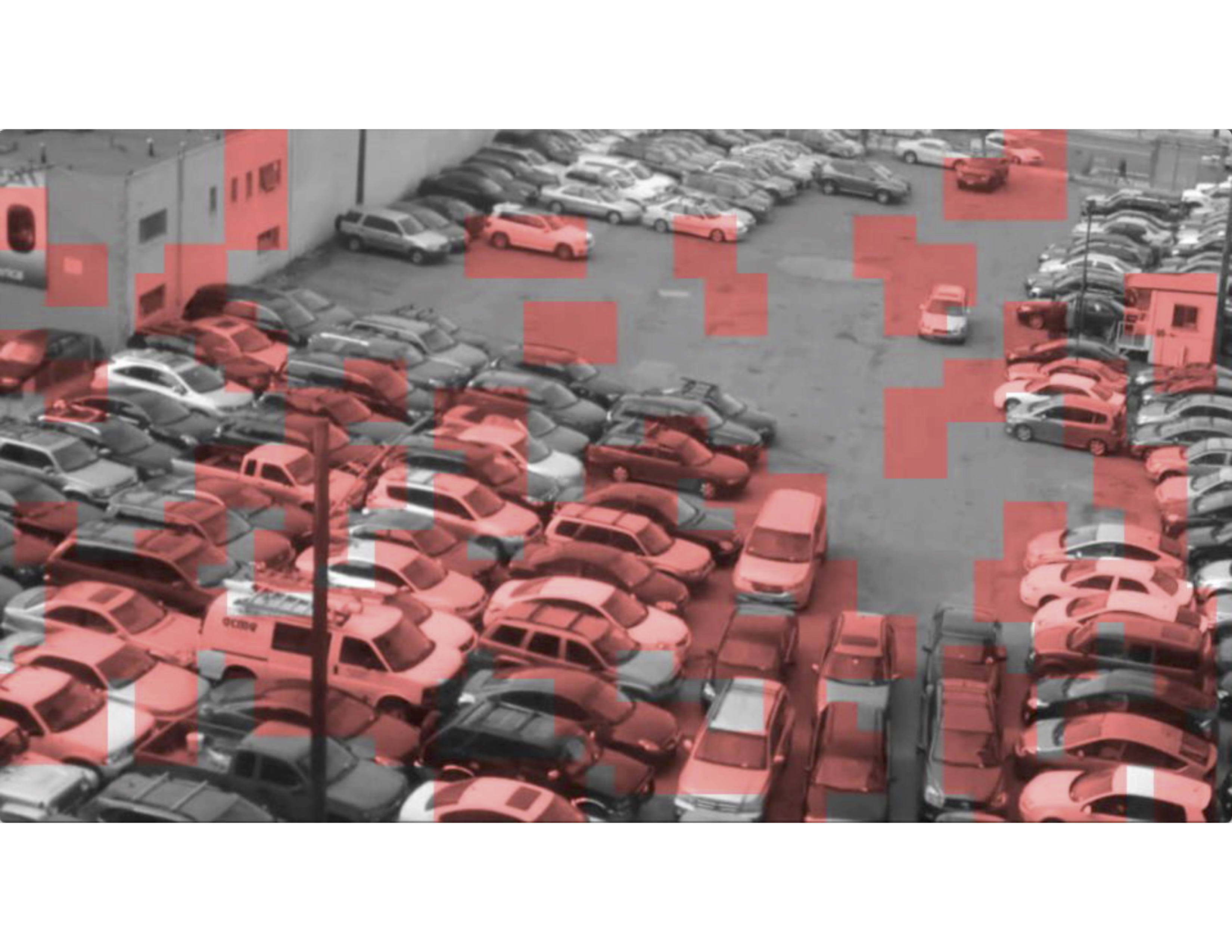}\label{fig:pklot222}} \\
\subfloat[Online GMM, frame 21]{\includegraphics[width=0.33\textwidth]{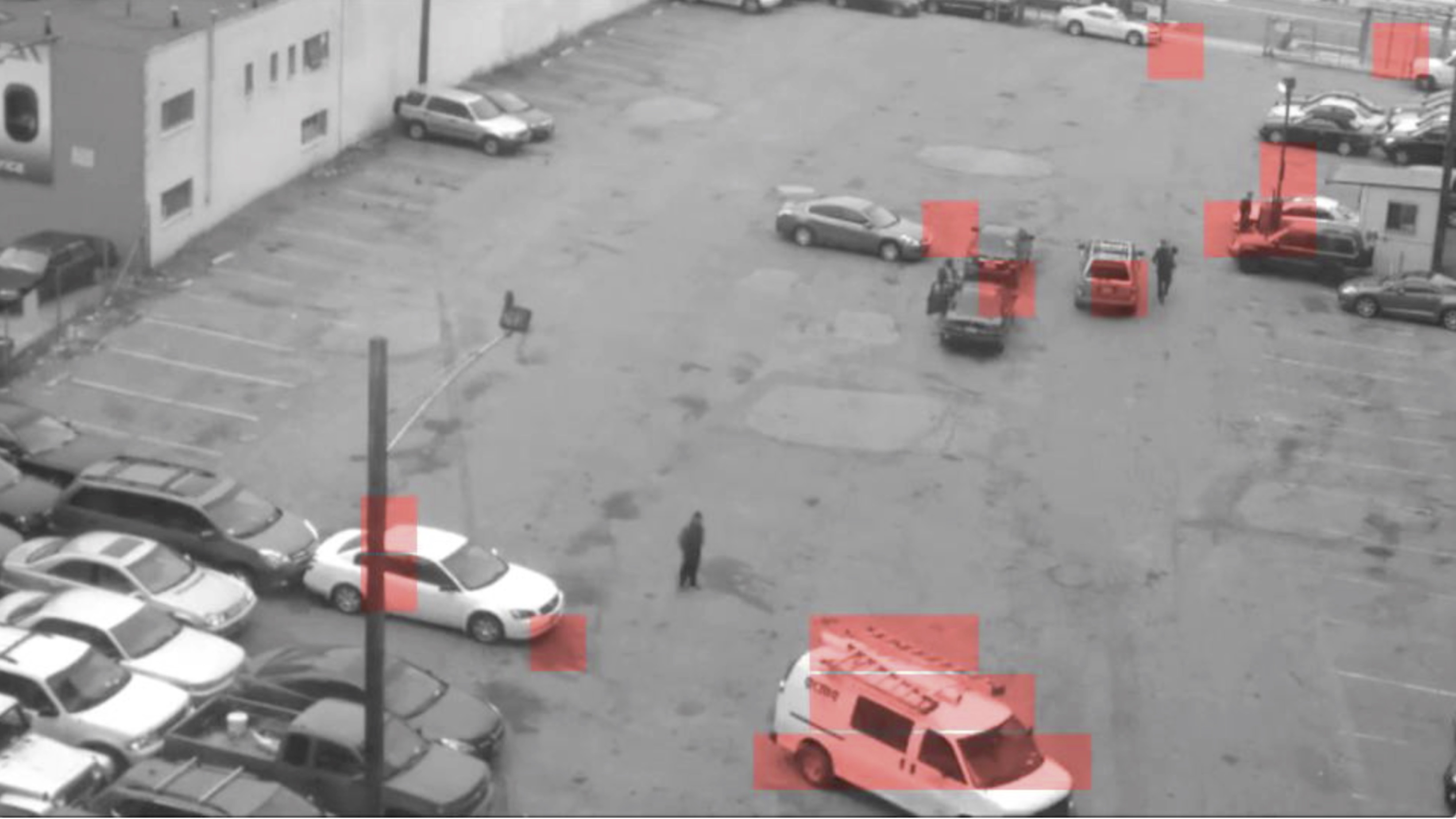}\label{fig:pklot213}} ~
\subfloat[Online GMM, frame 232]{\includegraphics[width=0.33\textwidth]{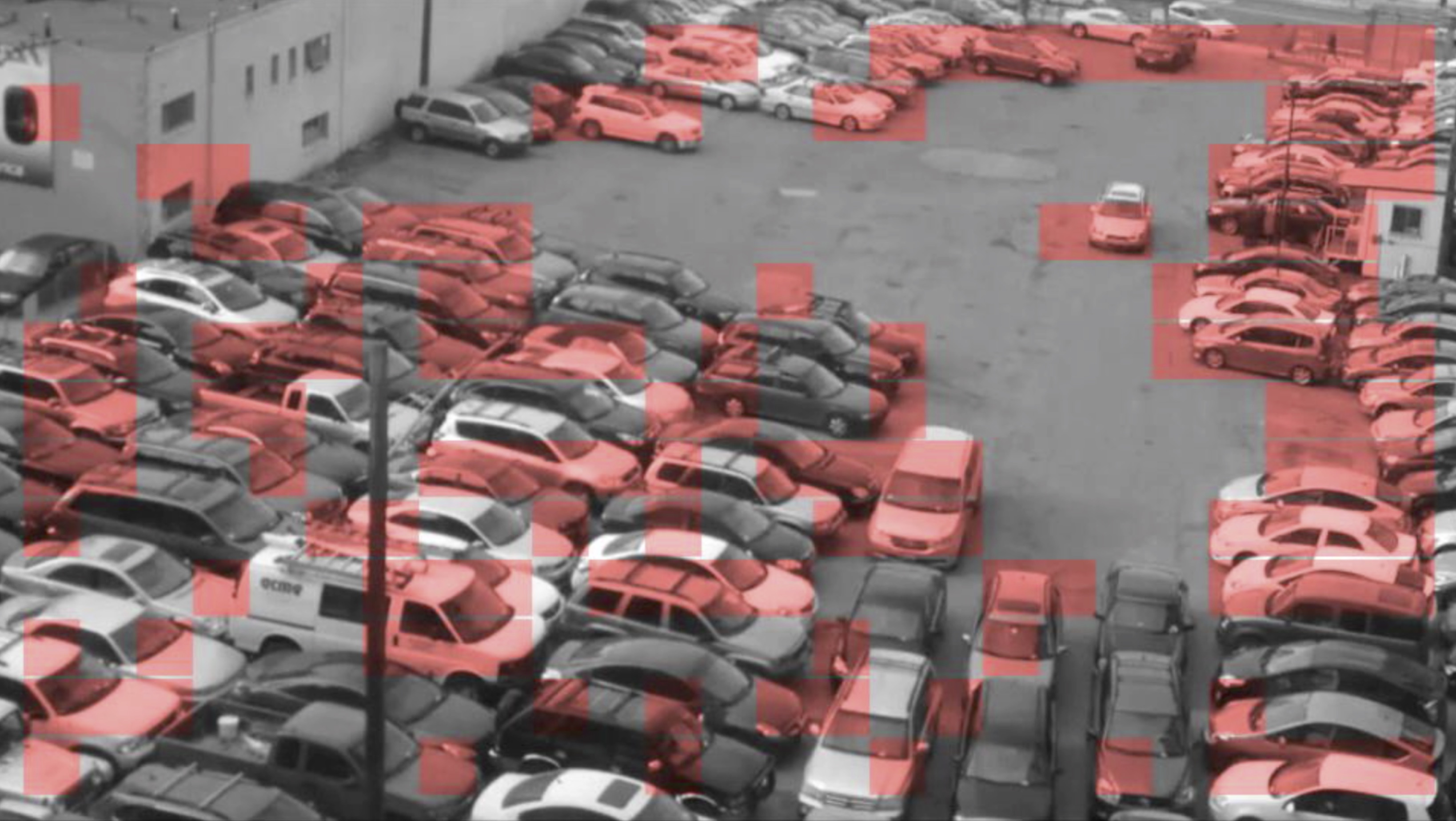}\label{fig:pklot223}} 
\caption{Result of Online Thinning and classical batch and online GMM
  algorithms on the surveillance video at frames 21 and
  232. Red-colored patches are flagged as salient according to the
  different probability
  models. \ref{fig:pklot211},~\ref{fig:pklot212},
  and~\ref{fig:pklot213} show the result on frame 21, where the lot is
  still relatively empty, and all three algorithms flag similar items
  in the scene (incoming car, people in the
  lot). \ref{fig:pklot221},~\ref{fig:pklot222}, and~\ref{fig:pklot223}
  show the result on frame 232, when the lot is about half full.  The
  Online Thinning algorithm has learned that cars are common objects
  in the video, and has thus adapted to assigning lower anomalousness
  scores to most cars.  The batch GMM algorithm does not adapt to the
  video, and assigns most cars with high anomalousness scores. The
  online GMM algorithm flags less cars with high anomalousness scores
  than batch GMM algorithm, but still flags more cars than the Online
  Learning algorithm. This suggests the online GMM algorithm adapts to
  the environment slower than the Online Learning algorithm.}
\label{fig:pklot2}
\end{figure}

Fig.~\ref{fig:pklot2} shows the result of Online Thinning
(Alg.~\ref{alg:OTM}) and both classical batch and online GMM
algorithms on the surveillance video at frames 21 and 232. Red-colored
patches are flagged as having high anomalousness scores.
Figures~\ref{fig:pklot211},~~\ref{fig:pklot212} and~\ref{fig:pklot213}
show the result on frame 21, where the lot is still relatively empty,
and all three algorithms flagged similar items in the scene (incoming
car, people in the
lot). Figures~\ref{fig:pklot221},~\ref{fig:pklot222}
and~\ref{fig:pklot223} show the result on frame 232, when the lot is
about half full.  At frame 232, the Online Thinning algorithm has
learned that cars are common objects in the video, and has thus
adapted to assigning lower anomalousness scores to most cars.
Instead, the Online Thinning algorithm assigns higher anomalousness
scores to relatively uncommon objects like black pole, building
windows, and cars parked differently from others.  The batch GMM
algorithm does not adapt to the video, and assigns most cars with high
anomalousness scores.  The online GMM algorithm flags fewer patches
than the batch GMM algorithm when the parking lot is filled
up. However, online GMM still flags significantly more cars than the
Online Thinning algorithm. Note that in the video, the parking lot is
filled up gradually, and most of the cars in the parking lot at frame
232 has shown up in the scene for a long time. However, at frame 232,
the Online GMM algorithm still flags a significant amount of cars in
the parking lot, while the Online Thinning algorithm has stopped
flagging cars that have been in the scene for a long time. This
suggests that the online GMM algorithm adapts to the environment at a
slower rate than the Online Thinning algorithm.

\section{Enron email experiments}
\label{sec:Enronexp}
Data thinning can also be applied to text documents to find anomalous
texts and topics. The development of latent Dirichlet allocation
(LDA)~\cite{blei2003latent} for text document topic modeling and other
methods have allowed us to analyze the topics of a collection of
documents. The Enron data is a collection of about fifty thousand
emails within the Enron corporation between the year 1998 and
2002. The dataset has been explored in the context of social network
analysis \cite{diesner2005communication} and event detection
\cite{aggarwal2012event, raginsky2012sequential, horn2011online}. In
\cite{raginsky2012sequential, horn2011online}, the authors used the
email addresses and time stamps and successfully predicted major
events in the company by finding days during which email
correspondence shows abnormal patterns. In our work, we also try to
detect significant events in the company's history by using the Enron
database. However, we approach the problem by using the count of
``topic words'' found in the emails, instead of the contact
information which does not reflect the content of the emails.

The challenge here is that the count data cannot be modeled as
Gaussian, and {pre-}processing {is needed} before applying the method. We
see each of the word-count of topic words in the email as an
independent Poisson realization of some underlying rate. By using the
Anscombe transformation \cite{anscombe1948transformation}, we can
approximate the normalized data as arising from a Gaussian mixture
model, and thus apply the Online Thinning
Algorithm~\ref{alg:OTM}.

{This experiment applies} the Online Thinning algorithm to the
Enron email dataset for event detection. To process the Enron emails,
we first generate a five-hundred-word topic list using LDA
\cite{darling2011theoretical}, where the list includes fifty
topics, and each topic has ten associated keywords. For each
email, the number of times each keyword appears {is counted and recorded}
in a fifty-dimensional vector $y_t\in\mathbb{N}^{50}$ where each entry
$[y_{t}]_i$ corresponds to how many times the keywords in topic $i$
appears in this email. Here $[\cdot]_i$ indicates the $i^{\mbox{th}}$
element of a vector. The feature vectors {are then normalized} using the
Anscombe transform \cite{anscombe} by setting
$[x_t]_i = 2\sqrt{[y_t]_i + \frac{3}{8}};$
note that $[x_t]_i$ is asymptotically normal with mean
$2\sqrt{[y_t]_i + \frac{3}{8}} + \frac{1}{4\sqrt{[y_t]_i}}$ and unit
variance.  Online Thinning is then applied to the transformed data
data (the $x_t$'s), and we flag emails by thresholding the
anomalousness score assigned by the Online Thinning algorithm.

\begin{figure}[!t]
\centering
{\includegraphics[width=0.6\textwidth]{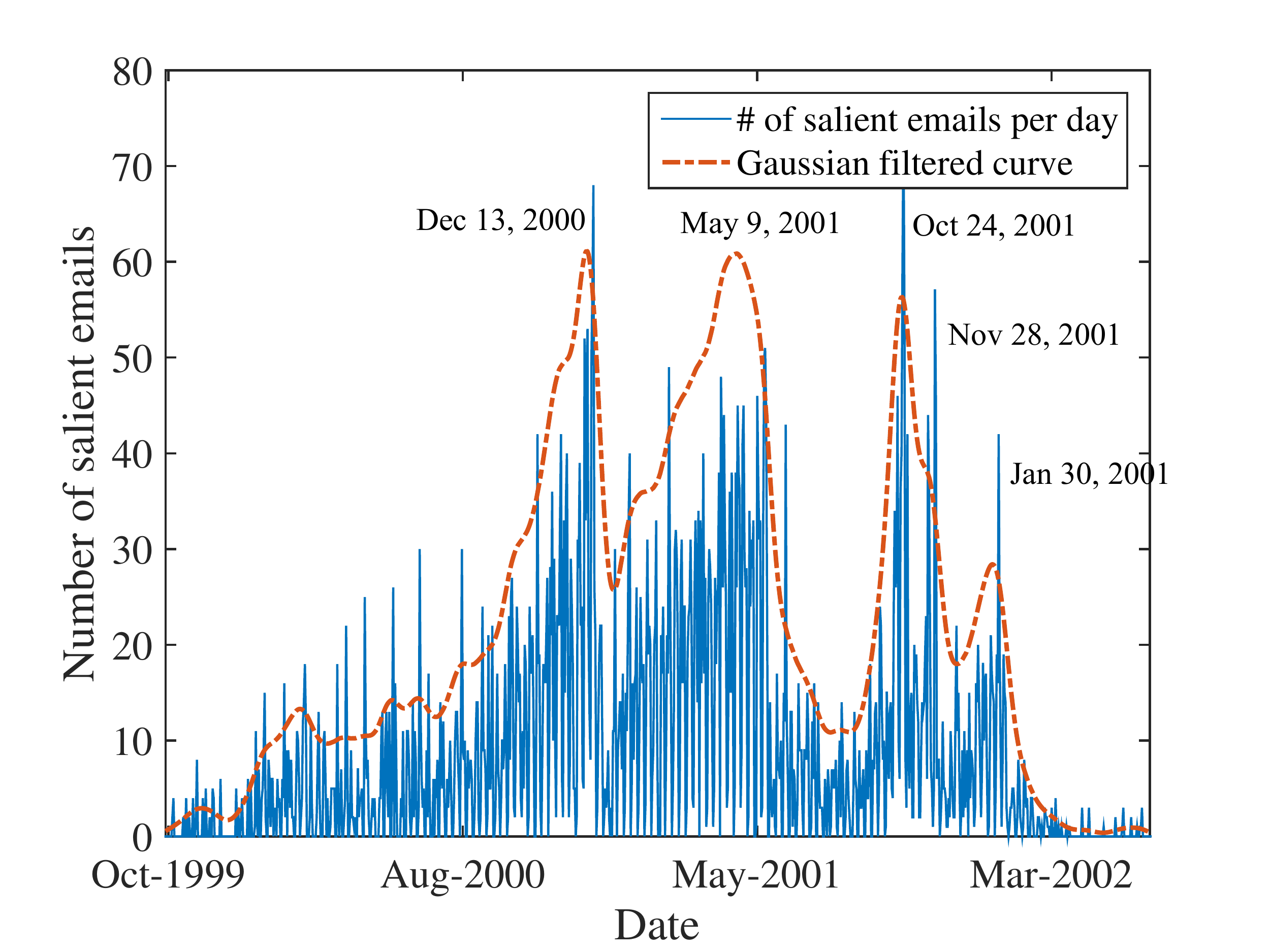}}
\caption{number of selected emails versus time (date). The large peaks in the plot all correspond to major events in the history of the company. The red curve is smoothed using a Gaussian filter.}
\label{fig:Enron}
\end{figure}

Fig.~\ref{fig:Enron} shows the number of selected emails versus time
(date).  The major peaks in the plot correspond to the following time
and events:
\begin{enumerate}
\item December 13, 2000: Enron announces that president and chief
  operating officer Jeffrey Skilling will take over as chief executive
  in February. \cite{Dec13}
\item May 9, 2001: ``California utility says prices of gas were
  inflated'' by Enron collaborator, and blackouts affect 167,00 Enron
  customers. \cite{May09,May08}
\item October 24, 2001: Enron ousts its chief financial officer Andrew
  S. Fastow, and the shares of Enron fell to the lowest price since
  early 1995 \cite{ousts}.
\item November 28, 2001: Enron shares plunge below \$1. \cite{Nov28}
\item January 30, 2001: Stephen Cooper takes over as Enron CEO, and
  Enron Metals is sold to a unit of Sempra Energy. \cite{Jan29, Jan30}
\end{enumerate}
As seen, the flagged dates cluster around the time when significant
events happen in the Enron company.

\section{Conclusion}
{This paper} proposed an online data thinning method for
high-dimensional data with changing environment. At the heart of the proposed
algorithm is a union of subspaces tracking algorithm,
which allows for fast and accurate data thinning in a variety of
applications with both subsampled data and mini-batch updates.

The core idea of the proposed approach is to track a Gaussian mixture
model whose covariance matrices each are dominated by a low-rank
component. Under this model, most observations are concentrated in a
union of subspaces, a model growing in popularity in image, video, and
text analysis because of its flexibility and robustness to
over-fittings. Unlike traditional GMMs, the low-rank structure
proposed here mitigates the curse of dimensionality and facilitates
efficient tracking in dynamic environments.  Furthermore, by
leveraging the recent advances in subspace tracking and subspace
clustering techniques, the proposed method is able to accurately
estimate the mixture density without adding a significant
computational burden. Another important feature of the proposed method
is the ability to track an arbitrary number of mixture components. The
adoption of a tree-like hierarchical structure for the union of
subspaces model allows the method to adaptively choose the number of
subspaces needed at each time stamp, and thus greatly {improves} the
flexibility of the method and accuracy when tracking highly dynamic
densities.

\bibliography{OnlineDataThinning_arxiv}

\begin{thebibliography}{100}

\bibitem{sept11}
W.~Pincus and D.~Priest.
\newblock {NSA} intercepts on eve of 9/11 sent a warning: Messages translated
  after attacks, June 20, 2002.
\newblock The Washington Post, Page A01.

\bibitem{JPLBigData}
W.~Clavin.
\newblock Managing the deluge of `big data' from space, 2013.
\newblock \url{http://www.jpl.nasa.gov/news/news.php?release=2013-299}.

\bibitem{WiredSKA}
J.~Marlow.
\newblock What to do with 1,000,000,000,000,000,000 bytes of astronomical data
  per day, 2012.
\newblock Wired,
  \url{http://www.wired.com/2012/04/what-to-do-with-1000000000000000000-bytes-of-astronomical-data-per-day/}.

\bibitem{LSSTWeb}
{LSST Team}.
\newblock {LSST}: A new telescope concept.
\newblock \url{http://www.lsst.org/lsst/public/tour_software}, retrieved June
  7, 2014.

\bibitem{LSST}
Z.~Ivezic and et~al. (108 additional authors~not shown).
\newblock {LSST}: from science drivers to reference design and anticipated data
  products.
\newblock \url{http://arxiv.org/abs/0805.2366}{arXiv:0805.2366}, 2008.

\bibitem{LHC}
E.~Dumbill.
\newblock What is big data? {A}n introduction to the big data landscape, 2012.
\newblock \url{http://strata.oreilly.com/2012/01/what-is-big-data.html}.

\bibitem{raginsky_OCP}
M.~Raginsky, R.~Willett, C.~Horn, J.~Silva, and R.~Marcia.
\newblock Sequential anomaly detection in the presence of noise and limited
  feedback.
\newblock {\em IEEE Trans. Info. Theory}, 58(8):5544 -- 5562, Aug. 2012.

\bibitem{onlineSocialAnomalies}
C.~Horn and R.~Willett.
\newblock Online anomaly detection with expert system feedback in social
  networks.
\newblock In {\em Proc. International Conference on Acoustics, Speech, and
  Signal Processing}, 2011.

\bibitem{bellman1961adaptive}
R.~Bellman.
\newblock {\em Adaptive control processes: a guided tour}, volume~4.
\newblock Princeton University Press, Princeton, NJ, 1961.

\bibitem{hastie2009element}
T.~Hastie, R.~Tibshirani, and J.~Friedman.
\newblock {\em The Elements of Statistical Learning: Data Mining, Inference,
  and Prediction}.
\newblock Springer-Verlag New York, 2009.

\bibitem{argus}
Doug Bezier.
\newblock {BAE} to develop surveillance system.
\newblock {\em The Washington Post}, 2007.
\newblock Retrieved 3-20-2012.

\bibitem{argus2}
D.~Hambling.
\newblock Special forcesÕ gigapixel flying spy sees all, 2009.
\newblock \url{http://www.wired.com/dangerroom/2009/02/gigapixel-flyin/}.

\bibitem{mosaic3}
D.~J. Brady, M.~E. Gehm, R.~A. Stack, D.~L. Marks, D.~S. Kittle, D.~R. Golish,
  E.~M. Vera, and S.~D. Feller.
\newblock Multiscale gigapixel photography.
\newblock {\em Nature}, 486:386--389, 2012.

\bibitem{mosaic4}
K.~Bourzac.
\newblock Gigapixel camera catches the smallest details.
\newblock {\em Nature News}, 2012.

\bibitem{mosaic2}
L.~Greenemeier.
\newblock Draw the curtains: Gigapixel cameras create highly revealing
  snapshots, 2011.
\newblock
  {\small{\url{http://www.scientificamerican.com/article.cfm?id=gigapixel-camera-revealed}}}.

\bibitem{SaliencyItti}
L.~Itti, C.~Koch, and E.~Niebur.
\newblock A model of saliency-based visual attention for rapid scene analysis.
\newblock {\em IEEE Transactions on Pattern Analysis \& Machine Intelligence},
  20(11):1254--1259, 1998.

\bibitem{SaliencySR}
X.~Hou and L.~Zhang.
\newblock Saliency detection: A spectral residual approach.
\newblock In {\em Computer Vision and Pattern Recognition, 2007. CVPR'07. IEEE
  Conference on}, pages 1--8. IEEE, 2007.

\bibitem{SaliencyGBVS}
J.~Harel, C.~Koch, and P.~Perona.
\newblock Graph-based visual saliency.
\newblock In {\em Advances in neural information processing systems}, pages
  545--552, 2006.

\bibitem{rao2010using}
N.~Rao, J.~Harrison, T.~Karrels, R.~Nowak, and T.~T. Rogers.
\newblock Using machines to improve human saliency detection.
\newblock In {\em Signals, Systems and Computers (ASILOMAR), 2010 Conference
  Record of the Forty Fourth Asilomar Conference on}, pages 80--84. IEEE, 2010.

\bibitem{erturk2002real}
S.~Ert{\"u}rk.
\newblock Real-time digital image stabilization using kalman filters.
\newblock {\em Real-Time Imaging}, 8(4):317--328, 2002.

\bibitem{hansen1994real}
M.~Hansen, P.~Anandan, K.~Dana, G.~Van~der Wal, and P.~Burt.
\newblock Real-time scene stabilization and mosaic construction.
\newblock In {\em Applications of Computer Vision, 1994., Proceedings of the
  Second IEEE Workshop on}, pages 54--62. IEEE, 1994.

\bibitem{ratakonda1998real}
K.~Ratakonda.
\newblock Real-time digital video stabilization for multi-media applications.
\newblock In {\em Circuits and Systems, 1998. ISCAS'98. Proceedings of the 1998
  IEEE International Symposium on}, volume~4, pages 69--72. IEEE, 1998.

\bibitem{chang2006robust}
H.-C. Chang, S.-H. Lai, and K.-R. Lu.
\newblock A robust real-time video stabilization algorithm.
\newblock {\em Journal of Visual Communication and Image Representation},
  17(3):659--673, 2006.

\bibitem{battiato2010robust}
S.~Battiato, A.~R. Bruna, and G.~Puglisi.
\newblock A robust block-based image/video registration approach for mobile
  imaging devices.
\newblock {\em Multimedia, IEEE Transactions on}, 12(7):622--635, 2010.

\bibitem{same2007online}
A.~Sam{\'e}, C.~Ambroise, and G.~Govaert.
\newblock An online classification em algorithm based on the mixture model.
\newblock {\em Statistics and Computing}, 17(3):209--218, 2007.

\bibitem{BalzanoNowakRecht2010}
L.~Balzano, R.~Nowak, and B.~Recht.
\newblock Online identification and tracking of subspaces from highly
  incomplete information.
\newblock In {\em Proc. Allerton Conf. on Comm., Control and Comp.}, pages 704
  -- 711, Sept. 2010.

\bibitem{ChiJournal2012}
Y.~Chi, Y.~C. Eldar, and R.~Calderbank.
\newblock Petrels: Parallel estimation and tracking of subspace by recursive
  least squares from partial observations.
\newblock {\em submitted to IEEE Trans. Sig. Proc., arXived.}, July 2012.

\bibitem{ChiEldarCalderbank2012}
Y.~Chi, Y.~C. Eldar, and R.~Calderbank.
\newblock {PETRELS: Subspace estimation and tracking from partial
  observations}.
\newblock In {\em IEEE Int. Conf. on Acoustics, Speech and Sig. Proc.
  (ICASSP)}, 2012.

\bibitem{roseta}
H.~Mansour and X.~Jiang.
\newblock A robust online subspace estimation and tracking algorithm.
\newblock In {\em IEEE International Conference on Acoustics, Speech and Signal
  Processing (ICASSP)}, April 2015.

\bibitem{aggarwal2000finding}
C.~C. Aggarwal and P.~S. Yu.
\newblock {\em Finding generalized projected clusters in high dimensional
  spaces}, volume~29.
\newblock ACM, 2000.

\bibitem{bradley2000k}
P.~S. Bradley and O.~L. Mangasarian.
\newblock k-plane clustering.
\newblock {\em Journal of Global Optimization}, 16(1):23--32, 2000.

\bibitem{bohm2004computing}
C.~B{\"o}hm, K.~Kailing, P.~Kr{\"o}ger, and A.~Zimek.
\newblock Computing clusters of correlation connected objects.
\newblock In {\em Proceedings of the 2004 ACM SIGMOD international conference
  on Management of data}, pages 455--466. ACM, 2004.

\bibitem{achtert2006mining}
E.~Achtert, C.~B{\"o}hm, P.~Kr{\"o}ger, and A.~Zimek.
\newblock Mining hierarchies of correlation clusters.
\newblock In {\em Scientific and Statistical Database Management, 2006. 18th
  International Conference on}, pages 119--128. IEEE, 2006.

\bibitem{achtert2007robust}
E.~Achtert, C.~B{\"o}hm, H.-P. Kriegel, P.~Kr{\"o}ger, and A.~Zimek.
\newblock Robust, complete, and efficient correlation clustering.
\newblock In {\em SDM}, pages 413--418. SIAM, 2007.

\bibitem{achtert2007exploring}
E.~Achtert, C.~B{\"o}hm, H.-P. Kriegel, P.~Kr{\"o}ger, and A.~Zimek.
\newblock On exploring complex relationships of correlation clusters.
\newblock In {\em Scientific and Statistical Database Management, 2007.
  SSBDM'07. 19th International Conference on}, pages 7--7. IEEE, 2007.

\bibitem{edgeworth1887xli}
F.~Y. Edgeworth.
\newblock Xli. on discordant observations.
\newblock {\em The London, Edinburgh, and Dublin Philosophical Magazine and
  Journal of Science}, 23(143):364--375, 1887.

\bibitem{chandola2009anomaly}
V.~Chandola, A.~Banerjee, and V.~Kumar.
\newblock Anomaly detection: A survey.
\newblock {\em ACM computing surveys (CSUR)}, 41(3):15, 2009.

\bibitem{de2000reject}
C.~De~Stefano, C.~Sansone, and M.~Vento.
\newblock To reject or not to reject: that is the question-an answer in case of
  neural classifiers.
\newblock {\em Systems, Man, and Cybernetics, Part C: Applications and Reviews,
  IEEE Transactions on}, 30(1):84--94, 2000.

\bibitem{barbara2001detecting}
D.~Barbara, N.~Wu, and S.~Jajodia.
\newblock Detecting novel network intrusions using bayes estimators.
\newblock In {\em SDM}, pages 1--17. SIAM, 2001.

\bibitem{scholkopf2001estimating}
B.~Sch{\"o}lkopf, J.~C. Platt, J.~Shawe-Taylor, A.~J. Smola, and R.~C.
  Williamson.
\newblock Estimating the support of a high-dimensional distribution.
\newblock {\em Neural computation}, 13(7):1443--1471, 2001.

\bibitem{roth2004outlier}
V.~Roth.
\newblock Outlier detection with one-class kernel fisher discriminants.
\newblock In {\em Advances in Neural Information Processing Systems}, pages
  1169--1176, 2004.

\bibitem{roth2006kernel}
V.~Roth.
\newblock Kernel fisher discriminants for outlier detection.
\newblock {\em Neural computation}, 18(4):942--960, 2006.

\bibitem{zhang2006detecting}
J.~Zhang and H.~Wang.
\newblock Detecting outlying subspaces for high-dimensional data: the new task,
  algorithms, and performance.
\newblock {\em Knowledge and information systems}, 10(3):333--355, 2006.

\bibitem{otey2006fast}
M.~E. Otey, A.~Ghoting, and S.~Parthasarathy.
\newblock Fast distributed outlier detection in mixed-attribute data sets.
\newblock {\em Data Mining and Knowledge Discovery}, 12(2-3):203--228, 2006.

\bibitem{ghoting2008fast}
A.~Ghoting, S.~Parthasarathy, and M.~E. Otey.
\newblock Fast mining of distance-based outliers in high-dimensional datasets.
\newblock {\em Data Mining and Knowledge Discovery}, 16(3):349--364, 2008.

\bibitem{tao2006mining}
Y.~Tao, X.~Xiao, and S.~Zhou.
\newblock Mining distance-based outliers from large databases in any metric
  space.
\newblock In {\em Proceedings of the 12th ACM SIGKDD international conference
  on Knowledge discovery and data mining}, pages 394--403. ACM, 2006.

\bibitem{wu2006outlier}
M.~Wu and C.~Jermaine.
\newblock Outlier detection by sampling with accuracy guarantees.
\newblock In {\em Proceedings of the 12th ACM SIGKDD international conference
  on Knowledge discovery and data mining}, pages 767--772. ACM, 2006.

\bibitem{ertoz2004finding}
L.~Ert{\"o}z, M.~Steinbach, and V.~Kumar.
\newblock {\em Finding topics in collections of documents: A shared nearest
  neighbor approach}.
\newblock Springer, 2004.

\bibitem{yu2002findout}
D.~Yu, G.~Sheikholeslami, and A.~Zhang.
\newblock Findout: finding outliers in very large datasets.
\newblock {\em Knowledge and Information Systems}, 4(4):387--412, 2002.

\bibitem{budalakoti2006anomaly}
S.~Budalakoti, A.~Srivastava, R.~Akella, and E.~Turkov.
\newblock Anomaly detection in large sets of high-dimensional symbol sequences.
\newblock {\em NASA Ames Research Center, Tech. Rep. NASA TM-2006-214553},
  2006.

\bibitem{pires2005using}
A.~Pires and C.~Santos-Pereira.
\newblock Using clustering and robust estimators to detect outliers in
  multivariate data.
\newblock In {\em Proceedings of the International Conference on Robust
  Statistics}, 2005.

\bibitem{solberg2005detection}
H.~E. Solberg and A.~Lahti.
\newblock Detection of outliers in reference distributions: performance of
  horn?s algorithm.
\newblock {\em Clinical chemistry}, 51(12):2326--2332, 2005.

\bibitem{aggarwal2008outlier}
C.~C. Aggarwal and P.~S. Yu.
\newblock Outlier detection with uncertain data.
\newblock In {\em SDM}, volume 483, page 493. SIAM, 2008.

\bibitem{chen2005simultaneous}
D.~Chen, X.~Shao, B.~Hu, and Q.~Su.
\newblock Simultaneous wavelength selection and outlier detection in
  multivariate regression of near-infrared spectra.
\newblock {\em Analytical Sciences}, 21(2):161--166, 2005.

\bibitem{agarwal2007detecting}
D.~Agarwal.
\newblock Detecting anomalies in cross-classified streams: a bayesian approach.
\newblock {\em Knowledge and information systems}, 11(1):29--44, 2007.

\bibitem{he2005optimization}
Z.~He, S.~Deng, and X.~Xu.
\newblock An optimization model for outlier detection in categorical data.
\newblock In {\em Advances in Intelligent Computing}, pages 400--409. Springer,
  2005.

\bibitem{ando2007clustering}
S.~Ando.
\newblock Clustering needles in a haystack: An information theoretic analysis
  of minority and outlier detection.
\newblock In {\em Data Mining, 2007. ICDM 2007. Seventh IEEE International
  Conference on}, pages 13--22. IEEE, 2007.

\bibitem{keogh2004towards}
E.~Keogh, S.~Lonardi, and C.~A. Ratanamahatana.
\newblock Towards parameter-free data mining.
\newblock In {\em Proceedings of the tenth ACM SIGKDD international conference
  on Knowledge discovery and data mining}, pages 206--215. ACM, 2004.

\bibitem{agovic20086}
A.~Agovic, A.~Banerjee, A.~R. Ganguly, and V.~Protopopescu.
\newblock Anomaly detection in transportation corridors using manifold
  embedding.
\newblock {\em Knowledge Discovery from Sensor Data}, pages 81--105, 2008.

\bibitem{dutta2007distributed}
H.~Dutta, C.~Giannella, K.~D. Borne, and H.~Kargupta.
\newblock Distributed top-k outlier detection from astronomy catalogs using the
  demac system.
\newblock In {\em SDM}, pages 473--478. SIAM, 2007.

\bibitem{ide2004eigenspace}
T.~Ide and H.~Kashima.
\newblock Eigenspace-based anomaly detection in computer systems.
\newblock In {\em Proceedings of the tenth ACM SIGKDD international conference
  on Knowledge discovery and data mining}, pages 440--449. ACM, 2004.

\bibitem{shyu2003novel}
M.-L. Shyu, S.-C. Chen, K.~Sarinnapakorn, and L.~Chang.
\newblock A novel anomaly detection scheme based on principal component
  classifier.
\newblock Technical report, DTIC Document, 2003.

\bibitem{breunig2000lof}
M.~M. Breunig, H.-P. Kriegel, R.~T. Ng, and J.~Sander.
\newblock Lof: identifying density-based local outliers.
\newblock In {\em ACM sigmod record}, volume~29, pages 93--104. ACM, 2000.

\bibitem{kriegel2008angle}
H.-P. Kriegel, A.~Zimek, et~al.
\newblock Angle-based outlier detection in high-dimensional data.
\newblock In {\em Proceedings of the 14th ACM SIGKDD international conference
  on Knowledge discovery and data mining}, pages 444--452. ACM, 2008.

\bibitem{ahmed2009online}
T.~Ahmed.
\newblock Online anomaly detection using kde.
\newblock In {\em Global Telecommunications Conference, 2009. GLOBECOM 2009.
  IEEE}, pages 1--8. IEEE, 2009.

\bibitem{SaliencyAIM}
N.~Bruce and J.~Tsotsos.
\newblock Saliency based on information maximization.
\newblock In {\em Advances in neural information processing systems}, pages
  155--162, 2005.

\bibitem{zhang2008sun}
L.~Zhang, M.~H. Tong, T.~K. Marks, H.~Shan, and G.~W. Cottrell.
\newblock {SUN}: A {B}ayesian framework for saliency using natural statistics.
\newblock {\em Journal of vision}, 8(7):32, 2008.

\bibitem{beyer1999nearest}
K.~Beyer, J.~Goldstein, R.~Ramakrishnan, and U.~Shaft.
\newblock When is ``nearest neighbor'' meaningful?
\newblock In {\em Database Theory?ICDT'99}, pages 217--235. Springer, 1999.

\bibitem{elhamifar2009sparse}
E.~Elhamifar and R.~Vidal.
\newblock Sparse subspace clustering.
\newblock In {\em Computer Vision and Pattern Recognition, 2009. CVPR 2009.
  IEEE Conference on}, pages 2790--2797. IEEE, 2009.

\bibitem{elhamifar2013sparse}
E.~Elhamifar and R.~Vidal.
\newblock Sparse subspace clustering: Algorithm, theory, and applications.
\newblock {\em Pattern Analysis and Machine Intelligence, IEEE Transactions
  on}, 35(11):2765--2781, 2013.

\bibitem{wang2016noisy}
Y.-X. Wang and H.~Xu.
\newblock Noisy sparse subspace clustering.
\newblock {\em Journal of Machine Learning Research}, 17(12):1--41, 2016.

\bibitem{vidal2010tutorial}
R.~Vidal.
\newblock A tutorial on subspace clustering.
\newblock {\em IEEE Signal Processing Magazine}, 28(2):52--68, 2010.

\bibitem{groupsparse_ssp}
D.~Pimentel-Alarcon, L.~Balzano, R.~Marcia, R.~Nowak, and R.~Willett.
\newblock Group-sparse subspace clustering with missing data.
\newblock In {\em Proc. Statistical Signal Processing Workshop}, 2016.

\bibitem{allard2012multi}
W.~K. Allard, G.~Chen, and M.~Maggioni.
\newblock Multi-scale geometric methods for data sets ii: Geometric
  multi-resolution analysis.
\newblock {\em Applied and Computational Harmonic Analysis}, 32(3):435--462,
  2012.

\bibitem{roweis2000nonlinear}
S.~T. Roweis and L.~K. Saul.
\newblock Nonlinear dimensionality reduction by locally linear embedding.
\newblock {\em Science}, 290(5500):2323--2326, 2000.

\bibitem{belkin2003laplacian}
M.~Belkin and P.~Niyogi.
\newblock Laplacian eigenmaps for dimensionality reduction and data
  representation.
\newblock {\em Neural computation}, 15(6):1373--1396, 2003.

\bibitem{ChenSilvaPaisley2012}
M.~Chen, J.~Silva, J.~Paisley, C.~Wang, D.~Dunson, and L.~Carin.
\newblock Compressive sensing on manifolds using a non-parametric mixture of
  factor analyzers: algorithm and performance bounds.
\newblock {\em IEEE Trans. on Signal Processing}, 58(12):6140 -- 6155, 2010.

\bibitem{xie2013change}
Y.~Xie, J.~Huang, and R.~Willett.
\newblock Change-point detection for high-dimensional time series with missing
  data.
\newblock {\em Selected Topics in Signal Processing, IEEE Journal of},
  7(1):12--27, 2013.

\bibitem{sampleimage}
Unsplash, 2015.
\newblock \url{https://pixabay.com/en/street-alley-trees-lined-road-828889/}.

\bibitem{donoho97cart}
D.~Donoho.
\newblock Cart and best-ortho-basis selection: A connection.
\newblock {\em Annals of Stat.}, 25:1870--1911, 1997.

\bibitem{CART}
L.~Breiman, J.~Friedman, R.~Olshen, and C.~J. Stone.
\newblock {\em Classification and Regression Trees}.
\newblock Wadsworth, Belmont, CA, 1983.

\bibitem{willett:tmi03}
R.~Willett and R.~Nowak.
\newblock Platelets: a multiscale approach for recovering edges and surfaces in
  photon-limited medical imaging.
\newblock {\em IEEE Transactions~Med.~Imaging}, 22(3):332--350, 2003.

\bibitem{willett:jsac04}
R.~Nowak, U.~Mitra, and R.~Willett.
\newblock Estimating inhomogeneous fields using wireless sensor networks.
\newblock {\em IEEE Journal on Selected Areas in Communications},
  22(6):999--1006, 2004.

\bibitem{WillettNowak2005}
R.~Willett and R.~Nowak.
\newblock Level set estimation via trees.
\newblock In {\em Proc. of ICASSP}, 2005.

\bibitem{scott2006minimax}
C.~Scott, R.~D. Nowak, et~al.
\newblock Minimax-optimal classification with dyadic decision trees.
\newblock {\em IEEE transactions on information theory}, 52(4):1335--1353,
  2006.

\bibitem{willett:density}
R.~Willett and R.~Nowak.
\newblock Multiscale {P}oisson intensity and density estimation.
\newblock {\em IEEE Transactions Information Theory}, 53(9):3171--3187, 2007.

\bibitem{woodbury1950inverting}
M.~A. Woodbury.
\newblock Inverting modified matrices.
\newblock {\em Memorandum report}, 42:106, 1950.

\bibitem{harville1998matrix}
D.~A. Harville.
\newblock Matrix algebra from a statistician's perspective.
\newblock {\em Technometrics}, 40(2):164--164, 1998.

\bibitem{borji2013state}
A.~Borji and L.~Itti.
\newblock State-of-the-art in visual attention modeling.
\newblock {\em Pattern Analysis and Machine Intelligence, IEEE Transactions
  on}, 35(1):185--207, 2013.

\bibitem{lowe1999object}
D.~G. Lowe.
\newblock Object recognition from local scale-invariant features.
\newblock In {\em Computer vision, 1999. The proceedings of the seventh IEEE
  international conference on}, volume~2, pages 1150--1157. Ieee, 1999.

\bibitem{vedaldi08vlfeat}
A.~Vedaldi and B.~Fulkerson.
\newblock {VLFeat}: An open and portable library of computer vision algorithms.
\newblock \url{http://www.vlfeat.org/}, 2008.

\bibitem{lee2009video}
K.-Y. Lee, Y.-Y. Chuang, B.-Y. Chen, and M.~Ouhyoung.
\newblock Video stabilization using robust feature trajectories.
\newblock In {\em Computer Vision, 2009 IEEE 12th International Conference on},
  pages 1397--1404. IEEE, 2009.

\bibitem{matsushita2005full}
Y.~Matsushita, E.~Ofek, X.~Tang, and H.-Y. Shum.
\newblock Full-frame video stabilization.
\newblock In {\em Computer Vision and Pattern Recognition, 2005. CVPR 2005.
  IEEE Computer Society Conference on}, volume~1, pages 50--57. IEEE, 2005.

\bibitem{blei2003latent}
D.~M. Blei, A.~Y. Ng, and M.~I. Jordan.
\newblock Latent dirichlet allocation.
\newblock {\em the Journal of machine Learning research}, 3:993--1022, 2003.

\bibitem{diesner2005communication}
J.~Diesner, T.~L. Frantz, and K.~M. Carley.
\newblock Communication networks from the {E}nron email corpus ``{I}t's always
  about the people. {E}nron is no different''€.
\newblock {\em Computational and Mathematical Organization Theory},
  11(3):201--228, 2005.

\bibitem{aggarwal2012event}
C.~C. Aggarwal and K.~Subbian.
\newblock Event detection in social streams.
\newblock In {\em SDM}, volume~12, pages 624--635. SIAM, 2012.

\bibitem{raginsky2012sequential}
M.~Raginsky, R.~M. Willett, C.~Horn, J.~Silva, and R.~F. Marcia.
\newblock Sequential anomaly detection in the presence of noise and limited
  feedback.
\newblock {\em Information Theory, IEEE Transactions on}, 58(8):5544--5562,
  2012.

\bibitem{horn2011online}
C.~Horn and R.~Willett.
\newblock Online anomaly detection with expert system feedback in social
  networks.
\newblock In {\em Acoustics, Speech and Signal Processing (ICASSP), 2011 IEEE
  International Conference on}, pages 1936--1939. IEEE, 2011.

\bibitem{anscombe1948transformation}
F.~J. Anscombe.
\newblock The transformation of poisson, binomial and negative-binomial data.
\newblock {\em Biometrika}, 35(3/4):246--254, 1948.

\bibitem{darling2011theoretical}
W.~M. Darling.
\newblock A theoretical and practical implementation tutorial on topic modeling
  and {G}ibbs sampling.
\newblock In {\em Proceedings of the 49th annual meeting of the association for
  computational linguistics: Human language technologies}, pages 642--647,
  2011.

\bibitem{anscombe}
F.~J. Anscombe.
\newblock The transformation of poisson, binomial and negative-binomial data.
\newblock {\em Biometrika}, 35(3/4):246--254, 1948.

\bibitem{Dec13}
CNN Money.
\newblock Enron names new {CEO}, 2000.
\newblock \url{http://money.cnn.com/2000/12/13/companies/enron/index.htm}.

\bibitem{May09}
New~York Times.
\newblock California utility says prices of gas were inflated, 2001.
\newblock
  \url{http://www.nytimes.com/2001/05/09/business/09GAS.html?ex=994046400&en=1809786a77861861&ei=5039}.

\bibitem{May08}
CNN.
\newblock Power turned off in california again, 2001.
\newblock \url{http://www.cnn.com/2001/US/05/08/calif.power.crisis.02/}.

\bibitem{ousts}
F.~Norris.
\newblock Enron ousts finance chief as s.e.c. looks at dealings, 2001.
\newblock
  \url{http://www.nytimes.com/2001/10/25/business/enron-ousts-finance-chief-as-sec-looks-at-dealings.html}.

\bibitem{Nov28}
CNN Money.
\newblock Dynegy scraps {E}nron deal, 2001.
\newblock \url{http://money.cnn.com/2001/11/28/companies/enron/}.

\bibitem{Jan29}
CNN Money.
\newblock Enron takes {C}ooper, 2002.
\newblock \url{http://money.cnn.com/2002/01/29/companies/enron_ceo/}.

\bibitem{Jan30}
CNN Money.
\newblock Enron sells metals unit, 2002.
\newblock \url{http://money.cnn.com/2002/01/30/deals/enron_sempra/}.

\end{thebibliography}
\end{document}